\RequirePackage[hyphens]{url}
\documentclass[mnsc,nonblindrev]{informs3h}
\OneAndAHalfSpacedXI

\usepackage[round]{natbib}
\usepackage{multibib}
\def\bibfont{\small}

\newcites{App}{References for Appendix}

\usepackage{secdot}
\usepackage{epsfig}
\usepackage{amssymb}
\usepackage{amsmath}
\usepackage{color}
\usepackage{graphicx}
\usepackage{multirow}
\usepackage{enumerate}
\usepackage{subfig}
\usepackage{natbib}

 \bibpunct[, ]{(}{)}{,}{a}{}{,}%
 \def\bibfont{\small}%
\usepackage[ruled,linesnumbered]{algorithm2e}
\usepackage[normalem]{ulem}
\usepackage{float}
\usepackage{enumitem}
\usepackage{amsmath}
\usepackage{epstopdf}
\usepackage{graphicx}
\usepackage{tikz}
\usepackage{pgfplots}
\usepackage{subfig}
\usepackage{tabularx}
\pgfplotsset{compat=newest}
\usepackage{pgfplotstable}
\usetikzlibrary{arrows.meta,babel,calc,intersections,backgrounds,patterns,plotmarks,decorations.pathreplacing}
\usepackage{multirow}
\usepgfplotslibrary{fillbetween}
\tikzstyle{every picture} += [>=stealth]
\makeatother
\usepackage{lipsum}
\usepackage[justification=centering]{caption}
\usepackage{booktabs}
\usepackage[hyphens]{url}
\usepackage{mathtools}

\TheoremsNumberedThrough    
\ECRepeatTheorems
\EquationsNumberedThrough  
\makeatletter
\newcommand\footnoteref[1]{\protected@xdef\@thefnmark{\ref{#1}}\@footnotemark}
\makeatother
\usepackage{apptools}
\usepackage{bbm}
\usepackage{chngcntr}
\usepackage{lipsum}
\usepackage{float}
\usepackage{hyperref}
\usepackage{cleveref}
\Crefname{line}{}{}
\usepackage{bm}
\hypersetup{
    hidelinks,
    colorlinks=true,
    linkcolor=black,
    citecolor=blue
} 

\usepackage{enotez}
\let\footnote=\endnote
\setenotez{backref = true}
\setenotez{list-name = {Endnotes}}
\usepackage[hang,flushmargin]{footmisc}
\makeatletter
\def\footnoterule{\relax
  \kern 1pt
  \hbox to \columnwidth{\vrule width 0.5\columnwidth height 0.5pt\hfill}
  \kern 1pt}
\makeatother

\tikzset{
 hatch distance/.store in=\hatchdistance,
 hatch distance=10pt,
 hatch thickness/.store in=\hatchthickness,
 hatch thickness=2pt
 }
 \makeatletter
 \pgfdeclarepatternformonly[\hatchdistance,\hatchthickness]{flexible hatch}
 {\pgfqpoint{0pt}{0pt}}
 {\pgfqpoint{\hatchdistance}{\hatchdistance}}
 {\pgfpoint{\hatchdistance-1pt}{\hatchdistance-1pt}}%
 {
 \pgfsetcolor{\tikz@pattern@color}
 \pgfsetlinewidth{\hatchthickness}
 \pgfpathmoveto{\pgfqpoint{0pt}{0pt}}
 \pgfpathlineto{\pgfqpoint{\hatchdistance}{\hatchdistance}}
 \pgfusepath{stroke}
 }
\makeatother
\usetikzlibrary{shapes.geometric, arrows, calc, shapes.misc, matrix, shapes, arrows, calc, positioning, shapes.geometric, shapes.symbols, shapes.misc, intersections, chains, scopes}
\tikzstyle{node1} = [circle, circle sides=6,minimum width=1.25cm, text badly centered, text width=2.35em,minimum height=1.15cm, draw=black, font=\small]%,fill=red!40
\tikzstyle{node2} = [rectangle,rounded corners, minimum width=1.25cm, minimum height=1cm, text-centered, draw=black, font=\small]%fill=orange!40,
\tikzstyle{arrow} = [thick,->,>=stealth]

\newcommand{\E}{\mathbb{E}}
\newcommand{\Var}{\textbf{Var}}
\newcommand{\muone}{\mu^{(1)}}
\newcommand{\muzero}{\mu^{(0)}}
\newcommand{\sigmaone}{\sigma^{(1)}}
\newcommand{\sigmazero}{\sigma^{(0)}}
\newcommand{\F}{\mathcal{F}}
\newcommand{\ALG}{\texttt{ALG}}
\begin{document}%%%%%%%%%%%%%%%%
%%%%%%%%%%%%%%%%%%%%%%%%%%%%%%%%

\RUNAUTHOR{Li et al.}
\RUNTITLE{Optimal Adaptive Experimental Design for Estimating Treatment Effect}
\TITLE{Optimal Adaptive Experimental Design for Estimating Treatment Effect\footnote{Authors are listed in alphabetical order.}}
\ARTICLEAUTHORS{
 \AUTHOR{Jiachun Li}
 \AFF{Laboratory for Information and Decision Systems, MIT, \url{jiach334@mit.edu}}
 \AUTHOR{David Simchi-Levi}
 \AFF{Laboratory for Information and Decision Systems, MIT, \url{dslevi@mit.edu}}
  \AUTHOR{Yunxiao Zhao}
 \AFF{School of Mathematics, Peking University, \url{zhaoyunxiao@stu.pku.edu.cn}}
}

\ABSTRACT{%
\textbf{Abstract.}
Given $n$ experiment subjects with potentially heterogeneous covariates and two possible treatments, namely active treatment and control, 
this paper addresses the fundamental question of determining the optimal accuracy in estimating the treatment effect.
Furthermore, we propose an experimental design that approaches this optimal accuracy, giving a (non-asymptotic) answer to this fundamental yet still open question. The methodological contributions are listed as follows.
First, we establish an idealized optimal estimator with minimal variance as benchmark, and then demonstrate that adaptive experiment is necessary to achieve near-optimal estimation accuracy. Secondly, by incorporating the doubly robust method into sequential experimental design, we frame the optimal estimation problem as an online bandit learning problem, bridging the two fields of statistical estimation and bandit learning. Using tools and ideas from both bandit algorithm design and adaptive statistical estimation, we propose a general low switching adaptive experiment framework, which could be a generic research paradigm for a wide range of adaptive experimental designs. Through novel lower bound techniques for non-i.i.d. data, we demonstrate the optimality of our proposed experiment.
Numerical result indicates that the estimation accuracy approaches optimal with as few as two or three policy updates.

}
\maketitle
%%%%%%%%%%%%%%%%%%%%%
%%%%%
%%%%%
%%%%%
%%%%%
%%%%%
%%%%%
%%%%% This is a new section !
%%%%%
%%%%%
%%%%%
%%%%%
%%%%%
%%%%%
%%%%%%%%%%%%%%%%%%%%%
\vspace{-2mm}
\section{Introduction}\label{sec:intro}

\subsection{Background}\label{subsec:background}
%\textcolor{red}{Part 1: Background}

Over the past several decades, experimental design has emerged as a central research problem in both causal inference and the broader field of statistics. One of the most prominent methodologies within this domain is the Randomized Controlled Trial (RCT), which has become extensively utilized in fields such as clinical trials and recommendation systems (\cite{uk1996uk}, \cite{gilotte2018offline}). In these settings, participants are typically assigned to either Group A (standard treatment) or Group B (new treatment), allowing researchers to estimate the differences between alternative policies or interventions. This approach has 
been instrumental in advancing knowledge across various fields (\cite{zabor2020randomized},\cite{kohavi2020online}), contributing to our understanding of treatment effects and policy impact.

Efficiency is one of the most critical concerns in experimental design, largely due to the typically high costs associated with conducting experiments, ranging from tens to hundreds of millions 
(\cite{martin2017much}) and also time constraints like developing vaccines for COVID-19 during the pandemic.  Consider the following illustrative scenario: researchers seek to estimate the efficacy of a new drug and determine whether it is more effective than the traditional treatment. In such experiments, the number of available patients is often limited, especially for rare diseases, and the associated costs, such as extensive patient testing and monitoring, the treatment process, and data collection, are substantial (\cite{wong2014examination}).
Given a fixed number of 
$n$ experimental subjects, this leads to a natural question:

\textit{What is the best possible accuracy for estimating the average treatment effect (ATE), and can an efficient, robust experimental design be developed to achieve this level of accuracy?}

Following the aforementioned illustrative example, in the absence of prior knowledge, the most straightforward and theoretically optimal experimental design would be a randomized controlled trial (RCT) with equal probability of assignment to the treatment and control groups. However, if we had access to a clairvoyant who could inform us that the outcome variance of the treatment (i.e., the new drug) is significantly higher, with $(\sigma^{(1)})^2=81$, while the variance of the existing drug (control) is relatively small, with 
$(\sigma^{(0)})^2=1$, then the experimental design should be adjusted accordingly. Specifically, the experiment should allocate the majority of patients, approximately $90\%$, to the treatment group to account for the greater uncertainty in the new drug's effectiveness. This is known as \textbf{Neyman allocation} proposed in the seminal work (\cite{neyman1992two}), which recommends that the sizes of the treated and control groups should be proportional to their
respective standard deviations to maximize the variance of different in means estimator. But without the prior information about variance, any non-adaptive experimental design can be sub-optimal in some cases, as proved in theorem 1 in (\cite{zhao2023adaptive}):

\noindent  \textbf{Theorem (Informal):}  For any fixed allocation mechanism which decides the size of treatment and control group $(n(1),n(0))$ before the experiment, the variance of treatment effect estimation under this design will be at least twice as large as the optimal design in some cases.

Thus, a natural approach to improving experimental efficiency is to leverage information from historical experimental results to design more efficient future experiments. In the context of our example, this could involve estimating the variance of both the treatment and control groups, and subsequently reallocating the remaining participants between these two groups. This concept forms the foundation of adaptive experimental design, which can be traced back to the seminal work of Chernoff in the 1950s (\cite{chernoff1992sequential}),
which shows that adaptive sampling can reduce the sample size for testing hypothesis.
Since then, extensive research has been conducted, primarily driven by clinical trials, but adaptive design has also gained significant traction in A/B testing within major tech companies like Uber and Netflix (\cite{bojinov2023design},\cite{bojinov2022online}). A well-established framework, known as response-adaptive randomization, has been thoroughly studied. The central premise is to first determine the optimal allocation proportion between the treatment and control groups, considering parameters such as the expected outcome and variance for both groups. The experiment is then designed so that the allocation ratio between the treatment and control groups gradually converges toward the optimal proportion. This optimal ratio is defined by various criteria, such as minimizing the variance of the treatment effect estimate or minimizing the variance of the Wald test (\cite{hu2009efficient}).
When covariate information is available, the covariate-adjusted response-adaptive randomization scheme is considered. This approach typically seeks to balance the covariate distribution between the treatment and control groups (\cite{zhang2007asymptotic}). These methods have been successfully applied in clinical trials (\cite{rosenberger2012adaptive}). More recently, specific efforts have focused on minimizing treatment effect estimation error, particularly through designing experiments that achieve Neyman allocation (\cite{zhao2023adaptive},\cite{dai2023clip}).

However, adaptive methods have faced recurring critiques, particularly regarding their practical implementation and statistical validity. A primary concern is the increased risk of falsely detecting treatment effects, prematurely discarding promising therapies, and introducing statistical or operational bias, as emphasized by FDA guidance(\cite{bothwell2018adaptive}). One of the most prominent criticisms is the complexity of adaptive methods, which require continuous system maintenance and policy updates. This complexity leads to significant operational costs, especially in large-scale experiments such as those in tech companies (\cite{bojinov2022online}).
A second common concern is the non-robustness of many adaptive methods. Outliers can disproportionately influence the experiment, leading to severely biased estimates of treatment effects and potentially resulting in the early and unjustified rejection of an effective treatment. In clinical trials, another challenge is the delay in feedback from patients, which can take weeks. This delay makes many adaptive methods infeasible, as they require outcome information from each patient before determining the allocation mechanism for the next. Such delays hinder the applicability of adaptive designs, particularly in settings where timely decision-making is critical.

The final, and perhaps most nuanced, concern is that although adaptive experimental methods are initially designed to leverage historical information to improve statistical accuracy, there is, in fact, \textbf{no non-asymptotic statistical guarantee for the output of an adaptive experiment}. This arises because, when the data-generating process exhibits strong correlations, standard statistical properties, such as high probability concentration and the analysis of bias and variance under the assumption of i.i.d. (independent and identically distributed) data, no longer hold. Consequently, while under an i.i.d. setting, an optimal design corresponds to an optimal allocation between the treatment and control groups, achieving this allocation adaptively provides no statistical guarantee for the experiment's output, as the data are no longer independent.
Thus, while adaptivity can enhance decision-making efficiency, excessive adaptivity undermines the very statistical properties that are essential for drawing valid inferences. In other words, over-adaptation can compromise the reliability of the experiment's conclusions.
At this stage, a natural question regarding adaptive experimental design arises:

\textit{Is it possible to design an adaptive experiment that is easy to implement, robust, maintains all the desired statistical guarantees, and still significantly enhances statistical accuracy?}

In the remaining sections of this paper, we provide a positive response to this question. Specifically, we introduce a general ``learning to design" framework, which is ``almost non-adaptive" in nature, retains the statistical guarantees typically associated with i.i.d. data, and achieves the best possible statistical accuracy in a rapid manner.

The paper is organized as follows. In section \ref{subsec-liter}, we go through the related literature and then describe the technical difficulties as well as methodology contributions in section \ref{subsec-contribution}. In section \ref{sec:formulation}, we formalize the problem setting as well as important definitions and notations. 
Then in section \ref{sec:nofeature}, we first introduce our framework to resolve the well-known optimal Neyman allocation problem and give the first finite time guarantee and analysis as a warm-up. Then in section \ref{sec:hetero-effect}, we consider the more challenging case with covariates and show that the optimal semi-parametric 
efficiency bound can be approached rapidly. We provide numerical results in section \ref{sec:numerical}, which shows that our method is extremely efficient, easy to implement, and significantly improves the statistical estimation accuracy. In section \ref{sec:future}, we point out several future directions that arise naturally under our framework, which shows that the ``learning to design" idea could be a general principle for broad class of adaptive experiment settings.
Finally, we conclude in section \ref{sec:conclusion}.

\subsection{Literature Review}\label{subsec-liter}
%We now start by establishing our problem settings.
\noindent\textbf{Adaptive experimental design.} 
This paper focuses on how adaptivity can improve statistical estimation accuracy. As previously mentioned, the framework for (covariate-adjusted) response-adaptive randomization is well established. Its core idea is to determine the optimal allocation ratio between the treatment and control groups and to develop an allocation mechanism that asymptotically approaches this ratio. Comprehensive overviews of these techniques can be found in standard texts (\cite{hu2006theory}). When covariate information is available, existing works provide allocation schemes aimed at balancing covariates between the treatment and control groups, a widely-used heuristic in experimental design (\cite{rosenberger2012adaptive},\cite{harshaw2024balancing}).
In contrast, this paper directly characterizes the optimal allocation mechanism in terms of semi-parametric efficiency bounds to minimize the treatment effect estimation error (\cite{hahn2011adaptive}). Extensions, such as handling delayed feedback or unobserved covariates, have also been explored (\cite{hu2008doubly},\cite{liu2023impacts}).
Recently, researchers in operations research have turned their attention to developing efficient and scalable A/B testing methodologies, increasingly popular in tech companies and online platforms (\cite{bojinov2023design},\cite{bojinov2022online}). An important area of study is dynamic treatment effect estimation, where past treatments influence future outcomes, and there is ongoing work to construct valid inferences in such settings (\cite{zhou2023offline}, \cite{hu2023off}). Another related line of research focuses on building valid confidence intervals for adaptively collected data, such as those generated by bandit or reinforcement learning algorithms (\cite{zhang2021statistical},\cite{zhang2022statistical}). However, none of these approaches offer non-asymptotic guarantees for the expectation or variance of the estimators, due to the complex correlations in the data-generating process. In many cases, it is even unknown whether the estimation is biased, except in certain specific scenarios (\cite{dimakopoulou2021online}). Most existing works rely on a martingale structure of the algorithm and apply the Martingale Central Limit Theorem to establish asymptotic normality (\cite{brown1971martingale}).
The most relevant works to this paper are (\cite{zhao2023adaptive},\cite{dai2023clip}), which aim to achieve Neyman allocation when variance is unknown and minimize estimation error. Additionally, \cite{zhao2023adaptive} was the first to adopt the low-switching experiment framework, which partially motivates this paper, though these works also lack non-asymptotic guarantees.

\noindent\textbf{ Doubly robust method in causal inference.}
The doubly robust method is a key tool for analyzing observational data in causal inference (\cite{chernozhukov2018double},\cite{kennedy2016semiparametric}), and it is also widely studied in the broader field of semi-parametric statistics (\cite{kennedy2017semiparametric},\cite{newey1994asymptotic},\cite{newey1990semiparametric}). The variance of any treatment effect estimator under i.i.d. data collection is bounded below by the semi-parametric efficiency bound, which extends the concepts of Fisher information and the Cramér-Rao lower bound (\cite{van2000asymptotic}). Notably, the doubly robust approach, specifically the AIPW (Augmented Inverse Probability Weighted) estimator, achieves the semi-parametric efficiency bound under mild conditions (\cite{funk2011doubly}), making it the gold standard for treatment effect evaluation.
Recent works have explored the application of doubly robust methods in adaptive experimental designs (\cite{kato2020efficient},\cite{kato2024active},\cite{cook2024semiparametric}), aiming to asymptotically reach the generalized Neyman allocation bound, which corresponds to the minimum possible variance. All previous works require consistent guarantees for the estimation of mean and variance functions based on certain assumptions, subsequently proposing adaptive designs and establishing asymptotic normality under these conditions. These designs exhibit a high degree of adaptivity; however, constructing such estimators with the desired guarantees remains challenging (particularly in a non-asymptotic framework). In this paper, we propose a more robust, efficient, and general framework that offers non-asymptotic statistical guarantees while preserving all the advantageous properties of adaptive experimental designs. Additionally, we introduce a framework that enables the transformation of an optimal adaptive experiment into a low-switching (contextual) bandit optimization, which can be applied in broader settings.

\noindent\textbf{Low switching (Contextual) bandit optimization.}
One of the most successful adaptive decision-making frameworks in the presence of missing data is the bandit framework (\cite{lai1985asymptotically}), which provides extensive results on how to adaptively collect information and make decisions to maximize total rewards. There are several key parallels between adaptive experimental design and bandit optimization: both aim to leverage historical data to enhance the accuracy of future decisions, and both only observe the outcomes of chosen actions (while un-chosen outcomes remain unobserved, which gives rise to the term ``bandit"), and both strive to minimize certain loss functions—cumulative regret in bandit optimization and estimation error in adaptive experimental design.

Given these similarities, many researchers have explored the application of bandit frameworks to experimental design (\cite{simchi2023multi}, \cite{li2024privacy}). Increasing interest has emerged around the question: can adaptive experimental design be viewed as a specific instance of bandit learning, allowing the adoption of well-established bandit methods to address adaptive design problems systematically under a unified framework? This paper provides such a framework by demonstrating that \textbf{an optimal experimental design is equivalent to a low-switching bandit optimization algorithm}. Low-switching algorithms have been extensively studied in bandit, contextual bandit, and reinforcement learning settings, and solutions for general settings have been well-established (\cite{simchilevi2021bypassing}, \cite{bai2019provably}, \cite{hanna2024efficient}). Consequently, many questions in adaptive experimental design could be resolved through the adoption of low-switching bandit algorithms. In this paper, we address two such problems: achieving the optimal Neyman allocation and the generalized Neyman allocation bound.

\noindent\textbf{Variance function estimation.}
The optimal allocation in minimizing the estimation error of treatment effect is only related to the variance of the outcomes of treatment and control, but not the expectation. Therefore, in this paper, we will need guarantees of (non-parametric) regression rates for the variance function. For the case without covariate, it reduces to estimate a single parameter, and we provide proof based on the concentration of U-statistics\cite{Bentkus2009TailProbabilities} in the appendix. When there exist covariates, this is still quite an open question, as we can only observe the outcomes, not the variance of them. For one dimension case, the minimax optimal rate has been proved for the smoothness class under certain regularity conditions (\cite{wang2008effect}, \cite{shen2020optimal}), while for high dimensional case, there are some results in fixed design (\cite{cai2009variance}), but a minimax rate for random design is still open in general. Within this paper, we will always assume that such a non-parametric regression oracle for variance function exists, which is the standard assumption in oracle-based learning algorithms (\cite{simchilevi2021bypassing}).   

\subsection{Technical Difficulties and Methodology Contribution}\label{subsec-contribution}

\textbf{More efficient and practical framework and paradigm with statistical guarantee.} As discussed earlier, a fundamental drawback of existing adaptive experimental designs is the lack of non-asymptotic statistical guarantees due to the correlated data-generating process. The key contribution of this paper is the observation that frequent policy updates are unnecessary; in fact, even a single update is sufficient to achieve asymptotic optimality (see the algorithms in Sections \ref{sec:nofeature} and \ref{sec:hetero-effect} for the case where the batch number 
$K=2$). While additional updates may improve the convergence rate, 
$O(\log n)$ updates are sufficient to attain optimal rates for most online learning tasks.
The advantage of low-switching is that data generated within each batch remain independent, making large-sample behavior easier to characterize. Although correlations exist between batches, these can be addressed using a union bound argument across the 
$K$ batches. This preserves the statistical guarantees associated with i.i.d. data, with only a negligible loss of 
$\text{polylog} (n)$. However, excessively frequent policy updates result in a polynomial loss in 
$n$, undermining the analysis.
The central insight of our framework is that \textbf{low-switching is both sufficient for learning unknown parameters and necessary for preserving optimal statistical guarantees, making it the ``best of both worlds."}
Moreover, a low switching experiment naturally solves all the issues about adaptive experiments as we mentioned. It's much more practical and less costly since the policy  
only needs to be updated very few times, typically no more than four as we show in our numerical results in section \ref{sec:numerical}. 
It's robust to outliers, missing data, or delayed feedback, as we only need the collection of data at the end of each batch, which is a much longer interval.

\noindent\textbf{Doubly robust method in adaptive experimental design.}
The core idea of the AIPW (Augmented Inverse Probability Weighted) estimator is that, if either the model estimation or the propensity score estimation is accurate, the treatment effect estimation will remain valid. In this paper, we introduce the doubly robust method into adaptive experimental design and demonstrate that similar to observational studies, an optimal experimental design can be divided into two key components: \textbf{model estimation} and \textbf{propensity score optimization}.
A key contribution of this work is that, instead of merely optimizing the allocation ratio as in existing methods, we focus directly on optimizing the statistical accuracy of the design. We argue that this is a more appropriate guiding principle for experimental design, as an improved design naturally leads to better model estimation, and vice versa. In contrast, optimizing the allocation ratio alone does not provide any statistical guarantees and cannot be easily extended to settings with covariates. Further discussions on our framework are provided at the end of Sections \ref{subsec:comparison} and \ref{sec:future}.

\noindent \textbf{New lower bound results and techniques for adaptive experimental design.}
In this paper, we propose new lower bounds for general statistical estimation error when data are collected adaptively. Traditional lower bounds, such as the Cramér-Rao theorem based on Fisher information, no longer apply directly because they rely on the assumption of independently collected data. 
In general exact non-asymptotic lower bound of statistical estimation for adaptively collected data are still quite open to the best of our knowledge. In this paper we prove two different types of lower bound.
For the first one, we  demonstrate that for any potentially adaptive experiment and unbiased estimator, the variance under (generalized) Neyman allocation serves as a valid non-asymptotic lower bound, representing the fundamental statistical limit of estimating treatment effect for any adaptive experimental design. This can be viewed as a generalization of the Cramer-Rao or semi-parametric efficiency lower bounds, where the bound is taken not only over all possible estimators but also over all possible algorithms (or equivalently, data-generating processes). Our approach is based on a more detailed analysis of Fisher information, and to the best of our knowledge, this argument is novel within the causal inference literature, due to the adaptive nature of the data-generating process.
For the second lower bound, in order to have a better analysis of lower bound capturing lower order terms, we adopt the approach outlined in (\cite{khamaru2021near}) using the Bayesian perspective, which is more appropriate for adaptively collected data, as it updates the posterior belief over time.
We also leverage information-theoretic lower bound techniques commonly used in the bandit literature to establish a lower bound on the convergence rate towards Neyman allocation. Both arguments and proof techniques could provide insights for analyzing statistical limit for adaptively collected data and might be of independent interests.

\section{Problem Formulation}
\label{sec:formulation}
There are $n$ different subjects participating in the experiment sequentially, where $n$ is fixed and known to the experimenter. 
The covariate $X$ is sampled from a distribution $P_X$ on some compact set $\mathcal{X}$. There are two possible allocations, namely treatment and control group. For given $X$, the outcome of treatment $Y^{(1)}(X)$ is generated from some distribution $P_{Y^{(1)}| X}$ with expectation $ \mu^{(1)}(X)$ and variance $(\sigma^{(1)}(X))^2$, and we can similarly define $Y^{(0)}(X)$,$P_{Y^{(0)}| X_i}$, $ \mu^{(0)}(X)$ and $(\sigma^{(0)}(X))^2$. We will assume that 
$ \mu^{(1)}(X)$, $ \mu^{(0)}(X)$ belongs to a (possibly non-parametric) bounded function class $\mathcal{F}_{\mu}$ and $\sigma^{(1)}(X)$, $\sigma^{(0)}(X)$ belongs to another function class $\mathcal{F}_{\sigma}$.
For the $i^{th}$ experiment subject, the experimenter observes the covariate $ X_i $ generated from $P_X$, 
then the experimenter assigns a random treatment $ W_i \in \{0,1\} $ to subject $ i $. We denote the propensity score $e_i(X_i):=\mathcal{P}(W_i=1 | X_i)$.
Finally, the result of treatment $ Y_i := Y^{(W_i)}(X_i) $ is generated and observed. Throughout this paper, we will assume that 
given feature $X_i$, the outcome $\{Y^{(1)}(X_i),Y^{(0)}(X_i)\}$ is conditionally independent of the treatment assignment $W_i(X_i)$. Also note that throughout this paper, \textbf{we don't assume that the variance $(\sigma^{(1)}(X))^2$, $(\sigma^{(0)}(X))^2$ are upper bounded, or the propensity score $\eta <e(x) <1-\eta$ for some constant $\eta$.} The latter condition is widely assumed in most causal inference literature known as ``overlap condition" (\cite{kennedy2016semiparametric}), but in cases where the variability between two treatments is extremely unbalanced, and the uncertainty of outcomes might be extremely large, our framework can still give a valid, reasonable estimation bound, which further shows the robustness of our framework.
We use the notation $ \mathcal{F}_i $ to denote all information collected in first $i$ experiment subjects, namely $ \{(X_1, W_1, Y_1), \cdots, (X_i, W_i, Y_i) \}$.
The distribution of treatment assignment $ W_i $ is a Bernoulli distribution with expectation (or the propensity score) $e(X_i)$. It is measured conditioned on the information available when $W_i$ is generated. For example, when $W_i$ is generated after observing covariate $X_i$, then $e(X_i)= \mathcal{P}(W_i=1 | \{X_i, \mathcal{F}_{i-1}\})$.
when $W_i$ is generated after  the $j$-th experiment for some $j<i$ before $X_{j+1}$ is generated, then $e(X_i)= \mathcal{P}(W_i=1 |  \mathcal{F}_{j})$, because in this case $X_i$ is no longer available information.

The causal effect of interest is the average treatment effect(ATE) defined as
\[
   \tau := \mathbb{E}[\mu^{(1)}(X) - \mu^{(0)}(X)],
\]
where the expectation is taken over $P_X$.
After collecting all data from the experiment, the experimenter constructs an estimator $ \widehat{\tau} $ of $ \tau $ 
which is a measurable function that depends on collected data $ \mathcal{F}_n $. 
The goal is to minimize the expected square loss of $ \widehat{\tau} $, $ \mathbb{E}[(\widehat{\tau} - \tau)^2] $, where the expectation is taken over all data and treatment randomness.

\section{A Warm-up: Optimal Adaptive Experimental Design without Covariate} \label{sec:nofeature}
In this section, we will first give an optimal experimental design to estimate the average treatment effect with homogeneous experiment units. This section can be regarded as a warm-up for the following, more complicated case with covariates, but we will see that the idea of doubly robust method is adopted in both scenarios, which helps to formulate the optimal adaptive experimental design problem as online optimization. Then we use Bayes argument and information theoretic lower bound to prove that our method is optimal among a natural class of experiments and estimators. 

Since we assume all experiment subjects are homogeneous in this section,
the covariate set $\mathcal{X}$ and the distribution $P_X$ degenerate, so we will abbreviate $ Y_i^{(1)}(X_i) $ as $ Y_i^{(1)} $, $ \mu^{(1)}(X) $ as $ \mu^{(1)} $ and $ \sigma^{(1)}(X) $ as $ \sigma^{(1)} $, and similarly for $Y_i^{(0)}$, $ \mu^{(0)} $ and $ \sigma^{(0)} $. 
We denote the propensity score $e_i=P(W_i=1)$ as the probability of unit $i$ to be allocated in the treatment group evaluated conditioned on the 
 available information based on which $W_i$ are generated. Throughout this section, we don't have any bounded assumptions for all the parameters $\mu^{(1)}$, $\mu^{(0)}$, $\sigma^{(1)}$, $\sigma^{(0)}$. 

\subsection{Two Types of Estimators} \label{subsec:two-estimators}

Assume that we have an experiment with collected dataset $\{W_i, Y_i\}_{i=1}^n$, the most natural estimator would be the difference in mean estimator:
\begin{equation}\label{mean-estimator}
    \begin{aligned}
        \widehat{\tau}_0= \frac{\sum_{i=1}^n Y_i W_i}{\sum_{i=1}^n W_i} -\frac{\sum_{i=1}^n Y_i (1-W_i)}{\sum_{i=1}^n (1-W_i)},
    \end{aligned}
\end{equation}
which is simply the difference
of mean outcome between the treatment group and the control group. Then an optimal experimental design will be reduced to finding the optimal allocation proportion of units assigned to treatment group $\sum_{i=1}^n W_i/n$. Indeed, this is the central idea most existing literature adopt (\cite{hu2006theory}). However, as we mentioned in section \ref{sec:intro}, in a sequential experiment, the indicator variable $W_i$ might be highly correlated, making the statistical analysis of estimator (\ref{mean-estimator}) messy (\cite{shin2021bias}).

Another widely adopted type of estimator in causal inference is the doubly robust estimator, which is known to be asymptotically efficient under mild conditions. Specifically, we will consider the augment IPW estimator in causal inference literature:
\begin{equation} \label{aipw-estimator}
    \begin{aligned}
        \widehat{\tau}_1=\widehat{\mu}^{(1)}-\widehat{\mu}^{(0)}
        +\sum_{i=1}^n \frac{1}{n} \left(\frac{Y_i-\widehat{\mu}^{(1)}}{e_i}W_i -\frac{Y_i-\widehat{\mu}^{(0)}}{1-e_i}(1-W_i) \right),
    \end{aligned}
\end{equation}
where $\widehat{\mu}^{(1)}$ and $\widehat{\mu}^{(0)}$ is the estimation for outcome mean $\mu^{(1)}$ and $\mu^{(0)}$.
Compared to the estimator (\ref{mean-estimator}), the AIPW estimator seems to be complicated and unnecessary. However, it has two significant advantages: \textbf{1. Takes the propensity score into account} and \textbf{2. Separate the (possibly correlated) indicator variable $W_i$}. Moreover, this estimator can be naturally extended to ATE estimation with covariates, while the mean difference estimator cannot. As we illustrate in the following, under the low switching experiment framework with some specific techniques, the AIPW estimator will be equivalent to a ``\textbf{weighted difference in mean estimator}", which is much easier and cleaner to analyze in adaptive experiment compared to the brutal difference in mean estimator.

\subsection{Introducing the Imaginary Intermediary Estimator} \label{subsec:intermediary}

A widely adopted idea used in analyzing the asymptotic efficiency of the doubly robust estimator in observational study is to introduce an imaginary intermediary estimator, and we introduce it into sequential experimental design to separate the estimation process into two parts: \textbf{model estimation} and \textbf{propensity score optimization}, which aligns with the idea of ``\textbf{doubly robust}". We then show that the second part, propensity score optimization, is naturally an online optimization problem.

Define the imaginary intermediary estimator as 
\begin{equation} \label{intermediary-estimator}
    \begin{aligned}
       \widehat{\tau}_2=\mu^{(1)}-\mu^{(0)}
        +\sum_{i=1}^n \frac{1}{n} \left(\frac{Y_i-\mu^{(1)}}{e_i}W_i -\frac{Y_i-\mu^{(0)}}{1-e_i}(1-W_i) \right),    \end{aligned}
\end{equation}
which uses the same experiment and dataset as estimator (\ref{aipw-estimator}), but has access to the underlying expectation of outcome. Similarly, we can define the imaginary optimal estimator
\begin{equation} \label{optimal-estimator}
    \begin{aligned}
        \widehat{\tau}^{*}=\mu^{(1)}-\mu^{(0)}
        +\sum_{i=1}^n \frac{1}{n} \left(\frac{Y_i-\mu^{(1)}}{e^*}W_i -\frac{Y_i-\mu^{(0)}}{1-e^*}(1-W_i) \right),
    \end{aligned}
\end{equation}
where $e^{*}=\sigma^{(1)}/(\sigma^{(1)}+\sigma^{(0)})$ is the optimal propensity score, and $\widehat{\tau}^*$ is the optimal estimator with variance $V^{*}=(\sigma^{(1)}+\sigma^{(0)})^2/n$, which is also the variance of Neyman allocation (\cite{neyman1992two}). Note that neither  $\widehat{\mu}_2$ nor $\widehat{\tau}^{*}$ is a feasible estimator, we only use them in analysis. Then we can decompose the mean square error of AIPW estimator $\widehat{\tau}_1$ as:
\begin{equation}
    \begin{aligned}
        \mathbb{E} \left( \widehat{\tau}_1- \tau \right)^2 -  \mathbb{E} \left( \widehat{\tau}^*- \tau \right)^2&=  \mathbb{E} \left( \widehat{\tau}_1- \widehat{\tau}_2+\widehat{\tau}_2- \tau \right)^2- \mathbb{E} \left( \widehat{\tau}^*- \tau \right)^2 \\
        &= \underbrace{\mathbb{E} \left( \widehat{\tau}_1- \widehat{\tau}_2\right)^2}_{\text{model estimation}} + \underbrace{\mathbb{E}\left(\widehat{\tau}_2- \tau \right)^2-\mathbb{E} \left( \widehat{\tau}^*- \tau \right)^2}_{\text{propensity score optimization}}\\
        & \qquad +\underbrace{2\mathbb{E} \left( (\widehat{\tau}_1- \widehat{\tau}_2)(\widehat{\tau}_2-\tau) \right)}_{\text{cross-term}}
    \end{aligned}
\end{equation}
Ideally, if we can prove that the expectation of cross term is $0$, then the experimental design task is separated into the two parts we promised above. This is achieved by the ``incomplete randomization" method, as described in the low switching adaptive experimental design below. But first, we will show that the propensity score optimization term $\mathbb{E}\left(\widehat{\tau}_2- \tau \right)^2-\mathbb{E} \left( \widehat{\tau}^*- \tau \right)^2$ is actually a standard bandit optimization.

\begin{lemma} \label{lem:propen-opti-nofeature}
For any ``non-anticipating" allocation mechanism and data generation process as described in section \ref{sec:formulation}, we have
     \begin{equation}
         \begin{aligned}     \mathbb{E}\left(\widehat{\tau}_2- \tau \right)^2-\mathbb{E} \left( \widehat{\tau}^*- \tau \right)^2= 
             \frac{1}{n} \E\left(\sum_{i=1}^n \frac{(\sigma^{(1)})^2}{e_i} +\sum_{i=1}^n\frac{(\sigma^{(0)})^2}{1-e_i}
             - n \left(\frac{(\sigma^{(1)})^2}{e^*} +\frac{(\sigma^{(0)})^2}{1-e^*}\right) \right),
         \end{aligned}
     \end{equation}
     where $e^*=\sigma^{(1)}/(\sigma^{(1)}+\sigma^{(0)})$. 
\end{lemma}
The most important information in lemma \ref{lem:propen-opti-nofeature} is that when comparing the variance of optimal estimator $\widehat{\tau}^*$ and $\widehat{\tau}_2$, no matter how the data is generated in a complex, adaptive manner, we only need to solve an online regret minimization problem, which has already been extensively studied. The only difference is that in propensity score optimization, instead of the expectation of the outcome distribution, we are interested in learning the variance. However, to have a good statistical guarantee for the model estimation 
\begin{equation}
    \begin{aligned}
        \mathbb{E} \left( \widehat{\tau}_1- \widehat{\tau}_2\right)^2= \mathbb{E} \left(\frac{1}{n} \sum_{i=1}^n \left(1-\frac{W_i}{e_i}\right) \left(\widehat{\mu}^{(1)}-\mu^{(1)}\right)-\frac{1}{n}\sum_{i=1}^n \left(1-\frac{1-W_i}{1-e_i}\right) \left(\widehat{\mu}^{(0)}-\mu^{(0)}\right)\right)^2,
    \end{aligned}
\end{equation}

 we need the data to be collected in a more clear, loosely correlated way. This motivates our low switching adaptive experimental design, which is a \textbf{low switching algorithm that tries to minimize the regret of propensity score optimization while maintaining a relatively simple process of data collection}.

\subsection{Low Switching Adaptive Experimental Design and Main Results} \label{subsec:exp-nofeature}
\begin{algorithm}[H]
   \caption{Low Switching Adaptive Experiment}
   \label{alg:ADM data collecting}
         \textbf{Input:} the number of periods $ n $, the number of batch $ K $  \\
   Denote $\widehat{\sigma}_0^{(0)} = \widehat{\sigma}_0^{(1)} = 0 $ and $\frac{0}{0+0} = \frac{1}{2}$\\
   \For {$k \in [K]$}{
   Set the batch length $ \gamma_1 = n^{\frac{1}{K}}(\log n)^{\frac{K-1}{K}} $ and for $ k \geq 2 $, $ \gamma_k = n^{\frac{k}{K}}(\log n)^{\frac{K - k}{K}}  - n^{\frac{k-1}{K}}(\log n)^{\frac{K - k + 1}{K}} $. \\
   Let truncated variance estimation be $ \widetilde{\sigma}_{k-1}^{(1)} := \max\{\widehat{\sigma}_{k-1}^{(1)}, \frac{\gamma_1}{\gamma_k}\widehat{\sigma}_{k-1}^{(0)} \} $, $ \widetilde{\sigma}_{k-1}^{(0)} := \max\{\widehat{\sigma}_{k-1}^{(0)},\frac{\gamma_1}{\gamma_k}\widehat{\sigma}_{k-1}^{(1)} \} $. \\
   Set $e_k=\min\{ \frac{1}{1+\gamma_1/\gamma_k }, \max\{\frac{\gamma_1/\gamma_k }{1+\gamma_1/\gamma_k }, \frac{\widehat{\sigma}_{k-1}^{ (1)} }{\widehat{\sigma}_{k-1}^{ (0)} + \widehat{\sigma}_{k-1}^{ (1)} }\}\}=\frac{\widetilde{\sigma}_{k-1}^{(1)}}{\widetilde{\sigma}_{k-1}^{(1)}+\widetilde{\sigma}_{k-1}^{(0)}}$.\\
      Uniformly randomly choose $ e_k\gamma_k $ periods in this batch and assign $ W_t = 1 $ for these periods. 
   Then set $ W_t = 0 $ for all remaining $ (1-e_k)\gamma_k $ periods. \\
   Use outcomes from treatment group in this batch $ \{Y_t : \Gamma_{k-1} + 1 \leq t \leq \Gamma_{k}, W_t = 1 \} $ and variance oracle to calculate $ \widehat{\sigma}_{k}^{(1)} \in \mathbb{R}_{\geq 0}  $. \\
   Use  outcomes from control group in this batch $ \{Y_t : \Gamma_{k-1} + 1 \leq t \leq \Gamma_{k}, W_t = 0 \} $ and variance oracle to calculate $ \widehat{\sigma}_{k}^{(0)} \in \mathbb{R}_{\geq 0}  $. \\
   }
\end{algorithm}
The low switching adaptive experiment described in algorithm \ref{alg:ADM data collecting} is clear, intuitive, and computationally efficient. Since the optimal allocation $e^{*}=\sigma^{(1)}/(\sigma^{(0)}+\sigma^{(1)})$, the most natural idea is to collect historical data, estimate the variance and plug it in. Indeed, this is exactly what the low switching adaptive experiment does, except for a lower threshold to enforce a certain level of exploration which guarantees that $\frac{\gamma_1/\gamma_k }{1+\gamma_1/\gamma_k } \le e_k \le \frac{1 }{1+\gamma_1/\gamma_k }$, which is also very natural in online learning literature (\cite{lattimore2020bandit}).
Note that it's equivalent to plug in the truncated variance estimation $\widetilde{\sigma}_{k-1}$, which equals to the variance estimation $ \widehat{\sigma}_{k-1}$ from estimation oracle unless 
it's too small compared to variance estimation of the other treatment. All the above intuitions cannot be reflected in previous works using the difference in mean estimator because, as we discussed before, they don't consider the propensity score and only aim at optimizing the relative ratio of the treatment and control group. The low switching structure of the experiment is necessary to bound the model estimation error, which will be proven to be $0$ later.

Easy as it may seem, the low switching adaptive experiment gives the first non-asymptotic statistical guarantee for adaptive experiment and achieves the best possible estimation accuracy. All the proofs are delayed in the appendix, and we will only show the key ideas and results for simplicity.
The following lemma shows that the ``incomplete randomization" guarantee the cross term $\mathbb{E} \left( (\widehat{\tau}_1- \widehat{\tau}_2)(\widehat{\tau}_2-\tau) \right)=0$ using the key observation that under our ``incomplete randomization" design, $\hat{\tau}_1-\hat{\tau}_2=0$. In other words under our design, there is no need for ``model estimation", and the AIPW estimator is reduced to the IPW estimator. This is actually a very specific property under the incomplete randomization design. We will use different techniques and analyses to get rid of the cross term when there are covariates in the experiment in section \ref{sec:hetero-effect}. 
\begin{lemma}\label{lem-crossterm-nofeature}
    Under the low switching adaptive experiment \ref{alg:ADM data collecting}, the cross term $\mathbb{E} \left( (\widehat{\tau}_1- \widehat{\tau}_2)(\widehat{\tau}_2-\tau) \right)=0$.
\end{lemma}

\begin{remark}
   In experimental design literature, given the propensity score $e$, more often each $W_i$ is sampled independently as a Bernoulli distribution with expectation $e$. However, in this case using a Cauchy-Schwarz type argument on the cross term can only give us the bound $\mathbb{E} \left( (\widehat{\tau}_1- \widehat{\tau}_2)(\widehat{\tau}_2-\tau) \right) \le O(n^{-1/2} V^{*})$, which will be a much larger term compared to model estimation error as well as the regret in propensity score optimization, and lead to a sub-optimal experiment. That's why we use the ``incomplete randomization" here since it can guarantee that the cross term is $0$. In the next section, we will use another technique, cross-fitting, to get rid of the cross term.
\end{remark}

The benefit of incomplete randomization is that it not only ensures the cross term to be $0$ but also rules out the model estimation term. Furthermore, the AIPW estimator $\widehat{\tau}_1$ in this experiment has an equivalent form that is much more intuitive. All the results are summarized in the following theorem, which is the highlight of this section.

\begin{theorem}\label{thm-upperbound-nofeature}
    Under the low switching adaptive experiment \ref{alg:ADM data collecting} for n larger than some universal constant, the model estimation error $\mathbb{E} \left( \widehat{\tau}_1- \widehat{\tau}_2\right)^2=0$, and the regret of propensity score optimization
    $ \mathbb{E}\left(\widehat{\tau}_2- \tau \right)^2-\mathbb{E} \left( \widehat{\tau}^*- \tau \right)^2= \widetilde{O}(n^{-1+\frac{1}{K}})V^*$. As a result, we have    $\E(\hat{\tau}_1)=\tau$, which is an unbiased estimator and
     $ \mathbf{Var}\left(\widehat{\tau}_1- \tau \right)^2\le  (1+\widetilde{O}(n^{-1+\frac{1}{K}})) V^{*} $. Moreover, the AIPW estimator $\widehat{\tau}_1$ in (\ref{aipw-estimator}) is equivalent to a weighted average of mean difference estimator using the i.i.d data collected in each batch. Specifically
     \begin{align}\label{equ-adm}
&\widehat{\tau}_{1}=\sum_{k=1}^K \frac{\gamma_k}{n} \widehat{\tau}_{k, D M}
\end{align}
where $\widehat{\tau}_{k, D M}$ is the difference in mean estimator(DM) only using data in $k$-th batch. Namely
\begin{align}
&\widehat{\tau}_{k, D M}:=\frac{\sum_{i=\Gamma_{k-1}+1}^{\Gamma_k} W_i Y_i}{\sum_{i=\Gamma_{k-1}+1}^{\Gamma_k} W_i}-\frac{\sum_{i=\Gamma_{k-1}+1}^{\Gamma_k}\left(1-W_i\right) Y_i}{\sum_{i=\Gamma_{k-1}+1}^{\Gamma_k} 1-W_i}
\end{align}

\end{theorem}

Theorem \ref{thm-upperbound-nofeature} validates the performance of low switching adaptive experiment. Also, while we start from the idea of doubly robust method, eventually it is reduced back to a weighted difference in mean estimator. The difference in mean estimator within each batch is the average of i.i.d. data with statistical guarantee, and we weight them according to the batch length $\gamma_k$. The estimation error in initial batches 
might be large, but their weight in the final estimation $\gamma_k/n$ is also pretty small. As we approach the optimal design, the estimation error in the final batches will be near-optimal, and they take the majority weight in the estimator, which also matches the intuition of low switching adaptive experiment.
%As a byproduct, in fact using the probabilistic method developed in this paper, we give a positive answer to the open question in \cite{zhao2023adaptive}: 

%\textit{is there some statistical guarantee for the mean squared error of adaptive Neyman allocation proposed in (\cite{zhao2023adaptive})?}

%\begin{corollary}
  %  The adaptive Neyman allocation algorithm proposed in (\cite{zhao2023adaptive}), together with the mean estimator $\widehat{\tau}$, has the following mean square error guarantee:    
%\end{corollary}

At the end of this section, we provide a novel lower bound of estimation accuracy for any regularized experiment and a wide range of natural estimators. Since the data generation process can be arbitrarily adaptive, it's not possible to adopt classical statistical lower bounds like the Cramer-Rao bound, but instead, we use the Bayes risk argument and information lower bound to prove the optimality of our low switching adaptive experiment.

\begin{theorem}\label{thm-lowerbound-nofeature}
    For any non-anticipating allocation mechanism and data generation process with any estimator $\widehat{\tau}$, there exists an outcome distribution $ \mathcal{D} $ with corresponding parameters $ \mu^{(1)}, \mu^{(0)}, \sigma^{(1)}, \sigma^{(0)} $, such that 
   \[
      \mathbb{E}[(\widehat{\tau} - \tau)^2] \geq \frac{(\sigma^{(1)} + \sigma^{(0)})^2}{n} + O(\frac{(\sigma^{(1)} + \sigma^{(0)})^2}{n^2})
   \] 
\end{theorem}
%Proof is deferred to Appendix \ref{}.
Theorem \ref{thm-lowerbound-nofeature} establishes the fundamental statistical limits for parameter estimation of any adaptive experiment, thus providing a generalization of Cramer-Rao lower bound for possibly non-iid data. It is first argued that the minimax risk for worst case instance $(\mu^{(1)},(\sigma^{(1)})^2)$, $(\mu^{(0)},(\sigma^{(0)})^2)$ is lower bounded by the Bayes risk if we generate $(\mu^{(1)},(\sigma^{(1)})^2)$, $(\mu^{(0)},(\sigma^{(0)})^2)$ according to some prior distribution $\pi$. Then an information lower bound argument is adopted to show that for sufficiently close $(\sigma^{(1)})^2$ and $(\sigma^{(0)})^2$, with constant probability any experiment cannot distinguish them, which leads to sub-optimal allocation mechanism. 

\begin{remark}
    This lower bound is different from the Cramer-Rao lower bound. On the one hand, it is stronger as it also captures dependence on lower-order terms. On the other hand, the Cramer-Rao lower bound applies universally for any instance of $\sigma^{(1)}$, $\sigma^{(0)}$, $\mu^{(1)}$, $\mu^{(0)}$ provided that the estimator is unbiased, whereas the lower bound in  \cref{thm-lowerbound-nofeature} only holds for some specific instance. This raises the question of whether it is possible to derive a Cramer-Rao type lower bound that uniformly applies across all instances. The answer is affirmative, and, to avoid redundancy, this is postponed to the next section, specifically in \cref{cor-lowerbound-Neyman}, after we establish a more general lower bound that incorporates covariates. The arguments and guarantees in \cref{thm-lowerbound-nofeature} and \cref{cor-lowerbound-Neyman} are fundamentally different, and both deviate from existing approaches that apply only in the i.i.d. setting, thus offering new insights into proving the statistical limits of adaptively collected data. 
\end{remark}

\section{Optimal Adaptive Experiment for Heterogeneous Treatment Effect}
\label{sec:hetero-effect}
In this section, we will extend the low switching adaptive experimental design to the scenarios where subjects with different covariates will have heterogeneous outcomes.
Throughout this section, we will assume that 
the expectation function $|\mu^{(i)}(x)| \le  M$ for any $x \in \mathcal{X}$ and $i\in\{0,1\}$. Both mean functions and variance functions are $L-$Lipschitz for some universal constant $L$. 
We will also assume that $\sigma(x) \ge \sigma$ for any $x\in \mathcal{X}$ and some positive universal constant $\sigma$ to ensure that there is minimum uncertainty for any covariate $x$ in the compact set $\mathcal{X}$.
\begin{remark}
    The assumption of Lipschitz continuous condition for functions in compact space is the most basic and widely adopted in statistical learning (\cite{chen2021nonparametric}, \cite{hu2020smooth}). But in the meanwhile, if they possess higher order smoothness or other structures, for example, the function class $\mathcal{F}_{\mu}$ consists of all second-order smooth functions or $L$-Lipschitz convex functions, the rate of regression oracle that we define later can be different (\cite{wainwright_2019}). The assumption of minimum constant uncertainty is needed as a regularity condition for theoretical analysis, which is very similar to the assumption that the density function $p(x) \ge \sigma$ of distribution $\mathcal{D}_x$ for some constant $\sigma$ in all the density estimation or general non-parametric regression literature (\cite{cai2009variance}, \cite{shen2020optimal}). Also in practice, it's not reasonable to allow the uncertainty level of outcome to be arbitrarily close to $0$, due to unavoidable measurement error. But again, we don't assume that there is an upper bound of the outcome variance, and the propensity score can also be extremely unbalanced.
\end{remark}

In this setting, for a non-adaptive experiment with fixed propensity score function $e(X)$, the variance lower bound of any unbiased estimator $\widehat{\tau}$, also known as semi-parametric efficiency bound (\cite{newey1990semiparametric}), will be
$$
\E \left(\widehat{\tau}-\tau\right)^2\ge \frac{1}{n}\left(\mathbb{E}_X 
\left( \frac{(\sigma^{(1)}(X))^2}{e(X)}+\frac{(\sigma^{(0)}(X))^2}{1-e(X)}\right)+ 
 \mathbb{E}_X \left( \tau(X)-\tau\right)^2\right). $$
And the doubly robust estimators like AIPW are known to asymptotically achieve this efficiency bound under certain regularity conditions. 
 Therefore, a clairvoyant knowing the variance function $\sigma^{(1)}(X)$ and $\sigma^{(0)}(X)$ in advance would set the propensity score $e^{*}(X)=\sigma^{(1)}(X)/(\sigma^{(1)}(X)+\sigma^{(0)}(X))$ to achieve the \textit{generalized Neyman allocation bound} (\cite{kato2020efficient}). 
 \begin{equation}\label{optimal-semi-efficiency}
     V^*= \frac{1}{n}\left(\mathbb{E}_X \left( \sigma^{(1)}(X)+\sigma^{(0)}(X)\right)^2+ 
 \mathbb{E}_X \left( \tau(X)-\tau\right)^2\right).
 \end{equation}

 Since the experimenter cannot know the variance function in advance and must learn through the experiment process, a natural question will be:

 \textit{Is the generalized Neyman allocation bound a valid lower bound for any non-anticipating adaptive experiment? If so, can we design an adaptive experiment to achieve such a lower bound?}
 
So far, most studies focus on observational data, and a non-asymptotic convergence rate to the semi-parametric efficiency bound hasn't been discussed much. Using the same idea as introduced in the last section, we adopt the doubly robust method in adaptive experiment and divide the estimation error into two parts: model estimation and propensity score optimization. Then we show that the propensity score optimization is naturally a contextual bandit optimization problem, and a low switching algorithm is desired for statistical guarantee in the model estimation.
Combining everything together, we propose a low-switching adaptive experiment with covariates, which is basically a low switching contextual bandit algorithm.  

\subsection{Low switching Adaptive Experiment to Achieve Generalized Neyman Allocation}\label{subsec:lowswitch-covariate}
We will use the AIPW estimator to estimate the treatment effect, which is defined as
\begin{equation}\label{equ-aipw}
    \begin{aligned}          
      \widehat{\tau}_1^X =&
    \sum_{i=1}^n  \frac{1}{n} \left(\widehat{\mu}^{ (1)}(X_i)-\widehat{\mu}^{ (0)}(X_i)+ \frac{Y_i-\widehat{\mu}^{ (1)}(X_i)}{e_i(X_i)}W_i -\frac{Y_i-\widehat{\mu}^{ (0)}(X_i)}{1-e_i(X_i)}(1-W_i) \right),
    \end{aligned}
\end{equation}
where we use the superscript $X$ to emphasize the existence of covariates, and $\widehat{\mu}^{(1)}(X)$, $\widehat{\mu}^{(0)}(X)$ as the estimation for outcome function of treatment and control $\mu^{(1)}(X)$, $\mu^{(0)}(X)$.
Similarly, we can define the intermediary and optimal estimator as 
\begin{equation}
    \begin{aligned}
       \widehat{\tau}_2^X&=
        \sum_{i=1}^n \frac{1}{n} \left(\mu^{(1)}(X_i)-\mu^{(0)}(X_i)+ \frac{Y_i-\mu^{(1)}(X_i)}{e_i(X_i)}W_i -\frac{Y_i-\mu^{(0)}(X_i)}{1-e_i(X_i)}(1-W_i) \right),
        \end{aligned}
\end{equation}
\begin{equation}\label{equ-def-optimaltauX}
    \begin{aligned}
         \widehat{\tau}^{*X}&=
        \sum_{i=1}^n \frac{1}{n} \left(\mu^{(1)}(X_i)-\mu^{(0)}(X_i)+ \frac{Y_i-\mu^{(1)}(X_i)}{e^*(X_i)}W_i -\frac{Y_i-\mu^{(0)}(X_i)}{1-e^*(X_i)}(1-W_i) \right),
    \end{aligned}
\end{equation}
where the optimal propensity score $e^*(X)=\sigma^{(1)}(X)/(\sigma^{(1)}(X)+\sigma^{(0)}(X))$ and the  estimator $\widehat{\tau}^{*X}$ achieves the  generalized Neyman allocation bound
$V^{*}$ in (\ref{optimal-semi-efficiency}).
Similar to what we have in section \ref{sec:nofeature}, we can divide the mean square error of the AIPW estimator into three parts: model estimation, propensity score optimization, and cross term:
\begin{equation}
    \begin{aligned}
         \mathbb{E} \left( \widehat{\tau}_1^X- \tau \right)^2 -  \mathbb{E} \left( \widehat{\tau}^{*X}- \tau \right)^2&=  \mathbb{E} \left( \widehat{\tau}_1^X- \widehat{\tau}_2^X+\widehat{\tau}_2^X- \tau \right)^2- \mathbb{E} \left( \widehat{\tau}^{*X}- \tau \right)^2 \\
        &= \underbrace{\mathbb{E} \left( \widehat{\tau}_1^X- \widehat{\tau}_2^X\right)^2}_{\text{model estimation}} + \underbrace{\mathbb{E}\left(\widehat{\tau}_2^X- \tau \right)^2-\mathbb{E} \left( \widehat{\tau}^{*X}- \tau \right)^2}_{\text{propensity score optimization}}\\
        & \qquad +\underbrace{2\mathbb{E} \left( (\widehat{\tau}_1^X- \widehat{\tau}_2^X)(\widehat{\tau}_2^X-\tau) \right)}_{\text{cross-term}}
    \end{aligned}
\end{equation}
Also like in section \ref{sec:nofeature}, we assume for now that the cross term is $0$ (which is non-trivial, see Lemma \ref{lem:cross term is 0}), and focus on the model estimation and propensity score optimization separately. The propensity score optimization is a contextual bandit learning problem:
\begin{lemma}\label{lem:propensityscore-feature}
    \begin{equation}
    \mathbb{E}\left(\widehat{\tau}_2^X- \tau \right)^2-\mathbb{E} \left( \widehat{\tau}^{*X}- \tau \right)^2= 
             \frac{1}{n^2} \mathbb{E}\left(\sum_{i=1}^n \frac{(\sigma^{(1)}(X_i))^2}{e_i(X_i)} +\sum_{i=1}^n
             \frac{(\sigma^{(0)}(X_i))^2}{1-e_i(X_i)}
             - n \left(\frac{(\sigma^{(1)}(X_i))^2}{e^*(X_i)} +\frac{(\sigma^{(0)}(X_i))^2}{1-e^*(X_i)}\right) \right).
\end{equation}
\end{lemma}
Lemma \ref{lem:propensityscore-feature} tells us that the propensity score optimization is equivalent to learning the variance function $(\sigma^{(1)}(X),\sigma^{(0)}(X))$ and optimize the policy $e_i(X)$, and the regret is measured against the optimal policy $e^{*}(X)$. 
On the other hand, unlike the homogeneous case, here the model estimation error will not be $0$, 
and need to be dealt with with care:
\begin{equation}
\begin{aligned}
    \label{eq: model estimation}
     \mathbb{E} \left( \widehat{\tau}_1^X- \widehat{\tau}_2^X\right)^2
     = &\mathbb{E} \left(\frac{1}{n} \sum_{i=1}^n \left(1-\frac{W_i}{e_i(X_i)}\right) \left(\widehat{\mu}^{ (1)} (X_i)-\mu^{(1)}(X_i)\right) -\frac{1}{n}\sum_{i=1}^n \left(1-\frac{1-W_i}{1-e_i(X_i)}\right) \left(\widehat{\mu}^{ (0)}(X_i)-\mu^{(0)}(X_i)\right) \right) .
\end{aligned}
\end{equation}
This model estimation term is hard to analyze if the propensity score $e_i(X)$ can be arbitrarily correlated. Also we need to be careful about the correlation between $e_i(X), W_i$ and data used to estimate $\widehat{\mu}^{(1)}(X)$, $\widehat{\mu}^{(0)}(X)$. To have a clean analysis of model estimation error, we propose the \textit{low switching adaptive experiment with covariates} framework. First, we need some assumptions of access to offline (non-parametric) regression oracle, which have been extensively studied.
\begin{assumption}
    \label{asm:oracle with convariate}
   Let $ (X_1, Y_1), \cdots, (X_m, Y_m) $ be  a batch of i.i.d data such that $ X \sim \mathcal{D}_{\mathcal{X}}, \mathbb{E}[Y | X = x] = \mu(x), \Var[Y | X = x] = \sigma^2(x)$. We assume that the size of each batch $m \ge C_3 \log n$ for sufficiently large constant $C_3$, such that
there exists a regression oracle which takes $ (X_1, Y_1), \cdots, (X_m, Y_m) $ as input and outputs regression functions $ \widehat{\mu} : \mathcal{X} \rightarrow \mathbb{R} $, $ \widehat{\sigma} : \mathcal{X} \rightarrow \mathbb{R}_{\geq 0}  $ for expectation and standard deviation function $\mu(x)$, $ \sigma(x) $ such that with probability $ 1 - \delta $ for some small $\delta= \frac{1}{n^4}$ and universal constant $C_4$, $C_5$, 
\begin{align}
\mathbb{E}_{x \sim \mathcal{D}_{\mathcal{X}} }[(\mu(x) - \widehat{\mu}(x))^2] \leq C_4 \frac{\max_{x \in \mathcal{X}}(\sigma(x))^2}{m^{\alpha}}\log (\frac{1}{\delta}),\\
     \mathbb{E}_{x \sim \mathcal{D}_{\mathcal{X}} }[(\sigma(x) - \widehat{\sigma}(x))^2] \leq C_5 \frac{\max_{x \in \mathcal{X}}(\sigma(x))^2}{m^{\beta}}\log (\frac{1}{\delta})
\end{align}
   where $ \alpha, \beta $ are the non-parametric regression rates depending on smoothness of $ \mu(x) $ and $ \sigma(x) $. 
\end{assumption}

\begin{remark}
    $\alpha$ and $\beta$ are the non-parametric regression rates which depend on the complexity of function class (\cite{wainwright_2019},\cite{shen2020optimal}) and should satisfy $0 < \alpha, \beta \leq 1$. The unusual term $\max_{x \in \mathcal{X}}(\sigma(x))^2$ exists as the variance $(\sigma(X))^2$ may not be upper bounded and can depend on $n$, so all the concentration bounds should also scale with $(\sigma(X))^2$.
\end{remark}

Our algorithm still runs in batches.
We still denote the length of the $ k$-th batch as $ \gamma_k $ and let $ \Gamma_k := \sum_{j=1}^{k} \gamma_k  $ be the total length of first $ k $ batches. Denote $ S_k := \sum_{i=0}^{k-1} \beta^i = \frac{1-\beta^k}{1-\beta}$ for $k\ge 2$ and by default set $S_1=1$. The details of our experiment are stated in algorithm \Cref{alg:AIPW data collecting}.

\begin{algorithm}[h]
   \label{alg:AIPW data collecting}
   \caption{Low Switching Adaptive Experiment with Covariates}
         \textbf{Input:} the number of periods $ n $, the number of batch $ K $.\\
      \caption{Low Switching Adaptive Experiment with Covariates}
         \textbf{Input:} the number of periods $ n $, the number of batch $ K $.\\
   Denote $\widehat{\sigma}_0^{(0)}(x) = \widehat{\sigma}_0^{(1)}(x) = 0 $ and $\frac{0}{0+0} = \frac{1}{2}$\\
   For any $ k \in [K] $, set the batch length
   $\gamma_k=(n / a)^{\frac{S_k}{S_K}}(\log n)^{1 - \frac{S_k}{S_K}}$ for $ a \in \mathbb{R}_{>0}$ such that $ \sum_{k=1}^K \gamma_k = n $. \\
   Since $ \sum_{k=1}^K n^{\frac{S_k}{S_K}}(\log n)^{1 - \frac{S_k}{S_K}} > n^{\frac{S_K}{S_K}}(\log n)^{1 - \frac{S_K}{S_K}} = n $
   and $ \sum_{k=1}^K (n/K)^{\frac{S_k}{S_K}}(\log n)^{1 - \frac{S_k}{S_K}} < K\frac{n}{K} = n$, we conclude that $ 1 < a < K $. \\
   \ForEach{$k \in [K]$}{
   Let $\widetilde{\sigma}_{k-1}^{(1)}(x) = \max\{\widehat{\sigma}_{k-1}^{(1)}(x), \frac{\gamma_1}{\gamma_k}\widehat{\sigma}_{k-1}^{(0)}(x)$ \}, \\
   Let $\widetilde{\sigma}_{k-1}^{(0)}(x) = \max\{\widehat{\sigma}_{k-1}^{(0)}(x), \frac{\gamma_1}{\gamma_k}\widehat{\sigma}_{k-1}^{(1)}(x)$ \}. \\
   \ForEach{$ \Gamma_{k-1} + 1 \leq i \leq \Gamma_{k} $ }{
   \If{$ i \in G_1 $ }{
      Set propensity score function as 
      $ \forall x \in \mathcal{X}, e_i(x) = \frac{\widetilde{\sigma}_{k-1}^{G_1, (1)}(x)}{\widetilde{\sigma}_{k-1}^{G_1, (0)}(x) + \widetilde{\sigma}_{k-1}^{G_1, (1)}(x)} 
      = \min\{ \frac{1}{1+\gamma_1/\gamma_k }, \max\{\frac{\gamma_1/\gamma_k }{1+\gamma_1/\gamma_k }, \frac{\widehat{\sigma}_{k-1}^{G_1, (1)} }{\widehat{\sigma}_{k-1}^{G_1, (0)} + \widehat{\sigma}_{k-1}^{G_1, (1)} }\}\}$ \\
   }
   \If{$ i \in G_2 $ }{
      Set propensity score function as $ \forall x \in \mathcal{X}, e_i(x) = \frac{\widetilde{\sigma}_{k-1}^{G_2, (1)}(x)}{\widetilde{\sigma}_{k-1}^{G_2, (0)}(x) + \widetilde{\sigma}_{k-1}^{G_2, (1)}(x)}
      = \min\{ \frac{1}{1+\gamma_1/\gamma_k }, \max\{\frac{\gamma_1/\gamma_k }{1+\gamma_1/\gamma_k }, \frac{\widehat{\sigma}_{k-1}^{G_2, (1)} }{\widehat{\sigma}_{k-1}^{G_2, (0)} + \widehat{\sigma}_{k-1}^{G_2, (1)} }\}\}$ \\
   }
      Observe $ X_i $. \\
      With probability $ e_i(X_i) $, let $ W_i = 1$. Otherwise let $ W_i = 0 $.
   }
   For arm $ w \in \{0, 1\} $, use group1 outcomes in this batch $ \{(X_i, Y_i) : \Gamma_{k-1} + 1 \leq i \leq \Gamma_{k}, W_i = w, i \in G_1\} $ and variance oracle to calculate $ \widehat{\sigma}_{k}^{G_1, (w)} : \mathcal{X} \rightarrow \mathbb{R}_{\geq 0}  $. \\
   For arm $ w \in \{0, 1\} $, use group2 outcomes in this batch $ \{(X_i, Y_i) : \Gamma_{k-1} + 1 \leq i \leq \Gamma_{k}, W_i = w, i \in G_2\} $ and variance oracle to calculate $ \widehat{\sigma}_{k}^{G_2, (w)} : \mathcal{X} \rightarrow \mathbb{R}_{\geq 0}  $. \\
   }
\end{algorithm}

Generally speaking, experiment \Cref{alg:AIPW data collecting} is a \textit{offline oracle based} low switching contextual bandit algorithm that has been extensively studied for regret minimization, but also surprisingly has the power to ensure \textit{near optimal estimation accuracy}. Again we are setting a threshold for the minimum level of exploration for every $x$ in every batch. Some other tricks like data splitting and cross-fitting are adopted to 
ensure the independence of dataset used for estimating $\mu^{(1)}(X)$ and $\mu^{(0)}(X)$. Intuitively speaking, 
data splitting divides the $n$ units into two parallel sets $G_1$ and $G_2$, and two experiments are conducted on these two sets independently. Then the model estimation from $G_1$ is used to optimize experimental design in $G_2$ and vice versa. In this way, the tasks of model estimation and propensity score estimation are separated and can be regarded as independent, simplifying the analysis. In particular, now we have a slightly different AIPW estimator $\hat{\tau}_1^X$ with cross-fitting:
\begin{equation}
    \begin{aligned}          
      \widehat{\tau}_1^X =&
     \frac{1}{n} \sum_{i \in G_1} \left(\widehat{\mu}^{G_2, (1)}(X_i)-\widehat{\mu}^{G_2, (0)}(X_i)+ \frac{Y_i-\widehat{\mu}^{G_2, (1)}(X_i)}{e_i(X_i)}W_i -\frac{Y_i-\widehat{\mu}^{G_2, (0)}(X_i)}{1-e_i(X_i)}(1-W_i) \right) \\
   &+ \frac{1}{n} \sum_{i \in G_2} \left(\widehat{\mu}^{G_1, (1)}(X_i)-\widehat{\mu}^{G_1, (0)}(X_i)+ \frac{Y_i-\widehat{\mu}^{G_1, (1)}(X_i)}{e_i(X_i)}W_i -\frac{Y_i-\widehat{\mu}^{G_1, (0)}(X_i)}{1-e_i(X_i)}(1-W_i) \right). 
    \end{aligned}
\end{equation}
cross-fitting validates that AIPW estimator $\widehat{\tau}_1^X$ is unbiased, and the cross term is $0$, as claimed in the following two lemmas.
\begin{lemma}
    \label{lem:AIPW unbiasness}
    Under the low switching adaptive experiment with covariates, AIPW estimator $\mathbb{E} (\widehat{\tau}_1^X)=\tau$.
\end{lemma}
\begin{lemma}
    \label{lem:cross term is 0}
    Under the low switching adaptive experiment with covariates, the cross term
    $\mathbb{E} \left( (\widehat{\tau}_1^X- \widehat{\tau}_2^X)(\widehat{\tau}_2^X-\tau) \right)=0$.    
\end{lemma}
Making sure that the cross term equals $0$ is important since otherwise a Cauchy-Schwarz type argument would only give us $\mathbb{E} \left( (\widehat{\tau}_1^X- \widehat{\tau}_2^X)(\widehat{\tau}_2^X-\tau) \right) \le O(n^{-\alpha/2}) V^{*}$, which typically dominates both the model estimation and propensity score optimization. In other words, cross-fitting allows us to achieve ``\textbf{faster rate}" in approaching the semiparametric efficiency bound.
Moreover, the cross-fitting techniques also greatly simplify the analysis of model estimation error, given by the next inequality.
\begin{equation}\label{eq:model estimation, Cauchy}
    \begin{aligned}
          \mathbb{E} \left( \widehat{\tau}_1^X- \widehat{\tau}_2^X\right)^2
     \le& \frac{4}{n^2}
     \left(\sum_{i \in G_1}\mathbb{E}\left( \left(1-\frac{W_i}{e_i(X_i)}\right) \left(\widehat{\mu}^{G_2, (1)} (X_i)-\mu^{(1)}(X_i)\right)\right)^2\right.\\
     &+\left. \sum_{i \in G_2}\mathbb{E}\left( \left(1-\frac{W_i}{e_i(X_i)}\right) \left(\widehat{\mu}^{G_1, (1)} (X_i)-\mu^{(1)}(X_i)\right)\right)^2\right.\\
     &+ \left. \sum_{i \in G_1}\mathbb{E}\left( \left(1-\frac{1-W_i}{1-e_i(X_i)}\right) \left(\widehat{\mu}^{G_2, (0)}(X_i)-\mu^{(0)}(X_i)\right)\right)^2\right.\\
     &+ \left. \sum_{i \in G_2}\mathbb{E}\left( \left(1-\frac{1-W_i}{1-e_i(X_i)}\right) \left(\widehat{\mu}^{G_1, (0)}(X_i)-\mu^{(0)}(X_i)\right)\right)^2\right).
    \end{aligned}
\end{equation}
\noindent The power of inequality (\ref{eq:model estimation, Cauchy})
is that while the data generating process and allocation decision $W_i$ can be adaptive and very complicated, the martingale structure with cross-fitting enables us to separate all the correlations and bound the ``weighted" model estimation error for each $X_i$.
As a consequence, we have
\begin{lemma}
    \label{lem:model estimation}
        There exist a constant $ C $ depend on $ K, \alpha, \beta $, the diameter of $ \mathcal{X} $, the lipschitz constant $ L $ for $ \sigma^{(1)}(x), \sigma^{(0)}(x) $ and the lower bound $ \sigma $  for $ \sigma^{(1)}(x), \sigma^{(0)}(x) $ such that 
    for any $ n > C $,
    \begin{equation}
         \mathbb{E} \left( \widehat{\tau}_1^X- \widehat{\tau}_2^X\right)^2
     \le O\left(\frac{\log n}{n^\alpha} \right) V^*
    \end{equation}
\end{lemma}
 A standard argument of \textit{offline oracle based} contextual bandit algorithm is adopted to bound the regret of propensity score optimization:
 \begin{lemma}
 \label{lem:propensity optimization regret-feature}
     There exist a constant $ C $ depend on $ K, \alpha, \beta$, the diameter of $ \mathcal{X} $, the Lipschitz constant $ L $ for $ \sigma^{(1)}(x), \sigma^{(0)}(x) $ and the lower bound $ \sigma $  for $ \sigma^{(1)}(x), \sigma^{(0)}(x) $ such that 
    for any $ n > C $,
 \begin{equation}
        \mathbb{E}\left(\widehat{\tau}_2^X- \tau \right)^2-\mathbb{E} \left( \widehat{\tau}^{*X}- \tau \right)^2 \le O\left( K n^{-1 + \frac{1}{S_K}}(\log n)^{2 - \beta - \frac{1}{S_K}}\right)V^*
 \end{equation}    
 \end{lemma}
Putting everything together, we characterize the performance of low switching adaptive experiment with covariates in the following theorem.
\begin{theorem}
    The AIPW estimator $\widehat{\tau}_1^X$ satisfies that $\mathbb{E}\left(\widehat{\tau}_1^X\right)=\tau$ and $$\mathbb{E} \left( \widehat{\tau}_1^X- \tau \right)^2 =\Var \left(\widehat{\tau}_1^X\right) \le (1+ \widetilde{O}(n^{-\alpha}) + \widetilde{O}(Kn^{-1 + \frac{1}{S_K}})) V^*.$$
Specifically, when $K= O(\log n)$, we have 
$\Var (\widehat{\tau}_1^X)= (1+\widetilde{O}(n^{-\min(\alpha,\beta)}))V^*$.
\end{theorem}

The following theorem shows the fundamental statistical limit of estimating ATE with heterogeneous covariates, and can be understood as a generalization of classical semi-parametric efficiency bound in the sense that it not only holds over all possible estimators, but also among all (possibly adaptive) designs.

\begin{theorem}\label{thm-lowerbound-GNeyman}
    For any non-anticipating allocation mechanism and data generating process with any unbiased estimator $\hat{\tau}$, assume that the function classes $\mathcal{F}_{\mu}$ and $\mathcal{F}_{\sigma}$ that is an open set under infinite norm $\| \quad \|_{\infty}$, then for  any $\mu^{(1)}(X), \mu^{(0)}(X) \in \mathcal{F}_{\mu}$ and $\sigma^{(1)}(X), \sigma^{(0)}(X) \in \mathcal{F}_{\sigma}$, we have 
\begin{equation}
    \Var (\hat{\tau}) \ge \frac{1}{n}\left(\mathbb{E}_X \left( \sigma^{(1)}(X)+\sigma^{(0)}(X)\right)^2+ 
 \mathbb{E}_X \left( \tau(X)-\tau\right)^2\right)=V^{*},    
\end{equation}
where $\tau(X)=\mu_1(X)-\mu_0(X)$ is the conditional average treatment effect. Furthermore, the equality is achieved if and only if the allocation mechanism is the generalized Neyman allocation $e^{*}(X)$ and the estimator is the idealized AIPW estimator $\hat{\tau}^{*X}$ as defined in
(\ref{equ-def-optimaltauX}).
\end{theorem}
Theorem \ref{thm-lowerbound-GNeyman} shows that even in the broader class of adaptive experiments and unbiased estimators, the AIPW estimator with (non-adaptive) generalized
Neyman allocation is still universally optimal.
As a direct corollary, we show that Neyman allocation gives the estimation lower bound for any unbiased estimator and any (possibly adaptive) data collecting process without covariates.

\begin{corollary}\label{cor-lowerbound-Neyman}
      For any non-anticipating allocation mechanism and data generating process with any unbiased estimator $\hat{\tau}$, we have 
      \begin{equation}
          \Var(\hat{\tau})
          \ge \frac{(\sigma^{(1)}+\sigma^{(0)})^2}{n}= V^{*}
      \end{equation}
      for any $\mu^{(1)}$, $\mu^{(0)}$, $\sigma^{(1)}>0$, $\sigma^{(0)}>0$.
\end{corollary}

While we believe that this convergence rate of $(1+\widetilde{O}(n^{-\min(\alpha,\beta)}))$ is optimal, like the special case when $\alpha=\beta=1$ in section \ref{sec:nofeature}, we don't have a matching lower bound for it and leave this as an open question for future research.

To construct valid confidence interval, a central limit theorem of our design can be directly shown by checking the sufficient conditions in Theorem 1 in (\cite{cook2024semiparametric}):
\begin{theorem}\label{thm-CLT}
    Assume that $\sigmaone(X)$, $\sigmazero(X)$ are both upper and lower bounded by constants independent of $n$ for all $X$, then 
    our allocation algorithms (\ref{alg:ADM data collecting}) and \cref{alg:AIPW data collecting} together with our estimators $\hat{\tau}_1$ in \cref{equ-adm} and $\hat{\tau}_1^{X}$ in \cref{equ-aipw} satisfy
    \begin{equation}
        \begin{aligned}
         \sqrt{n} (\hat{\tau}_1-\tau)   &\xrightarrow{d} \mathcal{N}(0, (\sigmaone+\sigmazero)^2),\\
          \sqrt{n} (\hat{\tau}_1^X-\tau)   &\xrightarrow{d} \mathcal{N}(0, \mathbb{E}_X \left( \sigma^{(1)}(X)+\sigma^{(0)}(X)\right)^2+ 
 \mathbb{E}_X \left( \tau(X)-\tau\right)^2
          ).
        \end{aligned}
    \end{equation}
\end{theorem}

\subsection{Comparison of Doubly Robust Method in Experimental Design and Observational Study}\label{subsec:comparison}
 At this point, while we have stated our main mathematical results, it's still pretty important to compare the doubly robust method between adaptive experiment and observational study, and emphasize why it's a particular fit for adaptive experiment. 

 1. In experimental design, the adaptive experiment and the imaginary optimal experiment have different propensity score, and the data are generated independently according to their sampling scheme. In other words, the AIPW estimator $\widehat{\tau}_1^X$ is constructed based on one dataset $\{X_i,e_i(X_i),W_i,Y_i\}_{i=1}^n$, and the imaginary optimal estimator $\widehat{\tau}^{*X}$ is constructed using another independent dataset $\{X_i^{\prime},e_i^{\prime}(X_i^{\prime}),W_i^{\prime},Y_i^{\prime}\}_{i=1}^n$. While in observational study, all the analysis is based on the same dataset, which introduces additional correlation.  

 2. In observational study, the propensity score estimation tries to estimate the underlying true propensity score, and the name of ``doubly" robust comes from that if either one of the model estimation or propensity score estimation is estimated accurately, the doubly robust estimator will be asymptotically efficient. Therefore, in doubly robust analysis, typically we need some ``risk decay" condition (\cite{kennedy2016semiparametric}) like  $$\mathbb{E}\left(\widehat{\mu}(X)-\mu(X)\right)^2 \mathbb{E}\left(\widehat{e}(X)-e(X)\right)^2 =o (\frac{1}{n}).$$
While in experimental design, instead of estimating propensity score, we are trying to optimize the design, and as we prove above, optimizing the estimation error is equivalent to separately estimating the model and optimizing propensity score. \textbf{A better model estimation will lead to a better propensity score design, which will in turn lead to better model estimation.} So instead of the more conservative ``doubly robust", we think that in adaptive experiment they are actually ``\textbf{mutually reinforcing}".
\subsection{Why We Need Low Switching}
Carefully going through the proof in section \ref{sec:hetero-effect}, we can find that the only place we use low switching structure is adopting an offline oracle to return a good estimation of mean and variance function. In other words, if there is a somehow ``magical" online regression oracle that gives the same statistical guarantee, then is a low switching algorithm still necessary as we claimed? We want to argue this issue in several ways. First of all, a low switching algorithm is much more efficient and practical, which might be the most important thing in practice. Secondly, while there are some existing online oracles for mean estimation of certain function classes, they are much more computationally inefficient and cover only a narrow class of functions (see \cite{simchilevi2021bypassing} for a detailed discussion), and thus far there is no any online oracle for variance estimation with statistical guarantee. In contrast, offline oracles have been widely studied, implemented, and adopted in both academia and industry. 
Therefore, an efficient and robust low switching algorithm with well established offline regression oracle is always preferred. The third reason is that while the analysis of correlation structure can be simplified using only cross-fitting, as the somehow ``magical" inequality (\ref{eq:model estimation, Cauchy}) shows, it highly depends on the specific structure of the AIPW estimator.
In more general settings, the doubly robust method may no longer be valid, but this ``learning to design"
framework could still be a powerful and standard paradigm. In particular, assume that there is an optimal design $D^{*}$ and estimator $\widehat{\tau}^*$ if all the parameters and structures (or say the model) $M^{*}$ are known, and the target of adaptive experiment is to learn the parameters in the experimenting process and then construct a better design to approximate $D^{*}$ and $\widehat{\tau}^*$. 
We will set strictly increasing batch size $\gamma_1, \cdots, \gamma_K$, and within each batch of i.i.d. data, we can construct a model estimation $M_k$ for the design of the next batch $D_{k+1}$, as well as a treatment effect estimation $\widehat{\tau}_k$. The ultimate estimator can be constructed as 
$$\widehat{\tau}=\sum_{k=1}^K \frac{\gamma_k}{n}\widehat{\tau}_k. $$
 A union bound over $K$ estimations could be taken to deal with the correlation with a loss of factor $\text{poly}(K)$, which is $\text{poly}(\log n)$ at most:
\begin{equation}
    \begin{aligned}
    \mathbb{E} \left( \widehat{\tau}-\widehat{\tau}^*\right)^2&=\mathbb{E} \left( \sum_{k=1}^K \frac{n_k}{n}(\widehat{\tau}_k-\widehat{\tau}^*)\right)^2\\
    & \le \text{poly} (K)
   \max_{1 \le k\le K}
    \frac{n_k^2}{n^2}\mathbb{E}(\widehat{\tau}_k-\widehat{\tau}^*)^2.
    \end{aligned}
\end{equation}
For the beginning batches, the design is sub-optimal, and the performance of estimation $\widehat{\tau}_k$ could also be poor, but they also have small weights in the estimation.
As the batch size increases, hopefully we will have model estimation $M_k \to M^{*}$ and the design $D_k \to D^{*}$, hence the estimation $\widehat{\tau}_k$ will approach $\widehat{\tau}^{*}$ in latter batches. Overall, we can strike a balance between increasing batch length, model estimation, and design optimization to 
find a near optimal design $\hat{\tau}$.

\section{Numerical Experiment}
\label{sec:numerical}
In this section, we present the numerical results for our low switching adaptive experiment for both settings discussed in this paper.
\subsection{Experiment without covariates}
In this section,
we provide numerical results for estimating the treatment effect without covariates. We compare our method with the non-adaptive experiment, the adaptive Neyman allocation proposed in \cite{zhao2023adaptive}
and the optimal Neyman allocation experiment which knows the variance in advance.
The outcome of the treatment group is generated from a 
skewed Gaussian distribution 
$Y \sim SN(0,1,10)$ with expectation $\E(Y)=0.79$ and standard error $\sigma(Y)=0.61$.
The outcome of the control group is generated from another 
skewed Gaussian distribution 
$Y \sim SN(0,10,10)$ with expectation $\E(Y)=7.93$ and standard error $\sigma(Y)=6.08$. The length of experiment $n=2000$, and the estimation error is averaged over 1 million simulations. We consider 1 batch (non-adaptive), 2 batches with batch length $\gamma_1=60$, $\gamma_2=1940$ and 3 batches with batch length $\gamma_1=60$, $\gamma_2=240$, $\gamma_3=1700$.
The results are listed in the following tables. NA represents the non-adaptive experiment. LSAE represents the low switching adaptive experiment. ANA represents the adaptive Neyman allocation and ONA represents the optimal Neyman allocation.
\begin{table}[h]
\centering
\begin{tabularx}{0.4\linewidth}{|X|X|X|X|X|}
\hline
 Batch & NA &  LSAE &  ANA &  ONA \\
\hline
 1 & 0.037 & 0.037 & 0.037 & 0.022 \\
 2 & 0.037 & 0.023 & 0.023 & 0.022 \\
 3 & 0.037 & 0.023 & 0.022 & 0.022 \\
\hline
\end{tabularx}
\begin{tabularx}{0.44\linewidth}{|X|X|X|X|}
\hline
  Batch & NA &  LSAE &  ANA  \\
\hline
 1 & -5.35e-05 &  -2.61e-06 & -5.35e-05  \\
 2 & -5.35e-05 & -1.27e-04 & -1.46e-03  \\
 3 & -5.35e-05 & 1.48e-05 & -1.81e-03  \\
\hline
\end{tabularx}
\caption{L2 loss and bias of different experiments without covariates}
\label{table-nofeature}
\end{table}

On the left of table \ref{table-nofeature} we show the $L_2$ loss of different methods. The optimal Neyman allocation shows the best possible estimation accuracy, and both the low switching adaptive experiment and adaptive Neyman allocation achieve this estimation accuracy with only $1$ or $2$ policy updates, significantly outperforming the non-adaptive experiment with a reduction of error for more than $30 \%$.
On the right side of table \ref{table-nofeature}, in terms of bias, the low switching adaptive experiment is more than ten times better than the adaptive Neyman allocation, which is as stable as the non-adaptive experiment. In summary, the low switching adaptive experiment can achieve the best possible estimation accuracy using adaptivity, while still maintaining the desired statistical structure as a non-adaptive experiment.

\subsection{Experiment with covariates}

In this section, we provide numerical results for the low switching adaptive experiment with covariates. We have two features $X_1$ and $X_2$ with a uniform distribution. 
Given $X_1$, the outcome of the treatment group is generated from Gaussian distribution $Y(X_1) \sim N(10,1)$, the outcome of the treatment group is generated from Gaussian distribution $Y(X_1) \sim N(1,100)$. 
Given $X_2$, the outcome of the treatment group is generated from Gaussian distribution $Y(X_2) \sim N(1,100)$, and the outcome of the treatment group is generated from Gaussian distribution $Y(X_2) \sim N(10,1)$. Since so far there are few methods applicable for optimizing estimation accuracy with features, we compare our method with optimal Neyman allocation ignoring features, and the optimal generalized Neyman allocation.
The length of experiment $n=2000$, and the estimation error is averaged over 1 million simulations. We consider 1 batch (non-adaptive), 2 batches with batch length $\gamma_1=800$, $\gamma_2=1200$ and 3 batches with batch length $\gamma_1=500$, $\gamma_2=700$, $\gamma_3=800$.
The results are listed in the following tables. ONA represents the optimal Neyman allocation ignoring covariates. LSAE represents the low switching adaptive experiment. OGNA represents optimal generalized Neyman allocation.

\begin{table}[h]
\centering
\begin{tabularx}{0.4\linewidth}{|X|X|X|X|}
\hline
 Batch & ONA &  LSAE &  OGNA  \\
\hline
 1 & 0.14 & 0.14 & 0.10  \\
 2 & 0.14 & 0.12 & 0.10  \\
 3 & 0.14 & 0.12 & 0.10  \\
\hline
\end{tabularx}
\begin{tabularx}{0.2\linewidth}{|X|X|X|X|}
\hline
  Batch  &  LSAE   \\
\hline
 1 & -2.6e-04  \\
 2 & -7.3e-04  \\
 3 & -9.5e-04  \\
\hline
\end{tabularx}
\caption{L2 loss and bias of different experiments with covariates}
\label{table-feature}
\end{table}
Again on the left side, we show that our method outperforms the best possible experiment ignoring covariates, thus outperforming most existing methods. It shows that designing covariate-specific experiment can significantly improve estimation accuracy. It doesn't achieve the accuracy of optimal generalized Neyman allocation, which we suspect could be improved using more updates. Also, the bias of the low switching adaptive experiment is negligible.

\section{Discussion of Potential Future Directions}\label{sec:future}
In this paper, we adopt doubly robust method in adaptive experimental design and show that an optimal design is naturally equivalent to a low switching bandit optimization problem. The common issues, methods, or setups that frequently arise in experimental design can be formulated and addressed within this framework with a structured approach and standard research methodologies, which further emphasizes the generality of this adaptive experiment framework. We list a few of them here.

\textbf{1. Bayesian method for adaptive experimental design.}
Bayesian method is an extreme fit to correlated, adaptively collected data, as argued in literature (\cite{greenhill2020bayesian}). However, while Bayesian optimization has achieved widespread success due to its excellent empirical performance, there hasn't been much work on Bayesian adaptive experiments for statistical inference, mainly due to the reason that adaptive Bayesian method typically doesn't have a frequentist statistical guarantee. However, since we have proven that an optimal experimental design is equivalent to a low switching bandit algorithm, it seems likely that some fine-tuned version of the batched Thompson sampling algorithm (\cite{kalkanli2021batched}) can be transformed into an optimal experimental design.
It's interesting to see whether the Bayesian method can enjoy both outstanding empirical performance and optimal theoretical guarantee.

\textbf{2. Transferring knowledge of offline data.}
In many cases, there will be some existing experiment records from previous data, and how to adopt the information from offline data to construct a more efficient experiment has been widely considered (\cite{ledolter2020focus}). How offline data can be used to achieve better online decision making has been extensively studied (\cite{bu2020online},\cite{cheung2024leveraging}) in bandit literature, and in our framework this may also imply how existing data can help to design a more efficient experiment.

\textbf{3. Experimental design with limited resources.}
In practical experimental settings, conducting experiments often incurs significant costs, such as recruiting participants or maintaining the experimental platform. Consequently, there is frequently a fixed budget for resources and funding. To maximize experimental efficiency under this budget, it is crucial to allocate resources to areas where the uncertainty in information is greatest. When all parameters are known, this becomes a problem of resource allocation (\cite{hurwicz1973design}). However, in most cases, the uncertainty of distribution itself is unknown and should be learned adaptively during the experimenting process. This is known as online resource allocation or bandit with knapsack, which has been comprehensively studied in operations research (\cite{wang2022constant},\cite{badanidiyuru2018bandits}). Under our framework, a low-switching optimal online resource allocation algorithm
might be transformed into a resource constrained efficient experiment.

\subsection{Beyond Doubly Robust Method}

In the following, we list some important issues in experimental design that this ``learning to design" framework might be helpful, even if doubly robust methods may no longer be valid.

\textbf{1. Non-stationary outcome or distribution shift.}
In many common cases, when the outcome distribution itself is time varying (\cite{simchi2023non},\cite{glass1972estimating}), a static experimental design could fail to have a reasonable estimation. In contrast, using the techniques developed for detecting and analyzing non-stationarity or distribution shift
in online decision making (\cite{suk2021self},\cite{besbes2014stochastic}), it may be possible for the adaptive experimental design framework to learn about the non-stationarity, accordingly develop a better design and construct valid statistical inference.

\textbf{2. Learning the interference structure. } In many cases, a simple A/B testing might be biased if the treatment allocation of one unit could influence the outcome of the other one, which is known as the interference effect and has been widely considered in recent research (\cite{li2022interference},\cite{farias2022markovian}).
Typically a specific network structure or graphical model is assumed, and how the interference affects the outcome is considered as known. In practice, however, this may not be true (\cite{vansteelandt2012model}), and model misspecification may result in sub-optimal or even invalid estimation.
In our framework, this issue might be addressed by learning the interference structure in the experiment process.
At the beginning when the prior information on the interference effect is unclear, a conservative design and estimation should be adopted, and as the structure is estimated accurately during the process, the design and estimation could be more specific to the structure with better performance.

\textbf{3. Learning the pattern of dynamic treatment effect.}
In some other cases, an experiment for each unit is conducted across a period, and treatment allocations from the past could affect the outcome in the future. Evaluation and optimization of dynamic policy with observational data have been studied under this setting, and similar doubly robust and semi-parametric efficiency bound has been proposed (\cite{hu2023off}). It remains to be explored whether there exists ``optimal" experiment design that can estimate the dynamic policy or treatment effect most effectively when all the parameters are known and whether we can design an adaptive experiment with near optimal accuracy.

\section{Concluding Remarks}\label{sec:conclusion}
In this paper, we propose a generic learning to design adaptive experimental design framework. We give the first non-asymptotic optimal statistical guarantee for the well-known Neyman allocation problem. A matching lower bound is proposed using novel techniques combining Bayes risk argument and information lower bound. Then we extend this framework to the case with covariates and prove a similar non-asymptotic estimation accuracy that approaches the generalized Neyman allocation bound. The easy-to-implement algorithm is computationally efficient and achieves the theoretically optimal performance. We finally point out some future directions that might be solved under this adaptive experiment framework.

\printendnotes

% Reference
\bibliographystyle{chicago}
\bibliography{citation}

% Appendix
\section{Appendix}
\subsection{Proof of Lemma \ref{lem:propen-opti-nofeature}}\label{proof_lemma1}
Note first that $\E\left(\hat{\tau}_2\right)=\mu^{(1)}-\mu^{(0)}=\tau$, so  
\begin{equation}
\begin{aligned}
    \mathbb{E} \left(\widehat{\tau}_2-\tau\right)^2&= \mathbb{E}
       \left( \sum_{i=1}^n \frac{1}{n} \left(\frac{Y_i-\mu^{(1)}}{e_i}W_i -\frac{Y_i-\mu^{(0)}}{1-e_i}(1-W_i) \right)\right)^2\\
        &=\frac{1}{n^2} \left(\sum_{i=1}^n \mathbb{E}\left(\frac{Y_i-\mu^{(1)}}{e_i}W_i\right)^2+
        \sum_{i=1}^n \mathbb{E}\left(\frac{Y_i-\mu^{(0)}}{1-e_i}(1-W_i)\right)^2 \right.\\
&+\underbrace{\sum_{i=1}^n\mathbb{E}\left(\frac{Y_i-\mu^{(1)}}{e_i}\frac{Y_i-\mu^{(0)}}{1-e_i}W_i (1-W_i)\right)}_{\text{a}}
+\underbrace{\sum_{i,j}\mathbb{E}\left(\frac{Y_i-\mu^{(1)}}{e_i}\frac{Y_j-\mu^{(1)}}{e_j}W_i W_j\right)}_{\text{b}}\\
&\left.+\underbrace{\sum_{i,j}\mathbb{E}\left(\frac{Y_i-\mu^{(0)}}{1-e_i}\frac{Y_j-\mu^{(0)}}{1-e_j}(1-W_i)(1- W_j)\right)}_{\text{c}}+ \underbrace{\sum_{i,j}\mathbb{E}\left(\frac{Y_i-\mu^{(1)}}{e_i}\frac{Y_j-\mu^{(0)}}{1-e_j}W_i(1- W_j)\right)}_{\text{d}}\right)\\
&=\frac{1}{n^2} \left(\sum_{i=1}^n \mathbb{E}\left(\frac{Y_i-\mu^{(1)}}{e_i}W_i\right)^2+
        \sum_{i=1}^n \mathbb{E}\left(\frac{Y_i-\mu^{(0)}}{1-e_i}(1-W_i)\right)^2 \right)
        \end{aligned}
\end{equation}
We  will prove that all the remaining four terms a, b, c, d equal to $0$. For term a,we have
$\sum_{i=1}^n\mathbb{E}\left(\frac{Y_i-\mu^{(1)}}{e_i}\frac{Y_i-\mu^{(0)}}{1-e_i}W_i (1-W_i)\right)$ since $W_i(1-W_i)=0$ always holds. For the remaining three terms b, c, d where $i \neq j$, by symmetry we assume $i<j$ without loss of generality. We will just prove that each expectation in the term b, which is $\mathbb{E}\left(\frac{Y_i-\mu^{(1)}}{e_i}\frac{Y_j-\mu^{(1)}}{e_j}W_i W_j\right)=0$, as the remaining two terms follow the same proof.
$$\begin{aligned}
    \mathbb{E}\left(\frac{Y_i-\mu^{(1)}}{e_i}\frac{Y_j-\mu^{(1)}}{e_j}W_i W_j\right)
    = \mathbb{E}\left( \frac{Y_i-\mu^{(1)}}{e_i} W_i\mathbb{E}\left(\frac{Y_j-\mu^{(1)}}{e_j} W_j \bigg|  \frac{Y_i-\mu^{(1)}}{e_i} W_i\right)\right).
\end{aligned}$$
Since the history $\mathcal{F}_{j-1}$ only affects propensity score $e_j$, and for each fixed $e_j$, $\mathbb{E}\left(\frac{Y_j-\mu^{(1)}}{e_j} W_j \right)=0$, we conclude that the expected expectation always equal to $0$, so $ \mathbb{E}\left(\frac{Y_i-\mu^{(1)}}{e_i}\frac{Y_j-\mu^{(1)}}{e_j}W_i W_j\right)=0$. Similarly we can prove the other two terms to be $0$.
Therefore, we are only left with 
\begin{equation}\label{equ:var-tau2}
\begin{aligned}
    \mathbb{E} \left(\widehat{\tau}_2-\tau\right)^2&= \frac{1}{n^2} \left(\sum_{i=1}^n \mathbb{E}\left(\frac{Y_i-\mu^{(1)}}{e_i}W_i\right)^2+
        \sum_{i=1}^n \mathbb{E}\left(\frac{Y_i-\mu^{(0)}}{1-e_i}(1-W_i)\right)^2 \right)\\
       &=\frac{1}{n^2} \sum_{i=1}^n \E\left(\frac{(\sigma^{(1)})^2}{e_i}+\frac{(\sigma^{(0)})^2}{1-e_i}\right),
       \end{aligned}
       \end{equation}
Since $\mathbb{E}\left(\frac{Y_i-\mu^{(1)}}{e_i}W_i\right)^2=\frac{(\sigma^{(1)})^2}{e_i}$ and $\mathbb{E}\left(\frac{Y_i-\mu^{(0)}}{1-e_i}(1-W_i)\right)^2=\frac{(\sigma^{(0)})^2}{1-e_i}$, and the expectation in the last equation is taken with respect to $e_i$, since they may depend on random history $\mathcal{F}_{i-1}$.
Similarly, we can prove that 
\begin{equation}\label{equ:var-tau*}
\begin{aligned}
    \mathbb{E} \left(\widehat{\tau}^*-\tau\right)^2=\frac{1}{n^2} \sum_{i=1}^n \left(\frac{(\sigma^{(1)})^2}{e^*}+\frac{(\sigma^{(0)})^2}{1-e^*}\right)=\frac{1}{n}\left(\frac{(\sigma^{(1)})^2}{e^*}+\frac{(\sigma^{(0)})^2}{1-e^*}\right).
       \end{aligned}
       \end{equation}
       Combining (\ref{equ:var-tau2}) and (\ref{equ:var-tau*}) together, we complete the proof of lemma \ref{lem:propen-opti-nofeature}.
\subsection{Proof of Lemma \ref{lem-crossterm-nofeature}}\label{Proof_lemma2}
To prove $\mathbb{E} \left( (\widehat{\tau}_1- \widehat{\tau}_2)(\widehat{\tau}_2-\tau) \right)=0$, we only need to prove that $\hat{\tau}_1-\hat{\tau}_2=0$ holds with probability 1. 
Note that 
\begin{equation}
    \begin{aligned}
        \widehat{\tau}_1- \widehat{\tau}_2 &= \frac{1}{n} \sum_{i=1}^n \left(1-\frac{W_i}{e_i}\right) \left(\widehat{\mu}^{(1)}-\mu^{(1)}\right)-\frac{1}{n}\sum_{i=1}^n \left(1-\frac{1-W_i}{1-e_i}\right) \left(\widehat{\mu}^{(0)}-\mu^{(0)}\right)\\
        &=\frac{1}{n} \left(\sum_{i=1}^n \left(1-\frac{W_i}{e_i}\right) \right)\left(\widehat{\mu}^{(1)}-\mu^{(1)}\right)-\frac{1}{n}\left(\sum_{i=1}^n \left(1-\frac{1-W_i}{1-e_i}\right) \right)\left(\widehat{\mu}^{(0)}-\mu^{(0)}\right)
    \end{aligned}
\end{equation}
By the definition of incomplete randomization, for all $i$ in batch $k$ from $\Gamma_{k-1}+1$ to $\Gamma_k$ with length $\gamma_k$, the propensity score $e_i=P(W_i=1)=e_k$ for the same $e_k$ as we define in algorithm \ref{alg:ADM data collecting}. Moreover, we have exactly $e_k \gamma_k$ experiment subjects $i$ with $W_i=1$. Therefore, we always have 
$\sum_{i=\Gamma_{k-1}+1}^{\Gamma_k} (1-\frac{W_i}{e_i})=0.$
Combining all the $K$ batches, we conclude that $\sum_{i=1}^n \left(1-\frac{W_i}{e_i}\right)=0$. Similarly we have $\sum_{i=1}^n \left(1-\frac{1-W_i}{1-e_i}\right)=0$, so $\hat{\tau}_1=\hat{\tau}_2$ holds with probability 1.

\subsection{Proof of Theorem \ref{thm-upperbound-nofeature}}\label{proof_thm1}
First, as we have proved in section \ref{Proof_lemma2}, we have $\hat{\tau}_1-\hat{\tau}_2=0$ holds with probability 1, so $\E (\hat{\tau}_1-\hat{\tau}_2)^2=0$. Equivalently, the AIPW estimator is reduced to the IPW estimator:
\begin{equation}
    \begin{aligned}
\hat{\tau}_1&=\frac{1}{n}\sum_{i=1}^n \left(\frac{Y_i}{e_i}W_i-\frac{Y_i}{1-e_i}(1-W_i)\right)+
\frac{1}{n} \left(\sum_{i=1}^n \left(1-\frac{W_i}{e_i}\right)\hat{\mu}_1-\sum_{i=1}^n \left(1-\frac{1-W_i}{1-e_i}\right)\hat{\mu}_0\right)\\
&  =    \frac{1}{n}\sum_{i=1}^n \left(\frac{Y_i}{e_i}W_i-\frac{Y_i}{1-e_i}(1-W_i)\right),  
    \end{aligned}
\end{equation}
since we have proved in 
section 
\ref{Proof_lemma2} that
$\sum_{i=1}^n \left(1-\frac{W_i}{e_i}\right)=0$ and
$\sum_{i=1}^n \left(1-\frac{1-W_i}{1-e_i}\right)=0$. Furthermore, it's direct to show that $\E(\hat{\tau}_1)=\tau$, since
\begin{align}
    \E (\frac{Y_i}{e_i}W_i)=\mu_1, \quad
    \E (\frac{Y_i}{1-e_i}(1-W_i))=\mu_0
\end{align}
 holds for $i=1 ,\cdots, n$, so $\E(\hat{\tau}_1)=\mu_1-\mu_0=\tau$.

Denote the difference in mean estimator with in each batch as
\begin{align}
&\widehat{\tau}_{k, D M}:=\frac{\sum_{i=\Gamma_{k-1}+1}^{\Gamma_k} W_i Y_i}{\sum_{i=\Gamma_{k-1}+1}^{\Gamma_k} W_i}-\frac{\sum_{i=\Gamma_{k-1}+1}^{\Gamma_k}\left(1-W_i\right) Y_i}{\sum_{i=\Gamma_{k-1}+1}^{\Gamma_k} 1-W_i},
\end{align}
the IPW estimator can be further simplified:
\begin{equation}
    \begin{aligned}
     \hat{\tau}_1&=\frac{1}{n}\sum_{i=1}^n \left(\frac{Y_i}{e_i}W_i-\frac{Y_i}{1-e_i}(1-W_i)\right)\\
     &=
     \sum_{k=1}^K \frac{1}{n}(\sum_{i=\Gamma_{k-1}+1}^{\Gamma_k}\left(\frac{Y_i}{e_i}W_i-\frac{Y_i}{1-e_i}(1-W_i)\right))\\
     &= \sum_{k=1}^K \frac{\gamma_k}{n}\left(\sum_{i=\Gamma_{k-1}+1}^{\Gamma_k}\left(\frac{Y_i}{e_k \gamma_k}W_i-\frac{Y_i}{(1-e_k)\gamma_k}(1-W_i)\right)\right)\\
     &=\sum_{k=1}^K \frac{\gamma_k}{n}\left(
\frac{\sum_{i=\Gamma_{k-1}+1}^{\Gamma_k} W_i Y_i}{\sum_{i=\Gamma_{k-1}+1}^{\Gamma_k} W_i}-\frac{\sum_{i=\Gamma_{k-1}+1}^{\Gamma_k}\left(1-W_i\right) Y_i}{\sum_{i=\Gamma_{k-1}+1}^{\Gamma_k} (1-W_i})\right)\\
&=\sum_{k=1}^K \frac{\gamma_k}{n} \hat{\tau}_{DM}^{(k)},\\
    \end{aligned}
\end{equation}
where $\gamma_k$ is the length of the $k$-th batch from $\Gamma_{k-1}+1$ to $\Gamma_k$. The second equality holds by dividing the total experiment into $K$ batches.
The third equality holds since within batch $k$,  the propensity score $e_i$ equals to the same $e_k$, and the fourth equality holds since by the definition of our experiment, we always have
$\sum_{i=\Gamma_{k-1}+1}^{\Gamma_k} W_i=e_k \gamma_k$ and $\sum_{i=\Gamma_{k-1}+1}^{\Gamma_k} (1-W_i)=(1-e_k) \gamma_k$.
In this form, the AIPW estimator is equivalent to an ``adaptive difference in mean estimator" which we also denote as $\hat{\tau}_{ADM}:=\hat{\tau}_1=\hat{\tau}_2$.
We will prove that when $n$ larger than some universal constant,
\begin{equation}\label{equ:admvariance}    \E[(\widehat{\tau}_{ADM} - \tau)^2] \leq \frac{(\sigma^{(1)} + \sigma^{(0)})^2}{n} + O(\frac{(\sigma^{(0)} + \sigma^{(1)})^2K(\log n)^{\frac{K-1}{K}} }{n^{2 - 1/K} }),
\end{equation}
which is more direct, and all the claims in theorem \ref{thm-upperbound-nofeature} holds as the three estimators $\hat{\tau}_{ADM}$, $\hat{\tau}_1$ and $\hat{\tau}_2$ are equivalent.

First, $\hat{\tau}_{ADM}$ is unbiased as we have proved. 
   To calculus the square error, note that for any two different batches $ 1 \leq l < k \leq K $, we have
   \begin{align*}
      \E[(\widehat{\tau}_{DM}^{(l)} - \tau)(\widehat{\tau}_{DM}^{(k)} - \tau)] &= \E[\E[(\widehat{\tau}_{DM}^{(l)} - \tau)(\widehat{\tau}_{DM}^{(k)} - \tau)|\mathcal{F}_{\Gamma_{k-1} }  ]] \\
      &= \E[(\widehat{\tau}_{DM}^{(l)} - \tau)\E[(\widehat{\tau}_{DM}^{(k)} - \tau)|\mathcal{F}_{\Gamma_{k-1} }  ]] \\
      &= 0,\\
\E[(\widehat{\tau}_{DM}^{(k)} - \tau)^2] &= \E[\E[(\widehat{\tau}_{DM}^{(k)} - \tau)^2|\mathcal{F}_{\Gamma_{k-1} }  ]] \\
      &= \E[\frac{(\sigma^{(1)})^2}{\sum_{t=\Gamma_{k-1} +1}^{\Gamma_k}  W_i} + \frac{(\sigma^{(0)})^2}{\sum_{t=\Gamma_{k-1} +1}^{\Gamma_k}  1 - W_i}] \\
      &= \E[\frac{(\sigma^{(1)})^2(\widetilde{\sigma}_{k-1}^{(1)} + \widetilde{\sigma}_{k-1}^{(0)})}{\widetilde{\sigma}_{k-1}^{(1)}\gamma_k} + \frac{(\sigma^{(0)})^2(\widetilde{\sigma}_{k-1}^{(1)} + \widetilde{\sigma}_{k-1}^{(0)})}{\widetilde{\sigma}_{k-1}^{(0)}\gamma_k}]. 
   \end{align*}
   Hence 
   \begin{align}
      &\E[(\widehat{\tau}_{ADM} - \tau)^2] \nonumber \\
      =& \sum_{k=1}^{K} \frac{\gamma_k^2}{n^2}\E[(\widehat{\tau}_{DM}^{(k)} - \tau)^2] + \sum_{1 \leq l < k \leq K} \frac{\gamma_l\gamma_k}{n^2}\E[(\widehat{\tau}_{DM}^{(l)} - \tau)(\widehat{\tau}_{DM}^{(k)} - \tau)]  \nonumber\\
      =& \sum_{k=1}^{K} \frac{\gamma_k}{n^2}\E[\frac{(\sigma^{(1)})^2(\widetilde{\sigma}_{k-1}^{(1)} + \widetilde{\sigma}_{k-1}^{(0)})}{\widetilde{\sigma}_{k-1}^{(1)}} + \frac{(\sigma^{(0)})^2(\widetilde{\sigma}_{k-1}^{(1)} + \widetilde{\sigma}_{k-1}^{(0)})}{\widetilde{\sigma}_{k-1}^{(0)}}]  \label{1}\\
      =& \sum_{k=1}^{K} \frac{\gamma_k}{n^2} (\sigma^{(1)} + \sigma^{(0)})^2 + \sum_{k=1}^{K} \frac{\gamma_k}{n^2}\E[\frac{(\sigma^{(1)})^2(\widetilde{\sigma}_{k-1}^{(1)} + \widetilde{\sigma}_{k-1}^{(0)})}{\widetilde{\sigma}_{k-1}^{(1)}} + \frac{(\sigma^{(0)})^2(\widetilde{\sigma}_{k-1}^{(1)} + \widetilde{\sigma}_{k-1}^{(0)})}{\widetilde{\sigma}_{k-1}^{(0)}} - (\sigma^{(1)} + \sigma^{(0)})^2] \nonumber \\
      =& \frac{(\sigma^{(1)} + \sigma^{(0)})^2}{n} + \sum_{k=1}^{K} \frac{\gamma_k}{n^2} \E[(\frac{\sigma^{(1)}}{\widetilde{\sigma}_{k-1}^{(1)}} - \frac{\sigma^{(0)}}{\widetilde{\sigma}_{k-1}^{(0)}})^2\widetilde{\sigma}_{k-1} ^{(1)}\widetilde{\sigma}_{k-1} ^{(0)}] \nonumber \\
      \leq& \frac{(\sigma^{(1)} + \sigma^{(0)})^2}{n} + 2\sum_{k=1}^{K} \frac{\gamma_k}{n^2}\E[((\frac{\sigma^{(1)}}{\widetilde{\sigma}_{k-1}^{(1)}} - 1)^2 + (\frac{\sigma^{(0)}}{\widetilde{\sigma}_{k-1}^{(0)}} - 1)^2)\widetilde{\sigma}_{k-1} ^{(1)}\widetilde{\sigma}_{k-1} ^{(0)}] \nonumber \\
      =& \frac{(\sigma^{(1)} + \sigma^{(0)})^2}{n} + 2\sum_{k=1}^{K} \frac{\gamma_k}{n^2} \E[(\sigma^{(1)} - \widetilde{\sigma}_{k-1}^{(1)})^2\frac{\widetilde{\sigma}_{k-1}^{(0)}}{\widetilde{\sigma}_{k-1}^{(1)}} + (\sigma^{(0)} - \widetilde{\sigma}_{k-1}^{(0)})^2\frac{\widetilde{\sigma}_{k-1}^{(1)}}{\widetilde{\sigma}_{k-1}^{(0)}}] .\label{2}
   \end{align}
 For $ w \in \{1, 0\}, k \in [K] $, define events as follows, 
   \[
      \mathcal{E}_{k}^{(w)}  := \big\{ (\sigma^{(w)}  - \widehat{\sigma}_{k}^{(w)} )^2 \leq C_2(\frac{\widetilde{\sigma}_{k-1}^{(w)} }{\widetilde{\sigma}_{k-1} ^{(1)} + \widetilde{\sigma}_{k-1} ^{(0)}}\gamma_k)^{-1}(\log n)(\sigma^{(w)} )^2 \big\}
   \]
for some $C_2$ as defined in the following lemma.
\begin{lemma}\label{lem:concentration_of_variance}
(Concentration of standard error estimation)
 Let $ X_1, \cdots, X_m $ be $ m $ i.i.d sample drawn from a distribution $ \mathcal{P}_X $ with variance $ \sigma^2 $. 
   Define sample variance as $ \widehat{\sigma}^2 := \frac{1}{m-1}\sum_{i=1}^{m} (X_i - \sum_{j=1}^{m}X_j / m)^2 $.
   Suppose $ \mathcal{P}_X $ is sub-Gaussian and there is an absolute constant $ C $ such that for any $ \lambda \in \mathbb{R} $, 
   \begin{equation}
      \label{sub-Gaussian}
      \E_{X \sim \mathcal{D}_X} [\exp(\lambda(X - \E[X]))] \leq \exp(C^2\lambda^2\sigma^2).
   \end{equation}
   If $ m \geq C_1\log \delta^{-1} + 1  $ for $C_1=2$, then with probability $ 1 - 2\delta $, 
   \[
      ( \widehat{\sigma}-\sigma)^2 \leq 256C^4\sigma^2 m^{-1} (\log \delta^{-1}). 
   \] 
As a result, every event $\mathcal{E}_{k}^{(w)}$ happens with probability $ 1 - \frac{1}{2n^3} $ for some constant $C_2=256C^4$.
\end{lemma}
The proof of this lemma is delayed to section \Cref{proof:lem:concentration_of_variance}.
Assume lemma \ref{lem:concentration_of_variance} holds, then by union bound, with probability $ 1 - \frac{K}{n^3} \geq 1 - \frac{1}{n^2}$, all the events $ \mathcal{E} := \bigcap_{k \in [K]}(\mathcal{E}_k^{(1)} \cap \mathcal{E}_{k}^{(0)}) $ happen simultaneously. 
   Now we consider whether $ \mathcal{E} $ happens. 
   \subsubsection*{case1: $ \mathcal{E} $ doesn't happen.}
   Then by equation (\ref{1}) and the definition of $ \widetilde{\sigma}_{k-1} $, we have
   \[
      \E[(\widehat{\tau}_{ADM} - \tau)^2 | \lnot \mathcal{E}] \leq \sum_{k=1}^{K} \frac{\gamma_k}{n^2}((\sigma^{(1)})^2 + (\sigma^{(0)})^2)n = ((\sigma^{(1)})^2 + (\sigma^{(0)})^2)
   \] 
   \subsubsection*{case2: $ \mathcal{E} $ happens.}
   First note that for any $ k \in [K] $, by our definition of the exploration threshold, we know that $\frac{\widetilde{\sigma}_{k-1}^{(1)}}{\widetilde{\sigma}_{k-1}^{(1)} + \widetilde{\sigma}_{k-1}^{(0)}}\gamma_k \ge \frac{\gamma_1}{4}= n^{\frac{1}{K}} (\log(n))^{\frac{K-1}{K}} \ge 4C_2 \log n$ for $n$ larger than some constant, so we have
   \begin{align*}
      (\sigma^{(1)} - \widehat{\sigma}_{k}^{(1)})^2 \leq C_2 (\frac{\widetilde{\sigma}_{k-1}^{(1)}}{\widetilde{\sigma}_{k-1}^{(1)} + \widetilde{\sigma}_{k-1}^{(0)}}\gamma_k )^{-1} (\log n)(\sigma^{(1)})^2
      \leq C_2(4C_2\log n)^{-1}(\log n)(\sigma^{(1)})^2= \frac{(\sigma^{(1)})^2}{4}.
   \end{align*}
   Hence 
   \[ 
      \sigma^{(1)} / 2 \leq \widehat{\sigma}_k^{(1)} \leq 2\sigma^{(1)}, 
      \sigma^{(0)} / 2 \leq \widehat{\sigma}_k^{(0)} \leq 2\sigma^{(0)}.
   \]
   By the definition of $ \mathcal{E}_k^{(1)} $ and the fact that $ \frac{\widetilde{\sigma}_{k-1}^{(0)}}{\widetilde{\sigma}_{k-1}^{(1)}} \leq 1 + \frac{\widehat{\sigma}_{k-1}^{(0)}}{\widehat{\sigma}_{k-1}^{(1)}} \leq 1 + 4\frac{\sigma^{(0)}}{\sigma^{(1)} }$, 
   We have 
   \begin{equation}
      \label{eq:without X, variance precision, 1}   
      (\sigma^{(1)}  - \widehat{\sigma}_{k}^{(1)} )^2 \leq C_2(\frac{\widetilde{\sigma}_{k-1}^{(1)} }{\widetilde{\sigma}_{k-1} ^{(1)} + \widetilde{\sigma}_{k-1} ^{(0)}}\gamma_k)^{-1}(\log n)(\sigma^{(1)} )^2 \leq C_2(2 + 4\frac{\sigma^{(0)} }{\sigma^{(1)} })\gamma_k^{-1} (\log n)(\sigma^{(1)})^2.
   \end{equation}
   Now we analyse the terms in equation (\ref{2}) to get the final result.
   For any $ k \in[K] $, if $ \widehat{\sigma}_{k-1}^{(1)} \geq \frac{n^{1/K}(\log n)^{(K-1) / K} }{\gamma_k}\widehat{\sigma}_{k-1}^{(1)} $ or $ \sigma^{(1)} \geq \frac{n^{1/K}(\log n)^{(K-1)/K} }{\gamma_k}\widehat{\sigma}_{k-1}^{(0)} $, 
   then $ (\sigma^{(1)} - \widetilde{\sigma}_{k-1}^{(1)})^2 \leq (\sigma^{(1)} - \widehat{\sigma}_{k-1}^{(1)})^2 $.
   Hence by (\ref{eq:without X, variance precision, 1}) and 
   \begin{align}
      (\sigma^{(1)} - \widetilde{\sigma}_{k-1}^{(1)})^2\frac{\widetilde{\sigma}_{k-1}^{(0)}}{\widetilde{\sigma}_{k-1}^{(1)}} \nonumber 
      &\leq (\sigma^{(1)} - \widehat{\sigma}_{k-1}^{(1)})^2\frac{\widetilde{\sigma}_{k-1}^{(0)}}{\widetilde{\sigma}_{k-1}^{(1)}} \\  \nonumber
      &\leq C_2(2+4\frac{\sigma^{(0)}}{\sigma^{(1)}})(\gamma_{k-1} )^{-1}(\log n)(\sigma^{(1)})^2(1 + \frac{4\sigma^{(0)}}{\sigma^{(1)}}) \\  \nonumber
      &\leq 16C_2\gamma_{k-1}^{-1}\log n (\sigma^{(0)} + \sigma^{(1)})^2 \nonumber \\ 
      &\leq 16C_2\frac{n^{1/K}(\log n)^{-1/K} }{\gamma_k}\log n(\sigma^{(0)} + \sigma^{(1)})^2 \nonumber\\
      &\leq 16C_2\frac{n^{1/K}}{\gamma_k}(\log n)^{(K-1) / K} (\sigma^{(0)} + \sigma^{(1)})^2.\label{3}
   \end{align}
   Otherwise, if $ \widehat{\sigma}_{k-1}^{(1)} < \frac{n^{1/K}(\log n)^{(K-1)/K}  }{\gamma_k}\widehat{\sigma}_{k-1}^{(0)} $ and $ \sigma^{(1)} < \frac{n^{1/K}(\log n)^{(K-1)/K} }{\gamma_k}\widehat{\sigma}_{k-1}^{(0)} $, 
   then $ \widetilde{\sigma}_{k-1}^{(1)} = \widehat{\sigma}_{k-1}^{(0)}n^{1/K}(\log n)^{(K-1)/K} /\gamma_k $ and $ \widetilde{\sigma}_{k-1}^{(0)} = \widehat{\sigma}_{k-1}^{(0)} $.
   Hence 
   \begin{equation}
      \label{4}
      (\sigma^{(1)} - \widetilde{\sigma}_{k-1}^{(1)})^2\frac{\widetilde{\sigma}_{k-1}^{(0)}}{\widetilde{\sigma}_{k-1}^{(1)}} 
      \leq (\widetilde{\sigma}_{k-1}^{(1)})^2\frac{\widetilde{\sigma}_{k-1}^{(0)}}{\widetilde{\sigma}_{k-1}^{(1)}} 
      = \frac{n^{1/K}(\log n)^{(K-1)/K} }{\gamma_k}(\widehat{\sigma}_{k-1}^{(0)})^2 \leq 4\frac{n^{1/K}(\log n)^{(K-1)/K} }{\gamma_k}(\sigma^{(0)})^2.
   \end{equation}
   Combining the equation (\ref{3}) and (\ref{4}), we have 
   \[
      (\sigma^{(1)} - \widetilde{\sigma}_{k-1}^{(1)})^2\frac{\widetilde{\sigma}_{k-1}^{(0)}}{\widetilde{\sigma}_{k-1}^{(1)}} 
      \leq (16C_2 + 4)\frac{n^{1/K}(\log n)^{(K-1)/K} }{\gamma_k}(\sigma^{(0)} + \sigma^{(1)})^2.
   \] 
   By a symmetry argument, we also have
   \[
      (\sigma^{(0)} - \widetilde{\sigma}_{k-1}^{(0)})^2\frac{\widetilde{\sigma}_{k-1}^{(1)}}{\widetilde{\sigma}_{k-1}^{(0)}} 
      \leq (16C_2 + 4)\frac{n^{1/K} (\log n)^{(K-1)/K}}{\gamma_k}(\sigma^{(0)} + \sigma^{(1)})^2
   \] 
   For every $ k \in [K] $, plugging these into equation (\ref{2}), with $ \mathcal{E} $ happens, we have 
   \[
      \E[(\widehat{\tau}_{ADM} - \tau)^2 | \mathcal{E}] 
      \leq \frac{(\sigma^{(1)} + \sigma^{(0)})^2}{n} + \frac{(32C_2+8)K(\sigma^{(1)} + \sigma^{(0)})^2(\log n)^{(K-1)/K}}{n^{2 - 1/K} } .
   \] 
   Finally, by the Law of Total Probability, we have 
   \begin{align*}
      \E[(\widehat{\tau}_{ADM} - \tau)^2] &=  
      \Pr[\mathcal{E}]\E[(\widehat{\tau}_{ADM} - \tau)^2 | \mathcal{E}] +
      \Pr[\lnot\mathcal{E}]\E[(\widehat{\tau}_{ADM} - \tau)^2 | \lnot\mathcal{E}] \\
      &\leq \frac{(\sigma^{(1)} + \sigma^{(0)})^2}{n} + \frac{(32C_2 + 8)K(\sigma^{(1)}+ \sigma^{(0)})^2(\log n)^{(K-1)/K}}{n^{2 - 1/K} } + \frac{(\sigma^{(1)})^2 + (\sigma^{(0)})^2}{n^2},
   \end{align*}
   and the desired result in (\ref{equ:admvariance}) is proved.
\subsection{Proof of Lemma \ref{lem:concentration_of_variance}} \label{proof:lem:concentration_of_variance}
   First rewrite sample variance as 
   \[
      \widehat{\sigma}^2 = \frac{1}{n(n-1)}\sum_{1 \leq i < j \leq n} (X_i - X_j)^2.
   \] 
   Let $ \pi = (\pi(1), \cdots, \pi(n)) $ be a permutation of $ [n] $ and define
   \[
      V(\pi) := \frac{1}{2k}((X_{\pi(1)}  - X_{\pi(2)} )^2 + (X_{\pi(3)}  - X_{\pi(4)} )^2 + \cdots + (X_{\pi(2k-1)} - X_{\pi(2k)})^2), \text{ where } k = \lfloor \frac{n}{2} \rfloor,
   \] 
   then $ \widehat{\sigma}^2 $ can be written as the mean of $ V(\pi) $ over all permutation, namely
   \[
      \widehat{\sigma}^2 = \frac{1}{n!} \sum_{\pi} V(\pi).
   \] 
   Hence for any $ h > 0 $, 
   \begin{align}
      \label{Markov inequality}
      \Pr[\widehat{\sigma}^2 - \sigma^2 \geq t] &\leq \E[\exp(h(\widehat{\sigma}^2 - \sigma^2 - t)) ] \\
      &= e^{-ht}\E[\exp(h(\widehat{\sigma}^2 - \sigma^2)) ] \\
      &= e^{-ht}\E[\exp(h(\frac{1}{n!}\sum_{\pi}V(\pi) - \sigma^2))]  \\
      &\leq e^{-ht}(\frac{1}{n!}\sum_{\pi}\E[\exp(hV(\pi) - h\sigma^2)]) \label{Jensen inequality} \\
      &= e^{-ht}\E[\exp(\frac{h}{2k}((X_{1}  - X_{2} )^2 + (X_{3}  - X_{4} )^2 + \cdots + (X_{2k-1} - X_{2k})^2) - h\sigma^2)]\label{same distribution} \\
      &= e^{-ht}\E[\exp(\frac{h}{k}((X_1 - X_2)^2/2 - \sigma^2))]^k, \label{independence}
   \end{align}
where \eqref{Markov inequality} is due to $ \widehat{\sigma}^2 - \sigma^2 \geq t \implies \exp(h(\widehat{\sigma}^2 - \sigma^2 - t)) \geq 1 $;
\eqref{Jensen inequality} is due to the convexity of exponential function and Jensen inequality;
\eqref{same distribution} is due to the fact that for any permutation $ \pi $, the distribution of $ V(\pi) $ are same;
\eqref{independence} is due to the independence between $ X_1 - X_2, X_3 - X_4, \cdots, X_{2k-1} - X_{2k} $. \\
To further bound the above equation, we need to show that $ (X_1 - X_2)^2 / 2 $ is sub-exponential. 
\begin{lemma}\label{lem:subexponential}
For any $ - \frac{1}{4C^2\sigma^2} \leq \lambda \leq \frac{1}{4C^2   \sigma^2}$, we have 
\[
   \E[\exp(\lambda(X_1 - X_2)^2/2 - \sigma^2)] \leq \exp(16C^4\lambda^2\sigma^4).
\]  
\end{lemma}
\noindent\textbf{Proof of lemma \ref{lem:subexponential}.}

   By our assumption \ref{sub-Gaussian}, we can show that for any $ \lambda \in \mathbb{R} $,
   \begin{align*}
      \E[\exp(\lambda (X_1 - X_2) / \sqrt{2} )] 
      &= \E[\exp(\lambda (X_1 - \E[X_1] - X_2 + \E[X_2]) / \sqrt{2} )] \\
      &= \E[\exp(\lambda (X_1 - \E[X_1]) / \sqrt{2} )] \E[\exp(-\lambda(X_2 - \E[X_2]) / \sqrt{2})] \\
      &\leq \exp(C^2\lambda^2\sigma^2/2)\exp(C^2\lambda^2\sigma^2/2) = \exp(C^2\lambda^2\sigma^2).
   \end{align*}
   Let $ \Gamma(r) $ be the Gamma function, then the moments of the sub-Gaussian variable $ (X_1 - X_2)/\sqrt{2} $ are bounded as follows:
   \[
      \forall r \geq 0, \E[|(X_1 - X_2)/\sqrt{2} |^r] \leq r2^{r/2}(C\sigma)^r \Gamma(r/2).
   \]  
Noting that $ \E[(X_1 - X_2)^2 / 2] = \sigma^2 $ and by power series expansion, we have
\begin{align*}
\mathbb{E}[\exp(\lambda((X_1 - X_2)^2/2 - \sigma^2))]
&=1+\lambda\mathbb{E}[(X_1 - X_2)^2 - \sigma^2] + \sum_{r=2}^\infty\frac{\lambda^r\mathbb{E}[((X_1 - X_2)^2/2 - \sigma^2)^r]}{r!} \\
&\leq1+\sum_{r=2}^\infty\frac{\lambda^r\mathbb{E}[|(X_1 - X_2) / \sqrt{2} |^{2r}] }{r!} \\
&\leq1+\sum_{r=2}^\infty\frac{\lambda^r2r2^r(C\sigma)^{2r}\Gamma(r)}{r!} \\
&=1+\sum_{r=2}^\infty \lambda^r2^{r+1}(C\sigma)^{2r} \\
&=1+\frac{8\lambda^2(C\sigma)^4}{1-2\lambda(C\sigma)^2}.
\end{align*}
By restricting $ |\lambda| \leq 1 / 4C^2\sigma^2 $, we have $ 1/(1 - 2\lambda C^2\sigma^2) \leq 2 $. 
Since $ \forall x, 1 + x \leq e^{x}  $, we have proved that 
\[
   \forall |\lambda| \leq 1/4C^2\sigma^2, \E[\exp(\lambda(X_1 - X_2)^2/2 - \sigma^2)] \leq 1 + 16\lambda^2(C\sigma)^4 \leq \exp(16C^4\lambda^2\sigma^4)
\] 
and finish the proof of lemma \ref{lem:subexponential}.

Now let $ t = 8C^2k^{-1/2} \sigma^2\sqrt{\log \delta^{-1} }  $, 
$ h = \frac{1}{4}C^{-2}k^{1/2} \sigma^{-2}\sqrt{\log \delta^{-1} } $ in \eqref{independence}.
By $ k \geq (n - 1) / 2 \geq \log \delta^{-1}$, we have $ h / k = \frac{1}{4}C^{-2}k^{-1/2}\sigma^{-2}\sqrt{\log \delta^{-1} } \leq 1 / (4C^2\sigma^2) $.
Hence by the lemma and \eqref{independence}, we get
\[
   \Pr[\widehat{\sigma}^2 - \sigma^2 \geq 8C^{2} k^{-1/2}\sigma^2\sqrt{\log \delta^{-1} }] \leq e^{-ht}(\exp(16C^4\sigma^4h^2/k^2))^k = e^{-2\log \delta^{-1}+\log\delta^{-1} } =\delta.
\] 
Hence 
\begin{equation}
\label{bound1}
\begin{aligned}
   \Pr[\widehat{\sigma} - \sigma \geq 8C^{2} k^{-1/2}\sigma\sqrt{\log \delta^{-1} }]
   &= \Pr[\widehat{\sigma}^2 \geq  (1 + 8C^{2} k^{-1/2}\sqrt{\log \delta^{-1} })^2\sigma^2] \\
   &\leq \Pr[\widehat{\sigma}^2 \geq  (1 + 8C^{2} k^{-1/2}\sqrt{\log \delta^{-1} })\sigma^2] \\
   &= \Pr[\widehat{\sigma}^2 - \sigma^2 \geq 8C^{2} k^{-1/2}\sigma^2\sqrt{\log \delta^{-1} }] \leq \delta.
\end{aligned}
\end{equation}
Now we bound the probability that $ \widehat{\sigma} $ is too small. 
Similar to (\ref{Markov inequality}), for any $ h > 0 $, we have
\begin{align*}
   \Pr[\widehat{\sigma}^2 - \sigma^2 \leq -t] &\leq \E[\exp(-h(\widehat{\sigma}^2 - \sigma^2 + t)) ] \\
   &= e^{-ht}\E[\exp(-h(\widehat{\sigma}^2 - \sigma^2)) ] \\
   &= e^{-ht}\E[\exp(-h(\frac{1}{n!}\sum_{\pi}V(\pi) - \sigma^2))]  \\
   &\leq e^{-ht}(\frac{1}{n!}\sum_{\pi}\E[\exp(-hV(\pi) + h\sigma^2)]) \\
   &= e^{-ht}\E[\exp(\frac{-h}{2k}((X_{1}  - X_{2} )^2 + (X_{3}  - X_{4} )^2 + \cdots + (X_{2k-1} - X_{2k})^2) - h\sigma^2)] \\
   &= e^{-ht}\E[\exp(\frac{-h}{k}((X_1 - X_2)^2/2 - \sigma^2))]^k.  
\end{align*}
Again let $ t = 8C^2k^{-1/2} \sigma^2\sqrt{\log \delta^{-1} }  $, 
$ h = \frac{1}{4}C^{-2}k^{1/2} \sigma^{-2}\sqrt{\log \delta^{-1} } $.
By $ k \geq (n - 1) / 2 \geq \log \delta^{-1}$, we have $ -h / k = -\frac{1}{4}C^{-2}k^{-1/2}\sigma^{-2}\sqrt{\log \delta^{-1} } \geq -1 / (4C^2\sigma^2) $.
Hence by the lemma, we get 
\[
   \Pr[\widehat{\sigma}^2 - \sigma^2 \leq -8C^{2} k^{-1/2}\sigma^2\sqrt{\log \delta^{-1} }] \leq e^{-ht}(\exp(16C^4\sigma^4h^2/k^2))^k = e^{-2\log \delta^{-1}+\log\delta^{-1} } =\delta.
\] 
Therefore if $1 - 8C^{2} k^{-1/2}\sqrt{\log \delta^{-1} } \geq 0 $, 
\begin{equation}
\label{bound2}
\begin{aligned}
   \Pr[\widehat{\sigma} - \sigma \leq -8C^{2} k^{-1/2}\sigma\sqrt{\log \delta^{-1} }]
   &= \Pr[\widehat{\sigma}^2 \leq  (1 - 8C^{2} k^{-1/2}\sqrt{\log \delta^{-1} })^2\sigma^2] \\
   &\leq \Pr[\widehat{\sigma}^2 \leq  (1 - 8C^{2} k^{-1/2}\sqrt{\log \delta^{-1} })\sigma^2] \\
   &= \Pr[\widehat{\sigma}^2 - \sigma^2 \leq - 8C^{2} k^{-1/2}\sigma^2\sqrt{\log \delta^{-1} }] \leq \delta.
\end{aligned}
\end{equation}
If $ 1 - 8C^{2} k^{-1/2}\sqrt{\log \delta^{-1} } < 0 $, 
then $   \Pr[\widehat{\sigma} - \sigma \leq -8C^{2} k^{-1/2}\sigma\sqrt{\log \delta^{-1} }] = 0 < \delta $.
Combined with \eqref{bound1} and \eqref{bound2}, we get the final result as follows:
\begin{align*}
   \Pr[(\sigma - \widehat{\sigma})^2 \leq 256C^4n^{-1}\sigma^2\log \delta^{-1} ] 
   &\geq 1 - \Pr[\widehat{\sigma} - \sigma \geq 8C^{2} k^{-1/2}\sigma\sqrt{\log \delta^{-1} }] - \Pr[\widehat{\sigma} - \sigma \leq -8C^{2} k^{-1/2}\sigma\sqrt{\log \delta^{-1} }] \\
   &\geq 1 - 2\delta.
\end{align*}

\subsection{Proof of Theorem \ref{thm-lowerbound-nofeature}}
First, we construct Gaussian Bayes instances of the experiment process. In a Bayes instance, $ \mu^{(1)}, \mu^{(0)} $ are no longer fixed parameters but random variables follow a specific prior distribution and $ \sigma^{(0)}, \sigma^{(1)} $ are still fixed being parameters of the instance.
\begin{definition}[Gaussian Bayes instances]
   \label{Bayes instance}
   Any Gaussian Bayes instance accepting three parameters, $ \rho, \sigma^{(1)}, \sigma^{(0)}  $. 
   Data generated as follows:
   \begin{equation}
      \begin{aligned}
         &(\mu^{(1)}, \mu^{(0)}) \sim \mathcal{N}(\bm 0, \rho^2\bm I_2) \\
         & \forall i \in [n], (Y_i^{(1)}, Y_i^{(0)}) | (\mu^{(1)}, \mu^{(0)}) \sim \mathcal{N}((\mu^{(1)}, \mu^{(0)}), \text{diag}((\sigma^{(1)})^2, (\sigma^{(0)})^2)) 
      \end{aligned}
   \end{equation}
\end{definition}
For any Gaussian Bayes instance and any data generation policy, we can compute the posterior distribution of $ \mu^{(1)}, \mu^{(0)} $. 
Specifically, we have that
\begin{lemma}[posterior of $ \mu $]
   For any $ i \in [n] \cup \{0\} $, 
\begin{align}
   \label{posterior}
   &(\mu^{(1)}, \mu^{(0)}) | \mathcal{F}_i \sim \nonumber \\
   &\mathcal{N}((\frac{\sum_{j=1}^{i}W_jY_j}{(\sigma^{(1)})^2/\rho^2 + \sum_{j=1}^{i} W_j}, \frac{\sum_{j=1}^{i}(1-W_j)Y_j}{(\sigma^{(0)})^2/\rho^2 + \sum_{j=1}^{i} (1 - W_j)}), \begin{bmatrix}
      \frac{(\sigma^{(1)})^2}{(\sigma^{(1)})^2/\rho^2 + \sum_{j=1}^{i} W_j}& 0 \\
      0& \frac{(\sigma^{(0)})^2}{(\sigma^{(0)})^2/\rho^2 + \sum_{j=1}^{i} (1 - W_j)} 
   \end{bmatrix})
\end{align}
\end{lemma}
We now use induction on $ i $ to prove the lemma. \\
\textbf{Base case}: For $ i = 0 $, \eqref{posterior} is same as the prior distribution of $ (\mu^{(1)}, \mu^{(0)}) $ and $ \mathcal{F}_0 = \emptyset $. Hence there is nothing to prove.\\
\textbf{Induction step}: Assume \eqref{posterior} holds for $ i - 1 $. Here we show that it also holds for $ i $. 
Since data generation policy is oblivious to the $ \mu^{(1)}, \mu^{(0)} $, the conditional distribution $ W_i | \mathcal{F}_i $ is independent of $ \mu^{(1)}, \mu^{(0)} $.
Let $ p(\mu^{(1)}, \mu^{(0)}|\mathcal{F}_{i-1}, W_i, Y_i) $ denote Radon-Nikodym derivative with respect to the Lebesgue measure on $ \mathbb{R}^2 $. 
By Bayes' rules, we have
\begin{align*}
   p(\mu^{(1)}, \mu^{(0)}|\mathcal{F}_{i-1}, W_i, Y_i) \propto& \exp(-\frac{1}{2}(\frac{\sum_{j=1}^{i-1}W_jY_j}{(\sigma^{(1)})^2/\rho^2 + \sum_{j=1}^{i-1} W_j} - \mu^{(1)})^2\frac{(\sigma^{(1)})^2/\rho^2 + \sum_{j=1}^{i-1} W_j}{(\sigma^{(1)})^2}) \\
   & \times \exp(-\frac{1}{2}(\frac{\sum_{j=1}^{i-1}(1-W_j)Y_j}{(\sigma^{(0)})^2/\rho^2 + \sum_{j=1}^{i-1} 1-W_j} - \mu^{(0)})^2\frac{(\sigma^{(0)})^2/\rho^2 + \sum_{j=1}^{i-1} 1-W_j}{(\sigma^{(0)})^2}) \\
   &\times \exp(-\frac{1}{2}((Y_i - \mu^{(W_i)})^2)/(\sigma^{(W_i)})^2) \\
   \propto& \exp(-\frac{1}{2}(\frac{\sum_{j=1}^{i}W_jY_j}{(\sigma^{(1)})^2/\rho^2 + \sum_{j=1}^{i} W_j} - \mu^{(1)})^2\frac{(\sigma^{(1)})^2/\rho^2 + \sum_{j=1}^{i-1} W_j}{(\sigma^{(1)})^2}) \\
   & \times \exp(-\frac{1}{2}(\frac{\sum_{j=1}^{i}(1-W_j)Y_j}{(\sigma^{(0)})^2/\rho^2 + \sum_{j=1}^{i} 1-W_j} - \mu^{(0)})^2\frac{(\sigma^{(0)})^2/\rho^2 + \sum_{j=1}^{i-1} 1-W_j}{(\sigma^{(0)})^2}),
\end{align*}
where the last $ \propto $ can be checked by plugging in $ W_i = 1 $ and $ W_i = 0 $ respectively.
This implies that \eqref{posterior} holds for $ i $, and hence completes the proof of the lemma.

Hence, condition on $ \mathcal{F}_n $, $ \mu^{(1)} $ and $ \mu^{(0)} $ are independent and has variance $ \frac{(\sigma^{(1)})^2}{(\sigma^{(1)})^2/\rho^2 + \sum_{j=1}^{i} W_j}, \frac{(\sigma^{(0)})^2}{(\sigma^{(1)})^2/\rho^2 + \sum_{j=1}^{i} (1 - W_j)}$ respectively.
So for any estimator $ \widehat{\tau}(\mathcal{F}_n) $, we have
\begin{align}
   \label{lower bound1}
   \E[(\widehat{\tau}(\mathcal{F}_n) - \tau)^2] &= \E[\E[(\widehat{\tau}(\mathcal{F}_n) - \tau)^2 | \mathcal{F}_n]] \nonumber \\
   &= \E[\widehat{\tau}(\mathcal{F}_n)^2 - 2\widehat{\tau}(\mathcal{F}_n)\E[\tau | \mathcal{F}_n] + \E[\tau^2 | \mathcal{F}_n]] \nonumber \\
   &\geq \E[\Var[\tau | \mathcal{F}_n]] = \E[\Var[\mu^{(1)} - \mu^{(0)}| \mathcal{F}_n]] \nonumber \\
   &= \E[\Var[\mu^{(1)} | \mathcal{F}_n] + \Var[\mu^{(0)}|\mathcal{F}_n]] \nonumber \\
   &= \E[\frac{(\sigma^{(1)})^2}{(\sigma^{(1)})^2/\rho^2 + \sum_{j=1}^{i} W_j} + \frac{(\sigma^{(0)})^2}{(\sigma^{(0)})^2/\rho^2 + \sum_{j=1}^{i} (1 - W_j)}] 
\end{align}    

To further get the lower bound in terms of $ n $, we need to invoke the information lower bound argument on two Bayes instances. 
We construct two Bayes instances such that any data generation policy cannot differentiate them well.

In the first instance $ \nu $, we parameterize Gaussian Bayes instance with $ \sigma_{\nu}^{(1)} = 1, \sigma_{\nu}^{(0)} = 1 + \epsilon  $,
and in the second instance $ \nu' $, let $ \sigma_{\nu'} ^{(1)} = 1 + \epsilon, \sigma_{\nu'} ^{(0)} = 1 $,
where $ \epsilon > 0 $ to be determined.
We use $ \mathbb{P}_{\nu}  $ to denote the probability measurement on the first instance. 
and $ \mathbb{P}_{\nu'}  $ to denote the probability measurement on the second instance. 
Let $ u(y; \mu, \sigma) $ be the probability density function of the Gaussian distribution $ \mathcal{N}(\mu, \sigma) $. 
Then by the definition \ref{Bayes instance}, Radon-Nikodym derivative of $ \mathbb{P}_{\nu}  $ with respect to the product measure $ \mathbb{R}^2 \times (\mathbb{R} \times \mathbb{R})^n $ is 
\begin{align*}
   &p(\mu^{(1)}, \mu^{(0)}, y_1, w_1, \cdots, y_n, w_n) \\
   =&u(\mu^{(1)}; 0, 1/\rho^2)u(\mu^{(0)}; 0, 1/\rho^2)\prod_{i=1}^{n} (u(y_i; \mu^{(w_i)}, \sigma_{\nu}^{(w_i)})\Pr_{\nu}[W_i = w_i|y_1, w_1, \cdots, y_{i-1}, w_{i-1}])
\end{align*}
Radon-Nikodym derivative of $ \mathbb{P}_{\nu'} $ is 
\begin{align*}
   &p(\mu^{(1)}, \mu^{(0)}, y_1, w_1, \cdots, y_n, w_n) \\
   =&u(\mu^{(1)}; 0, 1/\rho^2)u(\mu^{(0)}; 0, 1/\rho^2)\prod_{i=1}^{n} (u(y_i; \mu^{(w_i)}, \sigma_{\nu'}^{(w_i)})\Pr_{\nu'}[W_i = w_i|y_1, w_1, \cdots, y_{i-1}, w_{i-1}])
\end{align*}
Again, data generation policy is oblivious to $ \sigma^{(1)}, \sigma^{(0)}  $, so we have
\[
   \Pr_{\nu}[W_i = w_i|y_1, w_1, \cdots, y_{i-1}, w_{i-1}] = \Pr_{\nu'}[W_i = w_i|y_1, w_1, \cdots, y_{i-1}, w_{i-1}]
\] 
Hence 
\begin{align*}
   D_{\text{KL}}(\mathbb{P}_{\nu} ||\mathbb{P}_{\nu'} ) &= \E_{\nu}[\sum_{i=1}^{n} \log \frac{u(Y_i; \mu^{(W_i)}, \sigma_{\nu}^{(W_i)} )}{u(Y_i; \mu^{(W_i)}, \sigma_{\nu'}^{(W_i)} )}] \\
   &\leq n\max\{\E_{\nu}[\log \frac{u(Y_i; \mu^{(1)}, 1)}{u(Y_i; \mu^{(1)}, 1 + \epsilon)} | W_n = 1], \E_{\nu}[\log \frac{u(Y_i; \mu^{(0)}, 1 + \epsilon)}{u(Y_i; \mu^{(0)}, 1)} | W_i = 0]\} \\
   &= n\max\{\E_{\mu \sim \mathcal{N}(0, \rho^2)}[D_{\text{KL}}(\mathcal{N}(\mu, 1) || \mathcal{N}(\mu, (1 + \epsilon)^2)) ], \E_{\mu \sim \mathcal{N}(0, \rho^2)} [D_{\text{KL}}(\mathcal{N}(\mu, (1 + \epsilon)^2) || \mathcal{N}(\mu, 1)) ] \} \\
   &= n\max\{-\log (1 + \epsilon) + \frac{(1+\epsilon)^2}{2} - \frac{1}{2}, -\log \frac{1}{1+\epsilon} + \frac{1}{2(1+\epsilon)^2} - \frac{1}{2}\} \\
   &= n\max\{\log (1 - \frac{\epsilon}{1 + \epsilon}) + \frac{(1+\epsilon)^2}{2} - \frac{1}{2}, \log (1+\epsilon) + \frac{1}{2(1+\epsilon)^2} - \frac{1}{2}\} \\
   &\leq n\max\{-\frac{\epsilon}{1 + \epsilon} + \frac{(1+\epsilon)^2}{2} - \frac{1}{2}, \epsilon + \frac{1}{2(1+\epsilon)^2} - \frac{1}{2}\} \\
   &= n\max\{\epsilon^2(\frac{1}{2}+\frac{1}{1+\epsilon}), \epsilon^2(\frac{2\epsilon+3}{2(1+\epsilon)^2})\} \\
   &\leq 2n\epsilon^2,
\end{align*}
where the second inequality comes from the fact that $ x > 0 \implies \log(1+x) \leq x $ and $ 0 < x < 1 \implies \log (1-x) < -x $.
\\
Now we set $ \epsilon = \frac{1}{4}n^{-1/2}  $, which implies that $ D_{\text{KL}(\nu || \nu')} \leq \frac{1}{8}$. 
By Bretagnolle-Huber inequality, 
\[
   |\Pr_{\nu}(\sum_{i=1}^{n} W_i \geq \frac{n}{2}) - \Pr_{\nu'}(\sum_{i=1}^{n} W_i \geq \frac{n}{2})| \leq 1 - \frac{1}{2}\exp(-D_{\text{KL}}(\nu||\nu') ) \leq 1 - \frac{1}{2}\exp(-\frac{1}{8}) \leq \frac{2}{3}
\] 
This means that 
\begin{equation}
   \label{BH inequality}
   \Pr_{\nu}(\sum_{i=1}^{n} W_i \geq \frac{n}{2}) \geq \frac{1}{6} \quad \text{ or } \quad \Pr_{\nu'}(\sum_{i=1}^{n} (1 - W_i) \geq \frac{n}{2}) \geq \frac{1}{6}
\end{equation}
Recall that in \eqref{lower bound1}, we have bound the expected square error in terms of $ \sum_{i=1}^{n} W_i $.
By letting $ \rho $ large enough, we have
\begin{align*}
   &\E_{\nu}[(\widehat{\tau}(\mathcal{F}_n) - \tau)^2] \\
   \geq& \E_{\nu} [\frac{1}{1/\rho^2 + \sum_{j=1}^{i} W_j} + \frac{(1 + \epsilon)^2}{(1 + \epsilon)^2/\rho^2 + \sum_{j=1}^{i} (1 - W_j)}] \\
   \geq& \E_{\nu} [\frac{1}{\sum_{j=1}^{i} W_j} + \frac{(1 + \epsilon)^2}{\sum_{j=1}^{i} (1 - W_j)}] - \frac{1}{200n^2}\\
   =& \Pr_{\nu}(\sum_{i=1}^{n} W_i \geq \frac{n}{2}) \E_{\nu}[\frac{1}{\sum_{j=1}^{i} W_j} + \frac{(1 + \epsilon)^2}{\sum_{j=1}^{i} (1 - W_j) }| \sum_{i=1}^{n} W_i \geq \frac{n}{2}]\\
   &+ \Pr_{\nu}(\sum_{i=1}^{n} W_i < \frac{n}{2}) \E_{\nu}[\frac{1}{\sum_{j=1}^{i} W_j} + \frac{(1 + \epsilon)^2}{\sum_{j=1}^{i} (1 - W_j) }| \sum_{i=1}^{n} W_i < \frac{n}{2}] - \frac{1}{200n^2}\\
   \geq& \Pr_{\nu}(\sum_{i=1}^{n} W_i \geq \frac{n}{2})(\frac{1}{n/2} + \frac{(1+\epsilon)^2}{n/2}) + (1 - \Pr_{\nu}(\sum_{i=1}^{n} W_i \geq \frac{n}{2}))\frac{(2+\epsilon)^2}{n} - \frac{1}{200n^2} \\
   =& \frac{(2+\epsilon)^2}{n} + \Pr_{\nu}(\sum_{i=1}^{n} W_i \geq \frac{n}{2})\frac{\epsilon^2}{n} - \frac{1}{200n^2} \\
   =& \frac{(2+\epsilon)^2}{n} + \Pr_{\nu}(\sum_{i=1}^{n} W_i \geq \frac{n}{2})\frac{1}{16n^2} - \frac{1}{200n^2} 
\end{align*}
By a symmetry argument, we have 
\[
   \E_{\nu'}[(\widehat{\tau}(\mathcal{F}_n) - \tau)^2] \geq \frac{(2+\epsilon)^2}{n} + \Pr_{\nu'}(\sum_{i=1}^{n} (1 - W_i) \geq \frac{n}{2})\frac{1}{16n^2} - \frac{1}{200n^2} 
\] 
Hence \eqref{BH inequality} implies that 
\begin{align*}
   \max\{\E_{\nu}[(\widehat{\tau}(\mathcal{F}_n) - \tau)^2], \E_{\nu'}[(\widehat{\tau}(\mathcal{F}_n) - \tau)^2] \} \
   &\geq \frac{(2+\epsilon)^2}{n} + \frac{1}{96n^2} - \frac{1}{200n^2} \\
   &\geq \frac{(2+\epsilon)^2}{n} + \frac{1}{200n^2} \\
   &\geq \frac{(\sigma_{\nu}^{(1)} + \sigma_{\nu}^{(0)})^2}{n} + \frac{(\sigma_{\nu}^{(1)} + \sigma_{\nu}^{(0)})^2}{1000n^2} \\
   &= \frac{(\sigma_{\nu'}^{(1)} + \sigma_{\nu'}^{(0)})^2}{n} + \frac{(\sigma_{\nu'}^{(1)} + \sigma_{\nu'}^{(0)})^2}{1000n^2} 
\end{align*}
Finally, to convert the above lower bound in Bayes instances to the minimax lower bound, we make use of the inequality
\[
   \inf_{\widehat{\tau}} \sup_{\mu^{(1)}, \mu^{(0)}}\E_{\nu} [(\widehat{\tau} - \tau)^2 | \mu^{(1)}, \mu^{(0)} ] \geq \inf_{\widehat{\tau}} \E_{\nu}[(\widehat{\tau} - \tau)^2]
\]
and 
\[
   \inf_{\widehat{\tau}} \sup_{\mu^{(1)}, \mu^{(0)}}\E_{\nu'} [(\widehat{\tau} - \tau)^2 | \mu^{(1)}, \mu^{(0)} ] \geq \inf_{\widehat{\tau}} \E_{\nu'}[(\widehat{\tau} - \tau)^2]
\]  
Hence for any data generation policy and any $ \widehat{\tau} $ which is a measurable function of $ \mathcal{F}_n $, there exists $ \mu^{(1)}, \mu^{(0)}, \sigma^{(1)}, \sigma^{(0)} $ such that 
\[
   \E[(\widehat{\tau} - \tau)^2] \geq \frac{(\sigma^{(1)} + \sigma^{(0)})^2}{n} + \frac{(\sigma^{(1)} + \sigma^{(0)})^2}{1000n^2},
\] 
where the expectation is taken over a (non-Bayes) Gaussian instance with arms mean $ \mu^{(1)}, \mu^{(0)} $ and arms standard deviation $ \sigma^{(1)}, \sigma^{(0)}  $.

\subsection{Proof of Lemma \ref{lem:propensityscore-feature}}
   First note that 
\begin{equation}
   \label{eq:tau2X variance}
   \begin{aligned}
      &\E\left(\widehat{\tau}_2^X- \tau \right)^2
      = \E\left( \frac{1}{n} \sum_{i=1}^n \left(\mu^{(1)}(X_i)-\mu^{(0)}(X_i) - \tau + \frac{Y_i-\mu^{(1)}(X_i)}{e_i(X_i)}W_i -\frac{Y_i-\mu^{(0)}(X_i)}{1-e_i(X_i)}(1-W_i) \right) \right)^2 \\
      =& \frac{1}{n^2} \E[\sum_{i=1}^n \left(\mu^{(1)}(X_i)-\mu^{(0)}(X_i)- \tau\right)^2 + \sum_{i=1}^n \left(\frac{Y_i-\mu^{(1)}(X_i)}{e_i(X_i)}W_i - \frac{Y_i-\mu^{(0)}(X_i)}{1-e_i(X_i)}(1-W_i) \right)^2] \\
      &+ \frac{1}{n^2}\E[\sum_{i=1}^n\sum_{j=1}^n \left(\mu^{(1)}(X_i)-\mu^{(0)}(X_i)- \tau\right)\left(\frac{Y_j-\mu^{(1)}(X_j)}{e_j(X_j)}W_j - \frac{Y_j-\mu^{(0)}(X_j)}{1-e_j(X_j)}(1-W_j) \right)] \\
      &+ \frac{2}{n^2}\E[\sum_{1 \leq i < j \leq n} \left(\mu^{(1)}(X_i)-\mu^{(0)}(X_i)- \tau\right)\left(\mu^{(1)}(X_j)-\mu^{(0)}(X_j)- \tau\right)] \\
      &+ \frac{2}{n^2}\E[\sum_{1 \leq i < j \leq n} \left(\frac{Y_i-\mu^{(1)}(X_i)}{e_i(X_i)}W_i - \frac{Y_i-\mu^{(0)}(X_i)}{1-e_i(X_i)}(1-W_i) \right)\left(\frac{Y_j-\mu^{(1)}(X_j)}{e_j(X_j)}W_j - \frac{Y_j-\mu^{(0)}(X_j)}{1-e_j(X_j)}(1-W_j) \right)] 
   \end{aligned}
\end{equation}
   For any $ i \in [n] $, 
   \begin{align*}
      &\E\left(\frac{Y_i-\mu^{(1)}(X_i)}{e_i(X_i)}W_i - \frac{Y_i-\mu^{(0)}(X_i)}{1-e_i(X_i)}(1-W_i) \right) \\
      =& \E[\E[\left(\frac{Y_i-\mu^{(1)}(X_i)}{e_i(X_i)}W_i - \frac{Y_i-\mu^{(0)}(X_i)}{1-e_i(X_i)}(1-W_i)\right)^2 | \mathcal{F}_{i-1}, X_i]] \\
      =& \E[e_i(X_i)\E[\left(\frac{Y_i-\mu^{(1)}(X_i)}{e_i(X_i)}\right)^2 | \mathcal{F}_{i-1}, X_i] + (1 - e_i(X_i))\E[\left(\frac{Y_i-\mu^{(0)}(X_i)}{1-e_i(X_i)}\right)^2 | \mathcal{F}_{i-1}, X_i]] \\
      =& \E[\frac{(\sigma^{(1)}(X_i))^2}{e_i(X_i)} + \frac{(\sigma^{(0)}(X_i))^2}{1-e_i(X_i)}]
   \end{align*}
   For any $ 1 \leq i \leq j \leq n$, 
   \begin{align*}
      &\E \left(\mu^{(1)}(X_i)-\mu^{(0)}(X_i)- \tau\right)\left(\frac{Y_j-\mu^{(1)}(X_j)}{e_j(X_j)}W_j - \frac{Y_j-\mu^{(0)}(X_j)}{1-e_j(X_j)}(1-W_j) \right) \\
      =& \E[\E[\left(\mu^{(1)}(X_i)-\mu^{(0)}(X_i)- \tau\right)\left(\frac{Y_j-\mu^{(1)}(X_j)}{e_j(X_j)}W_j - \frac{Y_j-\mu^{(0)}(X_j)}{1-e_j(X_j)}(1-W_j) \right)| \mathcal{F}_{j-1}, X_j, W_j]] \\
      =& \E[\left(\mu^{(1)}(X_i)-\mu^{(0)}(X_i)- \tau\right)\E[\left(\frac{Y_j-\mu^{(1)}(X_j)}{e_j(X_j)}W_j - \frac{Y_j-\mu^{(0)}(X_j)}{1-e_j(X_j)}(1-W_j) \right)| \mathcal{F}_{j-1}, X_j, W_j]]  = 0
   \end{align*}
   For any $ 1 \leq j < i \leq n $ 
   \begin{align*}
      &\E \left(\mu^{(1)}(X_i)-\mu^{(0)}(X_i)- \tau\right)\left(\frac{Y_j-\mu^{(1)}(X_j)}{e_j(X_j)}W_j - \frac{Y_j-\mu^{(0)}(X_j)}{1-e_j(X_j)}(1-W_j) \right) \\
      =& \E[\E[\left(\mu^{(1)}(X_i)-\mu^{(0)}(X_i)- \tau\right)\left(\frac{Y_j-\mu^{(1)}(X_j)}{e_j(X_j)}W_j - \frac{Y_j-\mu^{(0)}(X_j)}{1-e_j(X_j)}(1-W_j) \right)| \mathcal{F}_{i-1}]] \\
      =& \E[\E[\left(\mu^{(1)}(X_i)-\mu^{(0)}(X_i)- \tau\right)| \mathcal{F}_{i-1}]\left(\frac{Y_j-\mu^{(1)}(X_j)}{e_j(X_j)}W_j - \frac{Y_j-\mu^{(0)}(X_j)}{1-e_j(X_j)}(1-W_j) \right)]  = 0
   \end{align*}
   For any $ 1 \leq i < j \leq n $, 
   \begin{align*}
      &\E[\left(\mu^{(1)}(X_i)-\mu^{(0)}(X_i)- \tau\right)\left(\mu^{(1)}(X_j)-\mu^{(0)}(X_j)- \tau\right)] \\
      =& \E[\E[\left(\mu^{(1)}(X_i)-\mu^{(0)}(X_i)- \tau\right)\left(\mu^{(1)}(X_j)-\mu^{(0)}(X_j)- \tau\right)| \mathcal{F}_{j-1} ]] \\
      =& \E[\left(\mu^{(1)}(X_i)-\mu^{(0)}(X_i)- \tau\right)\E[\left(\mu^{(1)}(X_j)-\mu^{(0)}(X_j)- \tau\right)| \mathcal{F}_{j-1} ]] = 0
   \end{align*}
   For any $ 1 \leq i < j \leq n $, 
   \begin{align*}
      &\E[\left(\frac{Y_i-\mu^{(1)}(X_i)}{e_i(X_i)}W_i - \frac{Y_i-\mu^{(0)}(X_i)}{1-e_i(X_i)}(1-W_i) \right)\left(\frac{Y_j-\mu^{(1)}(X_j)}{e_j(X_j)}W_j - \frac{Y_j-\mu^{(0)}(X_j)}{1-e_j(X_j)}(1-W_j) \right)] \\
      &\E[\E[\left(\frac{Y_i-\mu^{(1)}(X_i)}{e_i(X_i)}W_i - \frac{Y_i-\mu^{(0)}(X_i)}{1-e_i(X_i)}(1-W_i) \right)\left(\frac{Y_j-\mu^{(1)}(X_j)}{e_j(X_j)}W_j - \frac{Y_j-\mu^{(0)}(X_j)}{1-e_j(X_j)}(1-W_j) \right)| \mathcal{F}_{j-1}, X_j, W_j]] \\ 
      &\E[\left(\frac{Y_i-\mu^{(1)}(X_i)}{e_i(X_i)}W_i - \frac{Y_i-\mu^{(0)}(X_i)}{1-e_i(X_i)}(1-W_i) \right)\E[\left(\frac{Y_j-\mu^{(1)}(X_j)}{e_j(X_j)}W_j - \frac{Y_j-\mu^{(0)}(X_j)}{1-e_j(X_j)}(1-W_j) \right)| \mathcal{F}_{j-1}, X_j, W_j]] = 0
   \end{align*}
   Plugging everything into equation \ref{eq:tau2X variance}, we have
   \begin{align*}
      \E\left(\widehat{\tau}_2^X- \tau \right)^2
      =& \frac{1}{n^2}\E[\sum_{i=1}^n \left(\mu^{(1)}(X_i)-\mu^{(0)}(X_i)- \tau\right)^2] + \frac{1}{n^2}\E[\sum_{i=1}^n \frac{(\sigma^{(1)}(X_i))^2}{e_i(X_i)} + \frac{(\sigma^{(0)}(X_i))^2}{1-e_i(X_i)}]
   \end{align*}
   For the exactly same reason, we have
   \begin{align*}
      \E\left(\widehat{\tau}^{*X}- \tau \right)^2
      =& \frac{1}{n^2}\E[\sum_{i=1}^n \left(\mu^{(1)}(X_i)-\mu^{(0)}(X_i)- \tau\right)^2] + \frac{1}{n^2}\E[\sum_{i=1}^n\frac{(\sigma^{(1)}(X_i))^2}{e^*(X_i)} + \frac{(\sigma^{(0)}(X_i))^2}{1-e^*(X_i)}]
   \end{align*}
   The difference between the above two equations gives the result.
\subsection{Proof of Lemma \ref{lem:AIPW unbiasness} and Lemma \ref{lem:cross term is 0}}
\subsubsection{Proof of Lemma \ref{lem:AIPW unbiasness}}
   For any $ i \in G_1 $, condition on $ \mathcal{F}_{i-1}, \mathcal{F}_{G_2}, X_i $, we have that $ e_i(X_i), \widehat{\mu}^{G_2}(X_i), \mu(X_i) $ are fixed.
   Independence between $\mathcal{F}_{G_1}$ and $\mathcal{F}_{G_2}$ implies that $ \E[W_i | \mathcal{F}_{i-1}, \mathcal{F}_{G_2}, X_i] = e_i(X_i) $
   and $ \E[Y_i | \mathcal{F}_{i-1}, \mathcal{F}_{G_2}, X_i, W_i] = \mu^{(W_i)}(X_i)$.
   Hence, 
   \begin{align*}
      &\E[\widehat{\mu}^{G_2, (1)}(X_i)(1 - \frac{W_i}{e_i(X_i)})-\widehat{\mu}^{G_2, (0)}(X_i)(1 - \frac{1-W_i}{1-e_i(X_i)})] \\
      =& \E[[\widehat{\mu}^{G_2, (1)}(X_i)(1 - \frac{W_i}{e_i(X_i)})-\widehat{\mu}^{G_2, (0)}(X_i)(1 - \frac{1-W_i}{1-e_i(X_i)})| \mathcal{F}_{i-1}, \mathcal{F}_{G_2}, X_i]] \\
      =& \E[\widehat{\mu}^{G_2, (1)}(X_i)(1 - \frac{\E[W_i| \mathcal{F}_{i-1}, \mathcal{F}_{G_2}, X_i]}{e_i(X_i)})-\widehat{\mu}^{G_2, (0)}(X_i)(1 - \frac{\E[1-W_i| \mathcal{F}_{i-1}, \mathcal{F}_{G_2}, X_i]}{1-e_i(X_i)})] = 0\\
   \end{align*}
   \begin{align*}
      &\E[\frac{Y_i}{e_i(X_i)}W_i -\frac{Y_i}{1-e_i(X_i)}(1-W_i)] \\
      =& \E[[\frac{Y_i}{e_i(X_i)}W_i -\frac{Y_i}{1-e_i(X_i)}(1-W_i) | \mathcal{F}_{i-1}, \mathcal{F}_{G_2}, X_i]] \\
      =& \E[e_i(X_i)\E[\frac{Y_i}{e_i(X_i)}W_i | \mathcal{F}_{i-1}, \mathcal{F}_{G_2}, X_i, W_i = 1] - (1 - e_i(X_i))\E[\frac{Y_i}{1-e_i(X_i)}(1-W_i) | \mathcal{F}_{i-1}, \mathcal{F}_{G_2}, X_i, W_i = 0]] \\
      =& \E[\mu^{(1)}(X_i) - \mu^{(0)}(X_i)] = \tau
   \end{align*}
   By a symmetry argument, for any $ i \in G_2 $,
   \begin{align*}
      &\E[\widehat{\mu}^{G_1, (1)}(X_i)(1 - \frac{W_i}{e_i(X_i)})-\widehat{\mu}^{G_1, (0)}(X_i)(1 - \frac{1-W_i}{1-e_i(X_i)})] \\
      =& \E[[\widehat{\mu}^{G_1, (1)}(X_i)(1 - \frac{W_i}{e_i(X_i)})-\widehat{\mu}^{G_1, (0)}(X_i)(1 - \frac{1-W_i}{1-e_i(X_i)})| \mathcal{F}_{i-1}, \mathcal{F}_{G_1}, X_i]] = 0 \\
   \end{align*}
   \begin{align*}
      &\E[\frac{Y_i}{e_i(X_i)}W_i -\frac{Y_i}{1-e_i(X_i)}(1-W_i)] \\
      =& \E[[\frac{Y_i}{e_i(X_i)}W_i -\frac{Y_i}{1-e_i(X_i)}(1-W_i) | \mathcal{F}_{i-1}, \mathcal{F}_{G_1}, X_i]] = \tau
   \end{align*}
   Hence by linearity of expectation, $ \E[\widehat{\tau}_1^X] = \tau $.
\subsubsection{Proof of Lemma \ref{lem:cross term is 0}}
By the definition of $\widehat{\tau}_1^X, \widehat{\tau}_2^X$, we write
   \begin{align*}
      \widehat{\tau}_1^X- \widehat{\tau}_2^X &= 
      \frac{1}{n} \sum_{i \in G_1} \left(\widehat{\mu}^{G_2, (1)} (X_i)-\mu^{(1)}(X_i)\right)\left(1-\frac{W_i}{e_i(X_i)}\right)  
      - \frac{1}{n}\sum_{i \in G_1} \left(\widehat{\mu}^{G_2, (0)}(X_i)-\mu^{(0)}(X_i)\right)\left(1-\frac{1-W_i}{1-e_i(X_i)}\right)  \\
      &+ \frac{1}{n} \sum_{i \in G_2} \left(\widehat{\mu}^{G_1, (1)} (X_i)-\mu^{(1)}(X_i)\right)\left(1-\frac{W_i}{e_i(X_i)}\right)  
      - \frac{1}{n}\sum_{i \in G_2} \left(\widehat{\mu}^{G_1, (0)}(X_i)-\mu^{(0)}(X_i)\right)\left(1-\frac{1-W_i}{1-e_i(X_i)}\right)  
   \end{align*}
   \begin{align*}
      \widehat{\tau}_2^X-\tau = 
      \frac{1}{n} \sum_{i=1}^n \left(\mu^{(1)}(X_i) - \E[\mu^{(1)}(X)] - \mu^{(0)}(X_i)+ \E[\mu^{(0)}(X) ] + \frac{Y_i-\mu^{(1)}(X_i)}{e^*(X_i)}W_i -\frac{Y_i-\mu^{(0)}(X_i)}{1-e^*(X_i)}(1-W_i) \right),
   \end{align*}
         We analyze the cross terms case by case. \\
   First, by considering three cases, we prove that for any $ i \in G_1, j \in [n] $, 
   \begin{equation}
      \label{cross term1}
      \begin{aligned}
      &\E[\left((\widehat{\mu}^{G_2, (1)}(X_i) - \mu^{(1)}(X_t))(1 - \frac{W_i}{e_i(X_i)})-(\widehat{\mu}^{G_2, (0)}(X_i)-\mu^{(0)}(X_t))(1 - \frac{1-W_i}{1-e_i(X_i)})\right) \\
      &\times\left(\mu^{(1)}(X_j) - \E[\mu^{(1)}(X)] - \mu^{(0)}(X_j)+ \E[\mu^{(0)}(X) ] \right)] = 0
      \end{aligned}
   \end{equation}
   \textbf{Case 1.1:} For $ i, j \in G_1 $ and $ i < j $, condition on $ \mathcal{F}_{j-1}, \mathcal{F}_{G_2} $, we have that $ e_i(X_i), \widehat{\mu}^{G_2}(X_i), \mu(X_i), W_i $ are fixed. 
   Independence between $ X_j $ and $ (\mathcal{F}_{j-1} , \mathcal{F}_{G_2}) $ implies that 
   $ \E[\mu^{(1)}(X_j) - \E[\mu^{(1)}(X)] - \mu^{(0)}(X_j)+ \E[\mu^{(0)}(X) ] | \mathcal{F}_{i-1} , \mathcal{F}_{G_2}, X_i] = 0 $. 
   Hence, 
   \begin{align*}
      &\E[\left((\widehat{\mu}^{G_2, (1)}(X_i) - \mu^{(1)}(X_t))(1 - \frac{W_i}{e_i(X_i)})-(\widehat{\mu}^{G_2, (0)}(X_i)-\mu^{(0)}(X_t))(1 - \frac{1-W_i}{1-e_i(X_i)})\right) \\
      &\times\left(\mu^{(1)}(X_j) - \E[\mu^{(1)}(X)] - \mu^{(0)}(X_j)+ \E[\mu^{(0)}(X) ] \right)] \\
      =&\E[[\left((\widehat{\mu}^{G_2, (1)}(X_i) - \mu^{(1)}(X_t))(1 - \frac{W_i}{e_i(X_i)})-(\widehat{\mu}^{G_2, (0)}(X_i)-\mu^{(0)}(X_t))(1 - \frac{1-W_i}{1-e_i(X_i)})\right) \\
      &\times\left(\mu^{(1)}(X_j) - \E[\mu^{(1)}(X)] - \mu^{(0)}(X_j)+ \E[\mu^{(0)}(X) ] \right)| \mathcal{F}_{j-1}, \mathcal{F}_{G_2}]] = 0 
   \end{align*}
   \textbf{Case 1.2:} For $ i, j \in G_1 $ and $ i \geq j $, condition on $ \mathcal{F}_{i-1}, \mathcal{F}_{G_2}, X_i $, we have that $ e_i(X_i), \widehat{\mu}^{G_2}(X_i), \mu(X_i), \mu(X_j) $ are fixed. 
   Independence between $\mathcal{F}_{G_1}$ and $\mathcal{F}_{G_2}$ implies that $ \E[W_i | \mathcal{F}_{i-1}, \mathcal{F}_{G_2}, X_i] = e_i(X_i) $.
   Hence,
   \begin{align*}
      &\E[\left((\widehat{\mu}^{G_2, (1)}(X_i) - \mu^{(1)}(X_t))(1 - \frac{W_i}{e_i(X_i)})-(\widehat{\mu}^{G_2, (0)}(X_i)-\mu^{(0)}(X_t))(1 - \frac{1-W_i}{1-e_i(X_i)})\right) \\
      &\times\left(\mu^{(1)}(X_j) - \E[\mu^{(1)}(X)] - \mu^{(0)}(X_j)+ \E[\mu^{(0)}(X) ] \right)] \\
      =&\E[[\left((\widehat{\mu}^{G_2, (1)}(X_i) - \mu^{(1)}(X_t))(1 - \frac{W_i}{e_i(X_i)})-(\widehat{\mu}^{G_2, (0)}(X_i)-\mu^{(0)}(X_t))(1 - \frac{1-W_i}{1-e_i(X_i)})\right) \\
      &\times\left(\mu^{(1)}(X_j) - \E[\mu^{(1)}(X)] - \mu^{(0)}(X_j)+ \E[\mu^{(0)}(X) ] \right)| \mathcal{F}_{i-1}, \mathcal{F}_{G_2}, X_i]] \\
      =&\E[\left((\widehat{\mu}^{G_2, (1)}(X_i) - \mu^{(1)}(X_t))(1 - \frac{\E[W_i | \mathcal{F}_{i-1}, \mathcal{F}_{G_2}, X_i]}{e_i(X_i)})-(\widehat{\mu}^{G_2, (0)}(X_i)-\mu^{(0)}(X_t))(1 - \frac{\E[1 - W_i | \mathcal{F}_{i-1}, \mathcal{F}_{G_2}, X_i]}{1-e_i(X_i)})\right) \\
      &\times\left(\mu^{(1)}(X_j) - \E[\mu^{(1)}(X)] - \mu^{(0)}(X_j)+ \E[\mu^{(0)}(X) ] \right)] = 0
   \end{align*}
   \textbf{Case 1.3:} For $ i \in G_1, j \in G_2 $, condition on $ \mathcal{F}_{i-1}, \mathcal{F}_{G_2}, X_i $, we have that $ e_i(X_i), \widehat{\mu}^{G_2}(X_i), \mu(X_i), \mu(X_j) $ are fixed. 
   Independence between $\mathcal{F}_{G_1}$ and $\mathcal{F}_{G_2}$ implies that $ \E[W_i | \mathcal{F}_{i-1}, \mathcal{F}_{G_2}, X_i] = e_i(X_i) $.
   Hence, 
   \begin{align*}
      &\E[\left((\widehat{\mu}^{G_2, (1)}(X_i) - \mu^{(1)}(X_t))(1 - \frac{W_i}{e_i(X_i)})-(\widehat{\mu}^{G_2, (0)}(X_i)-\mu^{(0)}(X_t))(1 - \frac{1-W_i}{1-e_i(X_i)})\right) \\
      &\times\left(\mu^{(1)}(X_j) - \E[\mu^{(1)}(X)] - \mu^{(0)}(X_j)+ \E[\mu^{(0)}(X) ] \right)] \\
      =&\E[[\left((\widehat{\mu}^{G_2, (1)}(X_i) - \mu^{(1)}(X_t))(1 - \frac{W_i}{e_i(X_i)})-(\widehat{\mu}^{G_2, (0)}(X_i)-\mu^{(0)}(X_t))(1 - \frac{1-W_i}{1-e_i(X_i)})\right) \\
      &\times\left(\mu^{(1)}(X_j) - \E[\mu^{(1)}(X)] - \mu^{(0)}(X_j)+ \E[\mu^{(0)}(X) ] \right)| \mathcal{F}_{i-1}, \mathcal{F}_{G_2}, X_i]] \\
      =&\E[\left((\widehat{\mu}^{G_2, (1)}(X_i) - \mu^{(1)}(X_t))(1 - \frac{\E[W_i | \mathcal{F}_{i-1}, \mathcal{F}_{G_2}, X_i]}{e_i(X_i)})-(\widehat{\mu}^{G_2, (0)}(X_i)-\mu^{(0)}(X_t))(1 - \frac{\E[1 - W_i | \mathcal{F}_{i-1}, \mathcal{F}_{G_2}, X_i]}{1-e_i(X_i)})\right) \\
      &\times\left(\mu^{(1)}(X_j) - \E[\mu^{(1)}(X)] - \mu^{(0)}(X_j)+ \E[\mu^{(0)}(X) ] \right)] = 0
   \end{align*}

   Next, by considering three cases, we prove that for any $ i \in G_1, j \in [n] $, 
   \begin{equation}
      \label{cross term2}
      \begin{aligned}
      &\E[\left((\widehat{\mu}^{G_2, (1)}(X_i) - \mu^{(1)}(X_t))(1 - \frac{W_i}{e_i(X_i)})-(\widehat{\mu}^{G_2, (0)}(X_i)-\mu^{(0)}(X_t))(1 - \frac{1-W_i}{1-e_i(X_i)})\right) \\
      &\times\left(\frac{Y_j-\mu^{(1)}(X_j)}{e^*(X_j)}W_j -\frac{Y_j-\mu^{(0)}(X_j)}{1-e^*(X_j)}(1-W_j) \right)] = 0
      \end{aligned}
   \end{equation}
   \textbf{Case 2.1:} For $ i, j \in G_1 $ and $ i > j $, condition on $ \mathcal{F}_{i-1}, \mathcal{F}_{G_2}, X_i $, we have that $ e_i(X_i), \widehat{\mu}^{G_2}(X_i), \mu(X_i), Y_j, X_j, W_j $ are fixed. 
   Independence between $\mathcal{F}_{G_1}$ and $\mathcal{F}_{G_2}$ implies that $ \E[W_i | \mathcal{F}_{i-1}, \mathcal{F}_{G_2}, X_i] = e_i(X_i) $.
   Hence, 
   \begin{align*}
      &\E[\left((\widehat{\mu}^{G_2, (1)}(X_i) - \mu^{(1)}(X_t))(1 - \frac{W_i}{e_i(X_i)})-(\widehat{\mu}^{G_2, (0)}(X_i)-\mu^{(0)}(X_t))(1 - \frac{1-W_i}{1-e_i(X_i)})\right) \\
      &\times\left(\frac{Y_j-\mu^{(1)}(X_j)}{e^*(X_j)}W_j -\frac{Y_j-\mu^{(0)}(X_j)}{1-e^*(X_j)}(1-W_j) \right)] \\
      =&\E[[\left((\widehat{\mu}^{G_2, (1)}(X_i) - \mu^{(1)}(X_t))(1 - \frac{W_i}{e_i(X_i)})-(\widehat{\mu}^{G_2, (0)}(X_i)-\mu^{(0)}(X_t))(1 - \frac{1-W_i}{1-e_i(X_i)})\right) \\
      &\times\left(\frac{Y_j-\mu^{(1)}(X_j)}{e^*(X_j)}W_j -\frac{Y_j-\mu^{(0)}(X_j)}{1-e^*(X_j)}(1-W_j) \right)| \mathcal{F}_{i-1}, \mathcal{F}_{G_2}, X_i]] \\
      =&\E[\left((\widehat{\mu}^{G_2, (1)}(X_i) - \mu^{(1)}(X_t))(1 - \frac{\E[W_i | \mathcal{F}_{i-1}, \mathcal{F}_{G_2}, X_i]}{e_i(X_i)})-(\widehat{\mu}^{G_2, (0)}(X_i)-\mu^{(0)}(X_t))(1 - \frac{\E[1 - W_i | \mathcal{F}_{i-1}, \mathcal{F}_{G_2}, X_i]}{1-e_i(X_i)})\right) \\
      &\times\left(\frac{Y_j-\mu^{(1)}(X_j)}{e^*(X_j)}W_j -\frac{Y_j-\mu^{(0)}(X_j)}{1-e^*(X_j)}(1-W_j) \right)] = 0
   \end{align*}
   \textbf{Case 2.2:} For $ i, j \in G_1 $ and $ i \leq j $, condition on $ \mathcal{F}_{j-1}, \mathcal{F}_{G_2}, X_j, W_j $, we have that $ e_i(X_i), \widehat{\mu}^{G_2}(X_i), \mu(X_i), W_i $ are fixed. 
   Independence between $\mathcal{F}_{G_1}$ and $\mathcal{F}_{G_2}$ implies that $ \E[Y_j | \mathcal{F}_{j-1}, \mathcal{F}_{G_2}, X_j, W_j] = \mu^{(W_j)}(X_j) $.
   Hence, 
   \begin{align*}
      &\E[\left((\widehat{\mu}^{G_2, (1)}(X_i) - \mu^{(1)}(X_t))(1 - \frac{W_i}{e_i(X_i)})-(\widehat{\mu}^{G_2, (0)}(X_i)-\mu^{(0)}(X_t))(1 - \frac{1-W_i}{1-e_i(X_i)})\right) \\
      &\times\left(\frac{Y_j-\mu^{(1)}(X_j)}{e^*(X_j)}W_j -\frac{Y_j-\mu^{(0)}(X_j)}{1-e^*(X_j)}(1-W_j) \right)] \\
      =&\E[[\left((\widehat{\mu}^{G_2, (1)}(X_i) - \mu^{(1)}(X_t))(1 - \frac{W_i}{e_i(X_i)})-(\widehat{\mu}^{G_2, (0)}(X_i)-\mu^{(0)}(X_t))(1 - \frac{1-W_i}{1-e_i(X_i)})\right) \\
      &\times\left(\frac{Y_j-\mu^{(1)}(X_j)}{e^*(X_j)}W_j -\frac{Y_j-\mu^{(0)}(X_j)}{1-e^*(X_j)}(1-W_j) \right)| \mathcal{F}_{j-1}, \mathcal{F}_{G_2}, X_j, W_j]] \\
      =&\E[\left((\widehat{\mu}^{G_2, (1)}(X_i) - \mu^{(1)}(X_t))(1 - \frac{W_i}{e_i(X_i)})-(\widehat{\mu}^{G_2, (0)}(X_i)-\mu^{(0)}(X_t))(1 - \frac{1-W_i}{1-e_i(X_i)})\right) \\
      &\times\left(\frac{\mu^{(W_j)}(X_j)-\mu^{(1)}(X_j)}{e^*(X_j)}W_j -\frac{\mu^{(W_j)}(X_j) -\mu^{(0)}(X_j)}{1-e^*(X_j)}(1-W_j) \right)] = 0
   \end{align*}
   
   \textbf{Case 2.3:} For $ i \in G_1, j \in G_2 $, condition on $ \mathcal{F}_{i-1}, \mathcal{F}_{G_2}, X_i $, we have that $ e_i(X_i), \widehat{\mu}^{G_2}(X_i), \mu(X_i), X_j, Y_j, W_j $ are fixed. 
   Independence between $\mathcal{F}_{G_1}$ and $\mathcal{F}_{G_2}$ implies that $ \E[W_i | \mathcal{F}_{i-1}, \mathcal{F}_{G_2}, X_i] = e_i(X_i) $.
   Hence, 
   \begin{align*}
      &\E[\left((\widehat{\mu}^{G_2, (1)}(X_i) - \mu^{(1)}(X_t))(1 - \frac{W_i}{e_i(X_i)})-(\widehat{\mu}^{G_2, (0)}(X_i)-\mu^{(0)}(X_t))(1 - \frac{1-W_i}{1-e_i(X_i)})\right) \\
      &\times\left(\frac{Y_j-\mu^{(1)}(X_j)}{e^*(X_j)}W_j -\frac{Y_j-\mu^{(0)}(X_j)}{1-e^*(X_j)}(1-W_j) \right)] \\
      =&\E[[\left((\widehat{\mu}^{G_2, (1)}(X_i) - \mu^{(1)}(X_t))(1 - \frac{W_i}{e_i(X_i)})-(\widehat{\mu}^{G_2, (0)}(X_i)-\mu^{(0)}(X_t))(1 - \frac{1-W_i}{1-e_i(X_i)})\right) \\
      &\times\left(\frac{Y_j-\mu^{(1)}(X_j)}{e^*(X_j)}W_j -\frac{Y_j-\mu^{(0)}(X_j)}{1-e^*(X_j)}(1-W_j) \right)| \mathcal{F}_{i-1}, \mathcal{F}_{G_2}, X_i]] \\
      =&\E[\left((\widehat{\mu}^{G_2, (1)}(X_i) - \mu^{(1)}(X_t))(1 - \frac{\E[W_i | \mathcal{F}_{i-1}, \mathcal{F}_{G_2}, X_i]}{e_i(X_i)})-(\widehat{\mu}^{G_2, (0)}(X_i)-\mu^{(0)}(X_t))(1 - \frac{\E[1 - W_i | \mathcal{F}_{i-1}, \mathcal{F}_{G_2}, X_i]}{1-e_i(X_i)})\right) \\
      &\times\left(\frac{Y_j-\mu^{(1)}(X_j)}{e^*(X_j)}W_j -\frac{Y_j-\mu^{(0)}(X_j)}{1-e^*(X_j)}(1-W_j) \right)] = 0
   \end{align*}
   Finally, by a symmetry argument, we have the counterpart of equation \ref{cross term1} and \ref{cross term2}, i.e.,
   for any $ i \in G_2, j \in [n] $, 
   \begin{equation}
      \label{cross term3}
      \begin{aligned}
      &\E[\left((\widehat{\mu}^{G_1, (1)}(X_i) - \mu^{(1)}(X_t))(1 - \frac{W_i}{e_i(X_i)})-(\widehat{\mu}^{G_1, (0)}(X_i)-\mu^{(0)}(X_t))(1 - \frac{1-W_i}{1-e_i(X_i)})\right) \\
      &\times\left(\mu^{(1)}(X_j) - \E[\mu^{(1)}(X)] - \mu^{(0)}(X_j)+ \E[\mu^{(0)}(X) ] \right)] = 0
      \end{aligned}
   \end{equation}
   \begin{equation}
      \label{cross term4}
      \begin{aligned}
      &\E[\left((\widehat{\mu}^{G_1, (1)}(X_i) - \mu^{(1)}(X_t))(1 - \frac{W_i}{e_i(X_i)})-(\widehat{\mu}^{G_1, (0)}(X_i)-\mu^{(0)}(X_t))(1 - \frac{1-W_i}{1-e_i(X_i)})\right) \\
      &\times\left(\frac{Y_j-\mu^{(1)}(X_j)}{e^*(X_j)}W_j -\frac{Y_j-\mu^{(0)}(X_j)}{1-e^*(X_j)}(1-W_j) \right)] = 0
      \end{aligned}
   \end{equation}
   By equation \ref{cross term1}, \ref{cross term2}, \ref{cross term3}, \ref{cross term4} and linearity of expectation, we have proved Lemma \ref{lem:cross term is 0}.
\subsection{Proof of Lemma \ref{lem:model estimation} and Lemma \ref{lem:propensity optimization regret-feature}}
 Our proofs rely on a clean event analysis. Intuitively, we show that with high probability, $ \widehat{\sigma}_k $ and $ \widehat{\mu} $ in algorithm \Cref{alg:AIPW data collecting} have good estimation precision guarantee.
 In the case of all clean events happening, we bound $ \E \left( \widehat{\tau}_1^X- \widehat{\tau}_2^X\right)^2 $ 
 and $ \E \left(\widehat{\tau}_2^X- \tau \right)^2-\mathbb{E} \left( \widehat{\tau}^{*X}- \tau \right)^2 $ by the $ \widehat{\sigma}_k $ and $ \widehat{\mu} $ estimation precision guarantee.
 Otherwise, by the assumption that $ -M \leq \tau \leq M$, $ (\widehat{\tau}_2^X - \tau)^2 $ are upper bounded by a constant and won't contribute much error in expectation because clean events happen with high probability.

 For arm $ w \in \{0, 1\} $, batch $ k \in [K] $ we define events as follows:
 \begin{align*}
   &\mathcal{E}_{k}^{G_1, (w)} := \big\{ \E_{X \sim P_X} [(\sigma^{(w)}(X)  - \widehat{\sigma}_{k}^{G_1, (w)}(X) )^2] \leq C_5(\inf_{x \in \mathcal{X}}\frac{\widetilde{\sigma}_{k-1}^{G_1, (w)}(x) }{\widetilde{\sigma}_{k-1} ^{G_1, (1)}(x) + \widetilde{\sigma}_{k-1} ^{G_1, (0)}(x)}\gamma_k/4)^{-\beta}(\log n)(\sup_{x \in \mathcal{X}} \sigma^{(w)}(x))^2 \}\\ 
   &\mathcal{E}_{k}^{G_2, (w)} := \big\{ \E_{X \sim P_X} [(\sigma^{(w)}(X)  - \widehat{\sigma}_{k}^{G_1, (w)}(X) )^2] \leq C_5(\inf_{x \in \mathcal{X}}\frac{\widetilde{\sigma}_{k-1}^{G_1, (w)}(x) }{\widetilde{\sigma}_{k-1} ^{G_1, (1)}(x) + \widetilde{\sigma}_{k-1} ^{G_1, (0)}(x)}\gamma_k/4)^{-\beta}(\log n)(\sup_{x \in \mathcal{X}} \sigma^{(w)}(x))^2 \}\\ 
   &\mathcal{E}_{\mu}^{G_1, (w)} := \{\E_{X \sim P_X} [(\widehat{\mu}^{G_1, (w)}(X) - \mu^{(w)}(X))^2] \leq C_4(\inf_{x \in \mathcal{X}}\frac{\widetilde{\sigma}_{K-1}^{G_1, (w)}(x) }{\widetilde{\sigma}_{K-1} ^{G_1, (1)}(x) + \widetilde{\sigma}_{K-1} ^{G_1, (0)}(x)}\gamma_K/4)^{-\alpha}(\log n)(\sup_{x \in \mathcal{X}} \sigma^{(w)}(x))^2 \}\\ 
   &\mathcal{E}_{\mu}^{G_2, (w)} := \{\E_{X \sim P_X} [(\widehat{\mu}^{G_2, (w)}(X) - \mu^{(w)}(X))^2] \leq C_4(\inf_{x \in \mathcal{X}}\frac{\widetilde{\sigma}_{K-1}^{G_1, (w)}(x) }{\widetilde{\sigma}_{K-1} ^{G_1, (1)}(x) + \widetilde{\sigma}_{K-1} ^{G_1, (0)}(x)}\gamma_K/4)^{-\alpha}(\log n)(\sup_{x \in \mathcal{X}} \sigma^{(w)}(x))^2\} \\ 
 \end{align*}
 We prove that $ \mathcal{E} := \bigcap_{w \in \{0, 1\}k \in [K-1]}(\mathcal{E}_k^{G_1, (w)} \cap \mathcal{E}_{k}^{G_2, (w)}) \bigcap_{w \in \{0, 1\}} (\mathcal{E}_{\mu}^{G_1, (w)} \cap \mathcal{E}_{\mu}^{G_2, (w)} ) $ happens with probability more than $ 1 - \frac{1}{n^2} $ 
 if $ n $ large enough. 
\subsubsection{Proof of the union event happening with high probability}
For any batch $ k \in [K] $, there are $ \gamma_k / 2 $ periods in group $G_1$.
To utilize the variance estimation guarantee in assumption \ref{asm:oracle with convariate} to get $ \widehat{\sigma}_k^{G_1, (1)} $, we need to select data in a way that covariates follow distribution $ P_X $. 
For any $ i $ in batch $ k $ belonging to group1 such that $ W_i = 1 $, we select data $ (X_i, Y_i) $ with probability $ \frac{\min_{x \in \mathcal{X}} e_i(x)}{e_i(X_i)} $. 
Hence we have selected $ m $ i.i.d data$ (X, Y) $ for arm1 where $ m \sim Bin(\gamma_k / 2, \inf_{x \in \mathcal{X}} \frac{\widetilde{\sigma}_{k-1}^{G_1, (1)} (x)}{\widetilde{\sigma}_{k-1}^{G_1, (1)} (x) + \widetilde{\sigma}_{k-1}^{G_1, (0)} (x)}) $.
Now we need the following result of Binomial distribution.
\begin{lemma}
   \label{lem:binomial light tail}
   $ X $ follows an arbitrary binomial distribution, we have
   \[
      \Pr[X \leq \frac{1}{2}\E[X]] \leq \exp(-\E[X]/8), 
   \] 
\end{lemma}
Since $ n $ large enough, we can assume $ \frac{n}{a\log n} > \frac{n}{K\log n} > 1 $ and $ (\frac{n}{a\log n})^{1/S_K} > (\frac{n}{K\log n})^{1/S_K} > \max\{128, 8C_3\} $. 
Hence we have $ \gamma_k \geq \gamma_1 \geq \max\{128, 8C_3\}\log n$. 
Consequently, 
\begin{align*}
   \E[m] &= (\gamma_k/2)\inf_{x \in \mathcal{X}} \frac{\widetilde{\sigma}_{k-1}^{G_1, (1)} (x)}{\widetilde{\sigma}_{k-1}^{G_1, (1)} (x) + \widetilde{\sigma}_{k-1}^{G_1, (0)} (x)}
         \geq  (\gamma_k / 2)\frac{1}{1 + \gamma_k/\gamma_1} \geq \gamma_1/4 \geq \max\{32, 2C_3\}\log n
\end{align*}
By lemma \ref{lem:binomial light tail}, with probability $ 1 - \frac{1}{n^4} $, 
$ m \geq \E[m] / 2 = \inf_{x \in \mathcal{X}}\frac{\widetilde{\sigma}_{k-1}^{G_1, (w)}(x) }{\widetilde{\sigma}_{k-1} ^{G_1, (1)}(x) + \widetilde{\sigma}_{k-1} ^{G_1, (0)}(x)}\gamma_k/4 \geq C_3\log n$.\\
Using these $ m $ collected data and the variance estimation oracle in assumption \ref{asm:oracle with convariate}, condition on $ m \geq \E[m]/2 $, with another probability $ 1 - \frac{1}{n^4} $, 
$ \mathcal{E}_k^{G_1, (1)} $ happens. By a union bound, we have 
\[
   \Pr[\mathcal{E}_k^{G_1, (1)}] > 1 - \frac{2}{n^4}
\] 
By a symmetry argument, for any $ k \in [K-1], w \in \{0, 1\} $ 
\[
   \Pr[\mathcal{E}_k^{G_1, (w)}] > 1 - \frac{2}{n^4},
   \Pr[\mathcal{E}_k^{G_2, (w)}] > 1 - \frac{2}{n^4}
\]
With the same argument in the last batch, but using mean estimation oracle in assumption \ref{asm:oracle with convariate}, we have 
\[
   \Pr[\mathcal{E}_{\mu}^{G_1, (w)}] > 1 - \frac{2}{n^4},
   \Pr[\mathcal{E}_{\mu}^{G_2, (w)}] > 1 - \frac{2}{n^4}
\] 
By the union bound, 
\[
   \Pr[\mathcal{E}] > 1 - \frac{8K}{n^4} > 1 - \frac{1}{n^2}
\] 
\subsubsection{Some additional results}
Before our proof of lemma \ref{lem:model estimation} and lemma \ref{lem:propensity optimization regret-feature},
we still need to prove some useful results.
Let the diameter of $ \mathcal{X} $ be $ d_{X} := \sup_{x, x' \in \mathcal{X}} \|x - x'\| $. 
Note that $ \sigma^{(1)}(x), \sigma^{(0)}(x) $ are $ L-$Lipschitz, so we have $ \forall w \in \{0, 1\}, \forall x, x' \in \mathcal{X}, |\sigma^{(w)}(x) - \sigma^{(w)}(x')| \leq L d_X $.
Combined with the assumption that $ \sigma^{(1)}(x), \sigma^{(0)}(x) $ are lower bounded by $ \sigma $, 
there exist a constant $ C' $ depend on $L, d_X, \sigma $ such that 
\[
   \forall w \in \{0, 1\}, \frac{\sup_{x \in \mathcal{X}} \sigma^{(w)}(x)}{\inf_{x \in \mathcal{X}} \sigma^{(w)}(x)} \leq C'
\] 
If $ \mathcal{E} $ happens, with the assumption that $ n $ is large enough,
we can show that there exists some constant $H$ depends on $\beta$, $K$, $L, d_X, \sigma $ such that
\[
   \forall k \in [K-1], \forall x \in \mathcal{X}, \forall w \in \{0, 1\}, \sigma^{(w)}(x) / H \leq \widehat{\sigma}_k^{G_1, (w)}(x), \widehat{\sigma}_k^{G_2, (w)}(x) \leq H\sigma^{(w)}(x) 
\] 
For the sake of simplicity, we will abbreviate the supercript of $G_1$, $G_2$ since they hold uniformly given that $ \mathcal{E} $ happens.
First, consider the treatment group. By the statistical guarantee of the oracle, we have 
\[
 \E_{X \sim P_X} [(\sigma^{(1)}(X)  - \widehat{\sigma}^{(1)}_{k}(X) )^2] \leq C_5(\inf_{x \in \mathcal{X}}\frac{\widetilde{\sigma}^{(1)}_{k-1}(x) }{\widetilde{\sigma}^{(1)}_{k-1} (x) + \widetilde{\sigma}_{k-1} ^{ (0)}(x)}\gamma_k/4)^{-\beta}(\log n)(\sup_{x \in \mathcal{X}} \sigma^{(1)}(x))^2 
\]
Since our exploration threshold guarantees that
\[
(\inf_{x \in \mathcal{X}}\frac{\widetilde{\sigma}^{(1)}_{k-1}(x) }{\widetilde{\sigma}^{(1)}_{k-1} (x) + \widetilde{\sigma}_{k-1} ^{ (0)}(x)}\gamma_k/4)^{-\beta}\le (\gamma_1/8)^{-\beta} \le  O(n^{-\frac{\beta}{S_K}}),
\]
For $n$ sufficiently large, we will have some $x_0 \in \mathcal{X}$ such that for some small $\varepsilon$,
\begin{equation}\label{equ-constant-variance}
    (\sigma^{(1)}(x_0)  - \widehat{\sigma}^{(1)}_k(x_0) )^2 \le \varepsilon (\sup_{x \in \mathcal{X}} \sigma^{(1)}(x))^2.
\end{equation}
Since we know that both $\sigma^{(1)}(X)$ and $\widehat{\sigma}^{(1)}_k(X)$ are Lipschitz continuous, we know that 
there exists constant $H_1$,$H_2$ such that 
\[ H_1\sup_{x \in \mathcal{X}} \sigma^{(1)}(x)\le \sigma^{(1)}(x) \le H_2 \inf_{x \in \mathcal{X}} \sigma^{(1)}(x) \]
\[
H_1\sup_{x \in \mathcal{X}}\widehat{\sigma}^{(1)}(x)\le \widehat{\sigma}^{(1)}(x) \le H_2 \inf_{x \in \mathcal{X}}\widehat{\sigma}^{(1)}(x)
\]
By (\ref{equ-constant-variance}), we also have that for some $H_3$,
\[
 \widehat{\sigma}^{(1)}_k(x_0) /H_3 \le \sigma^{(1)}(x_0) \le H_3\widehat{\sigma}^{(1)}_k(x_0)
\]
Combining the above inequalities, we can prove that
\[
 \forall x\in \mathcal{X}
 \qquad \widehat{\sigma}^{(1)}_k(x) /H \le \sigma^{(1)}(x) \le H\widehat{\sigma}^{(1)}_k(x)
\]
for some constant $H$. The proof for the control group is similar.
And as a result, for any $ i \in [n] $ and $ x \in \mathcal{X} $, 
\begin{equation}
   \label{eq:propensity ratio bound}   
   \frac{1 - e_i(x)}{e_i(x)} = 
   \frac{\widetilde{\sigma}_{k-1}^{G_1, (0)}(x) }{\widetilde{\sigma}_{k-1} ^{G_1, (1)}(x)}
   \leq 1 + \frac{\widehat{\sigma}_{k-1}^{G_1, (0)}(x) }{\widehat{\sigma}_{k-1} ^{G_1, (1)}(x)}
   \leq 1 + H^2\frac{\sigma^{(0)}(x)}{\sigma^{(1)}(x)}
\end{equation}
\subsubsection{Proof of lemma \ref{lem:model estimation}}
\begin{equation}
    \label{proof:eq:model estimation, Cauchy}
   \begin{aligned}
      \mathbb{E} \left( \widehat{\tau}_1^X- \widehat{\tau}_2^X\right)^2
     \le& \frac{4}{n^2}
     \left(\mathbb{E}\left(\sum_{i \in G_1} \left(1-\frac{W_i}{e_i(X_i)}\right) \left(\widehat{\mu}^{G_2, (1)} (X_i)-\mu^{(1)}(X_i)\right)\right)^2\right.\\
     &+\left. \mathbb{E}\left(\sum_{i \in G_2} \left(1-\frac{W_i}{e_i(X_i)}\right) \left(\widehat{\mu}^{G_1, (1)} (X_i)-\mu^{(1)}(X_i)\right)\right)^2\right.\\
     &+ \left. \mathbb{E}\left(\sum_{i \in G_1} \left(1-\frac{1-W_i}{1-e_i(X_i)}\right) \left(\widehat{\mu}^{G_2, (0)}(X_i)-\mu^{(0)}(X_i)\right)\right)^2\right.\\
     &+ \left. \mathbb{E}\left(\sum_{i \in G_2} \left(1-\frac{1-W_i}{1-e_i(X_i)}\right) \left(\widehat{\mu}^{G_1, (0)}(X_i)-\mu^{(0)}(X_i)\right)\right)^2\right)
    \end{aligned}
\end{equation}
   We analyze equation \ref{proof:eq:model estimation, Cauchy} term by term. 
   \begin{equation}
   \label{eq:regret1, with X, 1}
   \begin{aligned}
      &\E \left( \sum_{i \in G_1} \left(1-\frac{W_i}{e_i(X_i)}\right) \left(\widehat{\mu}^{G_2, (1)} (X_i)-\mu^{(1)}(X_i)\right)\right)^2 \\
      =& \E[\sum_{i \in G_1} \left(1-\frac{W_i}{e_i(X_i)}\right)^2\left(\widehat{\mu}^{G_2, (1)} (X_i)-\mu^{(1)}(X_i)\right)^2 \\
      &+ 2\sum_{i, j \in G_1, i < j} \left(1-\frac{W_i}{e_i(X_i)}\right)\left(1-\frac{W_j}{e_j(X_j)}\right)\left(\widehat{\mu}^{G_2, (1)} (X_i)-\mu^{(1)}(X_i)\right)\left(\widehat{\mu}^{G_2, (1)} (X_j)-\mu^{(1)}(X_j)\right)]
   \end{aligned}
   \end{equation}
For any $ i \in G_1 $, 
\begin{align*}
   &\E[\left(1-\frac{W_i}{e_i(X_i)}\right)^2\left(\widehat{\mu}^{G_2, (1)} (X_i)-\mu^{(1)}(X_i)\right)^2] \\
   =& \E[\E[\left(1-\frac{W_i}{e_i(X_i)}\right)^2\left(\widehat{\mu}^{G_2, (1)} (X_i)-\mu^{(1)}(X_i)\right)^2 | \mathcal{F}_{G_2}, \mathcal{F}_{i-1}, X_i]] \\
   =& \E[\E[\left(1-\frac{W_i}{e_i(X_i)}\right)^2| \mathcal{F}_{G_2}, \mathcal{F}_{i-1}, X_i] \left(\widehat{\mu}^{G_2, (1)} (X_i)-\mu^{(1)}(X_i)\right)^2 ] \\
   =& \E[\frac{1 - e_i(X_i)}{e_i(X_i)} \left(\widehat{\mu}^{G_2, (1)} (X_i)-\mu^{(1)}(X_i)\right)^2 ] \\
\end{align*}
For any $ i, j \in G_1, i < j $, 
\begin{align*}
   &\E [\left(1-\frac{W_i}{e_i(X_i)}\right)\left(1-\frac{W_j}{e_j(X_j)}\right)\left(\widehat{\mu}^{G_2, (1)} (X_i)-\mu^{(1)}(X_i)\right)\left(\widehat{\mu}^{G_2, (1)} (X_j)-\mu^{(1)}(X_j)\right)] \\
   =& \E [\left(1-\frac{W_i}{e_i(X_i)}\right)\left(1-\frac{W_j}{e_j(X_j)}\right)\left(\widehat{\mu}^{G_2, (1)} (X_i)-\mu^{(1)}(X_i)\right)\left(\widehat{\mu}^{G_2, (1)} (X_j)-\mu^{(1)}(X_j)\right) | \mathcal{F}_{G_2}, \mathcal{F}_{j-1}, X_j] \\
   =& \E [\left(1-\frac{W_i}{e_i(X_i)}\right)\E[\left(1-\frac{W_j}{e_j(X_j)}\right)| \mathcal{F}_{G_2}, \mathcal{F}_{j-1}, X_j]\left(\widehat{\mu}^{G_2, (1)} (X_i)-\mu^{(1)}(X_i)\right)\left(\widehat{\mu}^{G_2, (1)} (X_j)-\mu^{(1)}(X_j)\right) ] = 0 
\end{align*}
Plugging these into equation \ref{eq:regret1, with X, 1} we have
\begin{equation}
   \begin{aligned}
      \E \left( \sum_{i \in G_1} \left(1-\frac{W_i}{e_i(X_i)}\right) \left(\widehat{\mu}^{G_2, (1)} (X_i)-\mu^{(1)}(X_i)\right)\right)^2 
      = \sum_{i \in G_1} \E[\frac{1 - e_i(X_i)}{e_i(X_i)} \left(\widehat{\mu}^{G_2, (1)} (X_i)-\mu^{(1)}(X_i)\right)^2 ] \\
   \end{aligned}
\end{equation}
   Now we consider whether $ \mathcal{E} $ happens. 
   
  \noindent \textbf{case1: $ \mathcal{E} $ doesn't happen.} \\
   Since for any $ w \in \{0, 1\} $, $ -M \leq \tau \leq M  $, 
   we have
   \begin{equation}
      \E(\widehat{\tau}_1^X - \widehat{\tau}_2^X)^2 \leq 4M^2
   \end{equation}
   \textbf{case2: $ \mathcal{E} $ happens.}
   By equation \ref{eq:propensity ratio bound}, we can bound the estimation error of $ \widehat{\mu}$ as follows:
   \begin{equation}
      \begin{aligned}
      \label{eq:with X, mean precision, 1}
         \E[(\widehat{\mu}^{G_2, (1)}(X) - \mu^{(1)}(X))^2] 
         &\leq C_4(\inf_{x \in \mathcal{X}}\frac{\widetilde{\sigma}_{K-1}^{G_2, (1)}(x) }{\widetilde{\sigma}_{K-1} ^{G_2, (1)}(x) + \widetilde{\sigma}_{K-1} ^{G_2, (0)}(x)}\gamma_K/4)^{-\alpha}(\log n)(\sup_{x \in \mathcal{X}} \sigma^{(1)}(x))^2 \\
         &= C_4\sup_{x \in \mathcal{X}}(1 + \frac{\widetilde{\sigma}_{K-1}^{G_2, (0)}(x) }{\widetilde{\sigma}_{K-1} ^{G_2, (1)}(x)})^{\alpha} (\gamma_K/4)^{-\alpha}(\log n)(\sup_{x \in \mathcal{X}} \sigma^{(1)}(x))^2 \\
         &\leq C_4\sup_{x \in \mathcal{X}}(1 + H^2\frac{\sigma^{(0)}(x)}{\sigma^{(1)}(x)})^{\alpha} (\gamma_K/4)^{-\alpha}(\log n)(\sup_{x \in \mathcal{X}} \sigma^{(1)}(x))^2 \\
         &\leq C_4(1 + H^2\frac{\sup_{x \in \mathcal{X}} \sigma^{(0)}(x)}{\inf_{x \in \mathcal{X}} \sigma^{(1)}(x)})^{\alpha} (\gamma_K/4)^{-\alpha}(\log n)(\sup_{x \in \mathcal{X}} \sigma^{(1)}(x))^2 \\
         &\leq C_4C'^2(1 + H^2 C'\frac{\inf_{x \in \mathcal{X}} \sigma^{(0)}(x)}{\inf_{x \in \mathcal{X}} \sigma^{(1)}(x)})^{\alpha} (\gamma_K/4)^{-\alpha}(\log n)(\inf_{x \in \mathcal{X}} \sigma^{(1)}(x))^2 \\
      \end{aligned}
   \end{equation}
   By equation \ref{eq:with X, mean precision, 1} and the fact that $ \alpha \leq 1 $, for any $ i \in G_1 $, we have
   \begin{align*}
      \E[\frac{1 - e_i(X)}{e_i(X)}(\mu^{(1)}(X) - \widetilde{\mu}^{G_2, (1)}(X))^2] 
      \leq& C_4C'^2(1 + H^2C'\frac{\inf_{x \in \mathcal{X}} \sigma^{(0)}(x)}{\inf_{x \in \mathcal{X}} \sigma^{(1)}(x)})^{\alpha} (n/8)^{-\alpha}(\log n)(\inf_{x \in \mathcal{X}} \sigma^{(1)}(x))^2 \\
      \leq& 8C_4C'^2(1 + H^2C'\frac{\inf_{x \in \mathcal{X}} \sigma^{(0)}(x)}{\inf_{x \in \mathcal{X}} \sigma^{(1)}(x)})n^{-\alpha}(\log n)(\inf_{x \in \mathcal{X}} \sigma^{(1)}(x))^2 \\
      \leq& 8C_4C'^2(1 + H^2 C')n^{-\alpha}\log n \E[(\sigma^{(0)}(X) + \sigma^{(1)}(X))^2]\\ 
   \end{align*}
   By equation \ref{eq:regret1, with X, 1}, we have 
   \begin{align*}
      \E \left( \sum_{i \in G_1} \left(1-\frac{W_i}{e_i(X_i)}\right) \left(\widehat{\mu}^{G_2, (1)} (X_i)-\mu^{(1)}(X_i)\right)\right)^2 
      \leq 8|G_1|C_4C'^2(1 + H^2 C')n^{-\alpha}\log n \E[(\sigma^{(0)}(X) + \sigma^{(1)}(X))^2]\\ 
   \end{align*}
   By a symmetry argument, 
   \begin{align*}
      \E \left( \sum_{i \in G_1} \left(1-\frac{W_i}{e_i(X_i)}\right) \left(\widehat{\mu}^{G_2, (0)} (X_i)-\mu^{(0)}(X_i)\right)\right)^2 
      \leq 8|G_1|C_4C'^2(1 + H^2 C')n^{-\alpha}\log n \E[(\sigma^{(0)}(X) + \sigma^{(1)}(X))^2]\\ 
   \end{align*}
   \begin{align*}
      \E \left( \sum_{i \in G_2} \left(1-\frac{W_i}{e_i(X_i)}\right) \left(\widehat{\mu}^{G_1, (1)} (X_i)-\mu^{(1)}(X_i)\right)\right)^2 
      \leq 8|G_2|C_4C'^2(1 + H^2C')n^{-\alpha}(\log n )\E[(\sigma^{(0)}(X) + \sigma^{(1)}(X))^2]\\ 
   \end{align*}
   \begin{align*}
      \E \left( \sum_{i \in G_2} \left(1-\frac{W_i}{e_i(X_i)}\right) \left(\widehat{\mu}^{G_1, (0)} (X_i)-\mu^{(0)}(X_i)\right)\right)^2 
      \leq 8|G_2|C_4C'^2(1 + H^2 C')n^{-\alpha}(\log n )\E[(\sigma^{(0)}(X) + \sigma^{(1)}(X))^2]\\ 
   \end{align*}
   Hence by \ref{proof:eq:model estimation, Cauchy}, 
   \begin{align*}
      \E[(\widehat{\tau}_1^X - \widehat{\tau}_2^X)^2 | \mathcal{E} ] &\leq \frac{4}{n^2}16nC_4C'^2(1 + H^2C')n^{-\alpha}\log n \E[(\sigma^{(0)}(X) + \sigma^{(1)}(X))^2]\\ 
      &= 64C_4C'^2(1 + H^2C')n^{-1-\alpha}(\log n )\E[(\sigma^{(0)}(X) + \sigma^{(1)}(X))^2]\\ 
   \end{align*}
   Finally, by the law of total probability, 
   \begin{align*}
      \E[(\widehat{\tau}_1^X - \widehat{\tau}_2^X)^2 ] &\leq \E[(\widehat{\tau}_1^X - \widehat{\tau}_2^X)^2 | \mathcal{E}] 
      + \Pr[\neg \mathcal{E}] \E[(\widehat{\tau}_1^X - \widehat{\tau}_2^X)^2 | \neg \mathcal{E}] \\
      &= 64C_4C'^2(1 + H^2C')n^{-1-\alpha}\log n \E[(\sigma^{(0)}(X) + \sigma^{(1)}(X))^2] + \frac{4M^2}{n^2}\\ 
   \end{align*}
\subsubsection{Proof of lemma \ref{lem:propensity optimization regret-feature}}
We first analyze propensity score regret in group $G_1$.
Note that there are $ \gamma_k/2$ periods in batch $ k $ belonging to group $G_1$, hence
\begin{align}
   &\E[\sum_{i \in G_1} \frac{(\sigma^{(1)}(X_i))^2}{e_i(X_i)} +\sum_{i\in G_1} \frac{(\sigma^{(0)}(X_i))^2}{1-e_i(X_i)}] \nonumber \\
   =& \sum_{k=1}^{K} \gamma_k/2 \E[\frac{(\sigma^{(1)}(X))^2(\widetilde{\sigma}_{k-1}^{G_1, (1)}(X) + \widetilde{\sigma}_{k-1}^{G_1, (0)}(X))}{\widetilde{\sigma}_{k-1}^{G_1, (1)}(X)} + \frac{(\sigma^{(0)}(X))^2(\widetilde{\sigma}_{k-1}^{G_1, (1)}(X) + \widetilde{\sigma}_{k-1}^{G_1, (0)}(X))}{\widetilde{\sigma}_{k-1}^{G_1, (0)}(X)}] \label{eq:regret2 with X, 1} \\
   =& \sum_{k=1}^{K} \gamma_k/2 \E[(\sigma^{(1)}(X) + \sigma^{(0)}(X))^2] \nonumber \\
   &+ \sum_{k=1}^{K} \gamma_k/2 \E[\frac{(\sigma^{(1)}(X))^2(\widetilde{\sigma}_{k-1}^{G_1, (1)}(X) + \widetilde{\sigma}_{k-1}^{G_1, (0)}(X))}{\widetilde{\sigma}_{k-1}^{G_1, (1)}(X)} + \frac{(\sigma^{(0)}(X))^2(\widetilde{\sigma}_{k-1}^{G_1, (1)}(X) + \widetilde{\sigma}_{k-1}^{G_1, (0)}(X))}{\widetilde{\sigma}_{k-1}^{G_1, (0)}(X)} \nonumber \\
   & \quad \quad \quad \quad \quad - (\sigma^{(1)}(X) + \sigma^{(0)}(X))^2] \nonumber \\
   =& n/2\E[(\sigma^{(1)}(X) + \sigma^{(0)}(X))^2] + \sum_{k=1}^{K} \gamma_k/2 \E[(\frac{\sigma^{(1)}(X)}{\widetilde{\sigma}_{k-1}^{G_1, (1)}(X)} - \frac{\sigma^{(0)}(X)}{\widetilde{\sigma}_{k-1}^{G_1, (0)}(X)})^2\widetilde{\sigma}_{k-1}^{G_1, (1)}(X)\widetilde{\sigma}_{k-1}^{G_1, (0)}(X)]  \nonumber \\
   \leq& n/2\E[(\sigma^{(1)}(X) + \sigma^{(0)}(X))^2] + \sum_{k=1}^{K} \gamma_k \E[((\frac{\sigma^{(1)}(X)}{\widetilde{\sigma}_{k-1}^{G_1, (1)}(X)} - 1)^2 + (\frac{\sigma^{(0)}(X)}{\widetilde{\sigma}_{k-1}^{G_1, (0)}(X)} - 1)^2)\widetilde{\sigma}_{k-1}^{G_1, (1)}(X)\widetilde{\sigma}_{k-1}^{G_1, (0)}(X)]  \nonumber \\
   =& n/2\E[(\sigma^{(1)}(X) + \sigma^{(0)}(X))^2] + \sum_{k=1}^{K} \gamma_k \E[(\sigma^{(1)}(X) - \widetilde{\sigma}_{k-1}^{G_1, (1)}(X))^2\frac{\widetilde{\sigma}_{k-1}^{G_1, (0)}(X)}{\widetilde{\sigma}_{k-1}^{G_1, (1)}(X)} + (\sigma^{(0)}(X) - \widetilde{\sigma}_{k-1}^{G_1, (0)}(X))^2\frac{\widetilde{\sigma}_{k-1}^{G_1, (1)}(X)}{\widetilde{\sigma}_{k-1}^{G_1, (0)}(X)}] \label{eq:regret2 with X, 2}
\end{align}
   Now we consider whether $ \mathcal{E} $ happens.
   
   \textbf{case1: $ \mathcal{E} $ doesn't happen.}
   Then by equation (\ref{eq:regret2 with X, 1}) and the definition of $ \widetilde{\sigma}_{k-1}^{G_1}(x) $, we have
   \begin{align*}
      \E[\sum_{i \in G_1} \frac{(\sigma^{(1)}(X_i))^2}{e_i(X_i)} +\sum_{i\in G_1} \frac{(\sigma^{(0)}(X_i))^2}{1-e_i(X_i)} | \neg \mathcal{E}] 
      &\leq \sum_{k=1}^{K} (\gamma_k/2)n\E[(\sigma^{(1)}(X))^2 + (\sigma^{(0)}(X))^2] \\
      &= n^2/2\E[(\sigma^{(1)}(X))^2 + (\sigma^{(0)}(X))^2]
   \end{align*}
   \textbf{case2: $ \mathcal{E} $ happens.}
   By equation \ref{eq:propensity ratio bound}, we can bound the estimation error of $ \widehat{\sigma}_{k}^{G_1, (1)} $ as follows:
   \begin{equation}
      \begin{aligned}
      \label{eq:with X, variance precision, 1}
         \E[(\sigma^{(1)}(X)  - \widehat{\sigma}_{k}^{G_1, (1)}(X) )^2] 
         &\leq C_4(\inf_{x \in \mathcal{X}}\frac{\widetilde{\sigma}_{k-1}^{G_1, (1)}(x) }{\widetilde{\sigma}_{k-1} ^{G_1, (1)}(x) + \widetilde{\sigma}_{k-1} ^{G_1, (0)}(x)}\gamma_k/4)^{-\beta}(\log n)(\sup_{x \in \mathcal{X}} \sigma^{(1)}(x))^2 \\
         &= C_4\sup_{x \in \mathcal{X}}(1 + \frac{\widetilde{\sigma}_{k-1}^{G_1, (0)}(x) }{\widetilde{\sigma}_{k-1} ^{G_1, (1)}(x)})^{\beta} (\gamma_k/4)^{-\beta}(\log n)(\sup_{x \in \mathcal{X}} \sigma^{(1)}(x))^2 \\
         &\leq C_4\sup_{x \in \mathcal{X}}(1 + H^2\frac{\sigma^{(0)}(x)}{\sigma^{(1)}(x)})^{\beta} (\gamma_k/4)^{-\beta}(\log n)(\sup_{x \in \mathcal{X}} \sigma^{(1)}(x))^2 \\
         &\leq C_4(1 + H^2\frac{\sup_{x \in \mathcal{X}} \sigma^{(0)}(x)}{\inf_{x \in \mathcal{X}} \sigma^{(1)}(x)})^{\beta} (\gamma_k/4)^{-\beta}(\log n)(\sup_{x \in \mathcal{X}} \sigma^{(1)}(x))^2 \\
         &\leq C_4C'^2(1 + H^2C'\frac{\inf_{x \in \mathcal{X}} \sigma^{(0)}(x)}{\inf_{x \in \mathcal{X}} \sigma^{(1)}(x)})^{\beta} (\gamma_k/4)^{-\beta}(\log n)(\inf_{x \in \mathcal{X}} \sigma^{(1)}(x))^2 \\
      \end{aligned}
   \end{equation}
   Now we analyse the terms in equation (\ref{eq:regret2 with X, 2}) to get the final result.
   For any $ k \in[K] $, for any $ x \in \mathcal{X} $, if $ \widehat{\sigma}_{k-1}^{G_1, (1)}(x) \geq \frac{\gamma_1}{\gamma_k}\widehat{\sigma}_{k-1}^{G_1, (1)}(x) $ or $ \sigma^{(1)}(x) \geq \frac{\gamma_1}{\gamma_k}\widehat{\sigma}_{k-1}^{G_1, (0)}(x) $, 
   then $ (\sigma^{(1)}(x) - \widetilde{\sigma}_{k-1}^{G_1, (1)}(x))^2 \leq (\sigma^{(1)}(x) - \widehat{\sigma}_{k-1}^{G_1, (1)}(x))^2 $ which means
   \begin{equation}
      \begin{aligned}
      \label{eq:with X, variance precision, 2}
      (\sigma^{(1)}(x) - \widetilde{\sigma}_{k-1}^{G_1, (1)}(x))^2\frac{\widetilde{\sigma}_{k-1}^{G_1, (0)}(x)}{\widetilde{\sigma}_{k-1}^{G_1, (1)}(x)} 
      &\leq (\sigma^{(1)}(x) - \widehat{\sigma}_{k-1}^{G_1, (1)}(x))^2\frac{\widetilde{\sigma}_{k-1}^{G_1, (0)}(x)}{\widetilde{\sigma}_{k-1}^{G_1, (1)}(x)} \\  
      &\leq (\sigma^{(1)}(x) - \widehat{\sigma}_{k-1}^{G_1, (1)}(x))^2(1 + \frac{\widehat{\sigma}_{k-1}^{G_1, (0)}(x)}{\widehat{\sigma}_{k-1}^{G_1, (1)}(x)}) \\
      &\leq (\sigma^{(1)}(x) - \widehat{\sigma}_{k-1}^{G_1, (1)}(x))^2(1 + H^2\frac{\sigma^{(1)}(x)}{\sigma^{(0)}(x)}) \\
      &\leq (\sigma^{(1)}(x) - \widehat{\sigma}_{k-1}^{G_1, (1)}(x))^2(1+H^2C'\frac{\inf_{x' \in \mathcal{X}} \sigma^{(0)}(x')}{\inf_{x' \in \mathcal{X}} \sigma^{(1)}(x')}) \\  
      \end{aligned}
   \end{equation}
   Otherwise, if $ \widehat{\sigma}_{k-1}^{G_1, (1)}(x) < \frac{\gamma_1}{\gamma_k}\widehat{\sigma}_{k-1}^{G_1, (1)}(x) $ and $ \sigma^{(1)}(x) < \frac{\gamma_1}{\gamma_k}\widehat{\sigma}_{k-1}^{G_1, (0)}(x) $, 
   then $ \widetilde{\sigma}_{k-1}^{G_1, (1)}(x) = \widehat{\sigma}_{k-1}^{G_1, (0)}(x)\gamma_1/\gamma_k $ and $ \widetilde{\sigma}_{k-1}^{G_1, (0)}(x) = \widehat{\sigma}_{k-1}^{G_1, (0)}(x) $ which means
   \begin{equation}
      \label{eq:with X, variance precision, 3}
      \begin{aligned}
      (\sigma^{(1)}(x) - \widetilde{\sigma}_{k-1}^{G_1, (1)}(x))^2\frac{\widetilde{\sigma}_{k-1}^{G_1, (0)}(x)}{\widetilde{\sigma}_{k-1}^{G_1, (1)}(x)} 
      &\leq (\widetilde{\sigma}_{k-1}^{G_1, (1)}(x))^2\frac{\widetilde{\sigma}_{k-1}^{G_1, (0)}(x)}{\widetilde{\sigma}_{k-1}^{G_1, (1)}(x)} \\
      &= \frac{\gamma_1}{\gamma_k}(\widehat{\sigma}_{k-1}^{G_1, (0)}(x))^2 \\
      &\leq 4\frac{n^{1/S_K}(\log n)^{1-1/S_K} }{\gamma_k}(\sigma^{(0)}(x))^2
      \end{aligned}
   \end{equation}
   Combining the equation \ref{eq:with X, variance precision, 2} and \ref{eq:with X, variance precision, 3}, we have
   \begin{align*}
      \forall x \in \mathcal{X}, 
      (\sigma^{(1)}(x) - \widetilde{\sigma}_{k-1}^{G_1, (1)}(x))^2\frac{\widetilde{\sigma}_{k-1}^{G_1, (0)}(x)}{\widetilde{\sigma}_{k-1}^{G_1, (1)}(x)} 
      \leq& (\sigma^{(1)}(x) - \widehat{\sigma}_{k-1}^{G_1, (1)}(x))^2(1 + H^2C'\frac{\inf_{x' \in \mathcal{X}} \sigma^{(0)}(x')}{\inf_{x' \in \mathcal{X}} \sigma^{(1)}(x')}) \\  
      &+ 4\frac{n^{1/S_K}(\log n)^{1-1/S_K} }{\gamma_k}(\sigma^{(0)}(x))^2
   \end{align*}
   Let $ x $ take expectation over $ P_X $ and we have
   \begin{equation}
   \begin{aligned}
      \E[(\sigma^{(1)}(X) - \widetilde{\sigma}_{k-1}^{G_1, (1)}(X))^2\frac{\widetilde{\sigma}_{k-1}^{G_1, (0)}(X)}{\widetilde{\sigma}_{k-1}^{G_1, (1)}(X)}]
      \leq& \E[(\sigma^{(1)}(X) - \widehat{\sigma}_{k-1}^{G_1, (1)}(X))^2](1 + H^2C'\frac{\inf_{x' \in \mathcal{X}} \sigma^{(0)}(x')}{\inf_{x' \in \mathcal{X}} \sigma^{(1)}(x')}) \\
      &+ 4\frac{n^{1/S_K}(\log n)^{1-1/S_K} }{\gamma_k}\E[(\sigma^{(0)}(X))^2] \\
   \end{aligned}
   \end{equation}
   By equation \ref{eq:with X, variance precision, 1} and the fact that $ \beta \leq 1 $, we have
   \begin{align*}
      \E[[(\sigma^{(1)}(X) - \widetilde{\sigma}_{k-1}^{G_1, (1)}(X))^2\frac{\widetilde{\sigma}_{k-1}^{G_1, (0)}(X)}{\widetilde{\sigma}_{k-1}^{G_1, (1)}(X)}]] 
      \leq& C_4C'^2(1 + H^2C'\frac{\inf_{x \in \mathcal{X}} \sigma^{(0)}(x)}{\inf_{x \in \mathcal{X}} \sigma^{(1)}(x)})^{\beta} (\gamma_{k-1} /4)^{-\beta}(\log n)(\inf_{x \in \mathcal{X}} \sigma^{(1)}(x))^2 \\
      \leq& 4C_4C'^2(1 + H^2C'\frac{\inf_{x \in \mathcal{X}} \sigma^{(0)}(x)}{\inf_{x \in \mathcal{X}} \sigma^{(1)}(x)})(\gamma_{k-1} )^{-\beta}(\log n)(\inf_{x \in \mathcal{X}} \sigma^{(1)}(x))^2 \\
      \leq& 4C_4C'^2(1 + H^2C')\gamma_{k-1}^{-\beta}\log n \E[(\sigma^{(0)}(X) + \sigma^{(1)}(X))^2]\\ 
      \leq& 4C_4C'^2(1 + H^2C')\frac{n^{1/S_K}(\log n)^{1 - \beta- 1/S_K} }{\gamma_k}\log n \E[(\sigma^{(0)}(X) + \sigma^{(1)}(X))^2]\\ 
      =& 4C_4C'^2(1 + H^2C')\frac{n^{1/S_K}(\log n)^{2 - \beta -1/S_K} }{\gamma_k}\E[(\sigma^{(0)}(X) + \sigma^{(1)}(X))^2]
   \end{align*}
   By a symmetry argument, 
   \begin{align*}
      \E[[(\sigma^{(0)}(X) - \widetilde{\sigma}_{k-1}^{G_1, (0)}(X))^2\frac{\widetilde{\sigma}_{k-1}^{G_1, (1)}(X)}{\widetilde{\sigma}_{k-1}^{G_1, (0)}(X)}]] 
      \leq& 4C_4C'^2(1 + H^2C')\frac{n^{1/S_K}(\log n)^{2 - \beta -1/S_K} }{\gamma_k}\E[(\sigma^{(0)}(X) + \sigma^{(1)}(X))^2]
   \end{align*}
   For every $ k \in [K] $, plugging these into equation (\ref{eq:regret2 with X, 2}), with $ \mathcal{E} $ happens, we have 
   \begin{align*}
      &\E[\sum_{i \in G_1} \frac{(\sigma^{(1)}(X_i))^2}{e_i(X_i)} +\sum_{i\in G_1} \frac{(\sigma^{(0)}(X_i))^2}{1-e_i(X_i)} | \mathcal{E}] \\
      &\leq n/2\E[(\sigma^{(1)}(X) + \sigma^{(0)}(X))^2] + 8KC_4C'^2(1 + H^2C')n^{1/S_K}(\log n)^{2 - \beta -1/S_K}\E[(\sigma^{(0)}(X) + \sigma^{(1)}(X))^2]
   \end{align*}
   By the Law of Total Probability, we have 
   \begin{align*}
      &\E[\sum_{i \in G_1} \frac{(\sigma^{(1)}(X_i))^2}{e_i(X_i)} +\sum_{i\in G_1} \frac{(\sigma^{(0)}(X_i))^2}{1-e_i(X_i)}] \\
      =& \Pr[\mathcal{E}]\E[\sum_{i \in G_1} \frac{(\sigma^{(1)}(X_i))^2}{e_i(X_i)} +\sum_{i\in G_1} \frac{(\sigma^{(0)}(X_i))^2}{1-e_i(X_i)} | \mathcal{E}]
      + \Pr[\neg \mathcal{E}]\E[\sum_{i \in G_1} \frac{(\sigma^{(1)}(X_i))^2}{e_i(X_i)} +\sum_{i\in G_1} \frac{(\sigma^{(0)}(X_i))^2}{1-e_i(X_i)} | \neg \mathcal{E}] \\
      \leq& n/2\E[(\sigma^{(1)}(X) + \sigma^{(0)}(X))^2] + 8C_4C'^2(1 + H^2C')Kn^{1/S_K}(\log n)^{2 - \beta -1/S_K}\E[(\sigma^{(0)}(X) + \sigma^{(1)}(X))^2] \\
      &+ 1/2\E[(\sigma^{(1)}(X) + \sigma^{(0)}(X))^2]
   \end{align*}
   Finally, by a symmetry argument on $ G_2 $, we have
   \begin{align*}
      &\E[\sum_{i \in G_2} \frac{(\sigma^{(1)}(X_i))^2}{e_i(X_i)} +\sum_{i\in G_2} \frac{(\sigma^{(0)}(X_i))^2}{1-e_i(X_i)}] \\
      \leq& n/2\E[(\sigma^{(1)}(X) + \sigma^{(0)}(X))^2] + 8C_4C'^2(1 + H^2C')Kn^{1/S_K}(\log n)^{2 - \beta -1/S_K}\E[(\sigma^{(0)}(X) + \sigma^{(1)}(X))^2] \\
      &+ 1/2\E[(\sigma^{(1)}(X) + \sigma^{(0)}(X))^2]
   \end{align*}
   Add the above two equations, we get the final result as follows:
   \begin{align*}
      &\E[\sum_{i=1}^{n} \frac{(\sigma^{(1)}(X_i))^2}{e_i(X_i)} +\sum_{i\in G_2} \frac{(\sigma^{(0)}(X_i))^2}{1-e_i(X_i)}] \\
      \leq& n\E[(\sigma^{(1)}(X) + \sigma^{(0)}(X))^2] + 16C_4C'^2(1 + H^2C')Kn^{1/S_K}(\log n)^{2 - \beta -1/S_K}\E[(\sigma^{(0)}(X) + \sigma^{(1)}(X))^2] \\
      &+ \E[(\sigma^{(1)}(X) + \sigma^{(0)}(X))^2]
   \end{align*}

\subsection{Proof of Theorem \ref{thm-lowerbound-GNeyman}}

We will first formally define notations. Assume that  covariate $X$ is generated from distribution $P_X$ with density function $p(x)$  and all $x \in \mathcal{X}$. 
Given covariate $x$, the outcomes are generated from Gaussian distribution 
$Y^{(1)}(x) \sim \mathcal{N}(\muone(x),\sigmaone(x)^2)$,
$Y^{(0)}(x) \sim \mathcal{N}(\muzero(x),\sigmazero(x)^2)$.
We also define a family of functions 
$\muone_{\theta}(x)=\muone(x)+\theta g^{(1)}(x)$ and $\muzero_{\theta}(x)=\muzero(x)+\theta g^{(0)}(x)$ for some perturbation functions $g^{(1)}(x)$ and $g^{(0)}(x)$
to be specified later. Since the function class $\mathcal{F}_{\mu}$ is an open set under infinite norm, we know that $\muone_{\theta}(x), \muzero_{\theta}(x) \in \mathcal{F}_{\mu}$ for sufficiently small $\theta$. We also define perturbation of the density function
as $p_{\theta}(x)=p(x)(1+\theta h(x))$ for some $h(x)$ (also to be specified later) satisfying $E_X(h(X))=0$ and $\theta$ sufficiently small such that it's a valid density function.  We also denote the distribution as $P_{\theta,X}$.
Recall that we use $\mathcal{F}_t=\{(X_i,W_i,Y_i)\}_{i=1}^t$
to denote the historical data at time period $t$.

As the classic way of proving Cramer-Rao lower bound, we need to calculate the log-likelihood as well as the score function.
For any non-anticipating allocation mechanism \texttt{ALG},  likelihood function of $\F_n$ can be formulated as
\[
P_{\theta}^{\texttt{ALG}}
(\mathcal{F}_n)=\prod_{t=1}^n P_{\theta,X}(X_t)\prod_{t=1}^n
\texttt{ALG}(W_t|X_t, \F_{t-1})
\prod_{t=1}^n \frac{1}{\sqrt{2\pi \sigma^{(W_t)}(X_t)^2}} \exp\left(-\frac{(Y_t-\mu_{\theta}^{(W_t)}(X_t))^2}{2\sigma^{(W_t)}(X_t)^2}\right),
\]
where we use the notation $P_{\theta}^{\texttt{ALG}}$ to 
account for the randomness under instance $P_{\theta,X}$, 
$\muone_{\theta}(X)$, $\muzero_{\theta}(X)$ and also the randomness from allocation mechanism \texttt{ALG}.
The first product accounts for the randomness of covariate vector $X_t$, the second for the randomness of allocation $W_t$ given $X_t$ and historical information $\F_{t-1}$, and the third for the (Gaussian) randomness of outcome given covariate $X_t$ and allocation $W_t$.
Since the mechanism is non-anticipating, we know that the randomness of the second term $\prod_{t=1}^n
\texttt{ALG}(W_t|X_t, \F_{t-1})$ is independent of the parameter $\theta$.
Therefore, the score function (i.e., the derivative of log-likelihood function with respect to $\theta$) is
\begin{equation}\label{equ-scorefunction}
    \begin{aligned}
        Z(\F_n)&:=\partial_{\theta} \log P_{\theta}^{\ALG}(\F_n)\left|_{\theta=0} \right.        =\frac{\partial_{\theta} P_{\theta}^{\ALG}(\F_n)}{P_{\theta}^{\ALG}(\F_n)} \left|_{\theta=0} \right.\\
        &        =\sum_{t=1}^n \partial_{\theta} \log  p(X_t)(1+\theta h(X_t)) -\sum_{t=1}^n
        \partial_{\theta} \frac{(Y_t-\mu_{\theta}^{(W_t)}(X_t))^2}{2\sigma^{(W_t)}(X_t)^2}\left|_{\theta=0} \right.\\
        &=\sum_{t=1}^n h(X_t)
        -\sum_{t=1}^n\frac{g^{(W_t)}(X_t)}{\sigma^{(W_t)}(X_t)^2}
        (\mu^{(W_t)}(X_t)-Y_t),
    \end{aligned}
\end{equation}
and in the following we  will also use $\varepsilon_t:=\mu^{(W_t)}(X_t)-Y_t$ to denote the Gaussian noise at time $t$.

An unbiased estimator $T$ maps $\mathcal{F}_n$ to an estimation of treatment effect
$T(\mathcal{F}_n)$ satisfying
\[
\E_{\theta}^{\ALG} (T(\mathcal{F}_n))= \E_{\theta}^{\ALG}(\muone_{\theta}(x)- \muzero_{\theta}(x)):=\tau_{\theta}.
\]
By definition, we have 
\[
\E_{\theta}^{\ALG}(\muone_{\theta}(x)- \muzero_{\theta}(x))=
\int_{X} p(x)(1+\theta h(x))
(\muone(x)-\muzero(x)+\theta(g^{(1)}(x)-g^{(0)}(x))) dx.
\]
Take the derivatives on both sides and evaluate at $\theta=0$, we have
\begin{equation}\label{equ-CovZT}
    \begin{aligned}
    \partial_{\theta} E_{\theta}^{\ALG} (T(\mathcal{F}_n))\left|_{\theta=0} \right.&= \partial_{\theta} \int_{X} P_{\theta}^{\ALG}(\F_n) T(\mathcal{F}_n)\left|_{\theta=0} \right.= \int_{X} \partial_{\theta}P_{\theta}^{\ALG}(\F_n) T(\mathcal{F}_n)\left|_{\theta=0} \right.\\
    &=\int_X Z(\F_n)P_{\theta=0}^{\ALG}(\F_n) T(\F_n)=\E_{\theta=0}^{\ALG}(Z(\F_n)T(\F_n)),
\end{aligned}
\end{equation}
\begin{equation}\label{equ-deriv-ATE}
\begin{aligned}    \partial_{\theta}E_{\theta}^{\ALG}(\muone_{\theta}(x)- \muzero_{\theta}(x))\left|_{\theta=0} \right.&=
    \int_X p(x) \left(h(x)(\muone(x)-\muzero(x))+ g^{(1)}(x)-g^{(0)}(x)\right)\\
    &=\E_{P_X} \left(h(x)(\muone(x)-\muzero(x))+ g^{(1)}(x)-g^{(0)}(x)\right).
\end{aligned}
\end{equation}
In the following, the probability and expectation are all taken under the original instance where $\theta=0$, using Cauchy-Schwarz, we have 
\begin{equation}\label{equ-variance-cauchy}
    \begin{aligned}
        \Var(T(\F_n)) =\E(T(\F_n)-\E(T(\F_n)))^2
        \ge \frac{\E^2(Z(\F_n)(T(\F_n)-\E(T(\F_n))))}{\E(Z^2(\F_n))}=\frac{\E^2(Z(\F_n)T(\F_n)))}{\E(Z^2(\F_n))},
    \end{aligned}
\end{equation}
where we use the fact that
the expectation of score function $\E(Z(\F_n))=0$, which can be proved from equation (\ref{equ-scorefunction}) by observing that $\E(h(X)=0$ and 
$\E(\mu^{W}(X)-Y)=0$ for any fixed $X$, $W$.
By (\ref{equ-scorefunction}), we also have
\begin{equation}
    \begin{aligned}
       \E(Z^2(\F_n)) &=\E\left(\sum_{t=1}^n h(X_t)
        -\sum_{t=1}^n\frac{g^{(W_t)}(X_t)}{\sigma^{(W_t)}(X_t)^2}
        \varepsilon_t\right)^2\\
        &=n \E(h(X))^2+
      \E_{X_t\sim P_X} \E^{\ALG}  \sum_{t=1}^n \frac{g^{(W_t)}(X_t)^2}{\sigma^{(W_t)}(X_t)^2 }\\
       &=n \E(h(X))^2+
       n \E_{X\sim P_X, W \sim \widehat{\ALG}(.|x)}
       \frac{g^{(W)}(X)^2}{\sigma^{(W)}(X)^2 },
    \end{aligned}
\end{equation}
where the second equality holds since all the cross term has expectation zero, and $\Var(\varepsilon_t)=\sigma^{(W_t)}(X_t)^2$.
And in the last equation, the non-adaptive allocation mechanism  $\widehat{\ALG}$ is defined as the "average" propensity score for adaptive mechanism $\ALG$ in $n$ periods.
Specifically, for each covariate $X$, we define

\[
e^{\widehat{\ALG}}(X)=P^{\widehat{\ALG}}(W=1 |X)=
\frac{1}{n}\sum_{t=1}^n
\int P(W=1 |X ,\F_{t-1}) dP^{\ALG}(\F_{t-1})
\]
as the average over $n$ time periods under the randomness of \ALG and historical information  $\F_{t-1}$ for $t=1,\cdots, n$.

By (\ref{equ-CovZT}) and (\ref{equ-deriv-ATE}), we also have 
\[
        \E^2(Z(\F_n)T(\F_n))=
        \E^2_{P_X} \left(h(x)(\muone(x)-\muzero(x))+ g^{(1)}(x)-g^{(0)}(x)\right)
\]
Notice that if we can prove
\begin{equation}\label{equ-def-gh}
\begin{aligned}
    &\E^2(Z(\F_n)T(\F_n))\\
    =&
        \E^2_{P_X} \left(h(x)(\muone(x)-\muzero(x))+ g^{(1)}(x)-g^{(0)}(x)\right)\\
        =&\E^2_{P_X} \left(h(x)(\muone(x)-\muzero(x)- \E_X(\muone(x)-\muzero(x)))+ g^{(1)}(x)-g^{(0)}(x)\right)\\
        =&n\E_{P_X} \left(
        (\muone(x)-\muzero(x)-\E_X(\muone(x)-\muzero(x)))^2+ \frac{\sigmaone(x)^2}{e^{\widehat{\ALG}}(x)}
        +\frac{\sigmazero(x)^2}{1-e^{\widehat{\ALG}}(x)}\right)\\
        & \qquad        \times \E_{P_X}\left(h(X)^2+
      e^{\ALG}(x) \frac{g^{(1)}(x)^2}{\sigma^{(1)}(x)^2 }+ (1-e^{\ALG}) \frac{g^{(0)}(x)^2}{\sigma^{(0)}(x)^2 }
        \right),
\end{aligned}
\end{equation}
where the second equality holds since $\E_{P_X}(h(x))\E_{P_x}(\muone(x)-\muzero(x))=0$, then the proof is finished. This is because the first term
\[
\E_{P_X} \left(
        (\muone(x)-\muzero(x)-\E_X(\muone(x)-\muzero(x)))^2+ \frac{\sigmaone(x)^2}{e^{\widehat{\ALG}}(x)}
        +\frac{\sigmazero(x)^2}{1-e^{\widehat{\ALG}}(x)}\right)\ge V^{*}
\]
and the equality holds if and only if $e^{\widehat{\ALG}}(x)=e^{*}(x)$ almost surely, and  the second term
\[
  n\E_{P_X}\left(h(x)^2+
      e^{\ALG}(x) \frac{g^{(1)}(x)^2}{\sigma^{(1)}(x)^2 }+ (1-e^{\ALG}) \frac{g^{(0)}(x)^2}{\sigma^{(0)}(x)^2 }
        \right)=\E(Z^2(\F_n)).
\]
But the last equality in (\ref{equ-def-gh}) actually has a Cauchy-Schwarz form, so we can specifically design $h(x)$, $g^{(1)}(x)$ and $g^{(0)}(x)$ to make the equality holds.
Indeed, by Cauchy Schwarz
we have
\begin{equation}
\begin{aligned}
        &\E^2_{P_X} \left(h(x)(\muone(x)-\muzero(x))+ g^{(1)}(x)-g^{(0)}(x)\right)\\
        \le&\E_{P_X} \left(
        (\muone(x)-\muzero(x)-\E_X(\muone(x)-\muzero(x)))^2+ \frac{\sigmaone(x)^2}{e^{\widehat{\ALG}}(x)}
        +\frac{\sigmazero(x)^2}{1-e^{\widehat{\ALG}}(x)}\right)\E(Z^2(\F_n))        
\end{aligned}
\end{equation}
and the equality holds
if we take \[\frac{\muone(x)-\muzero(x)-\E_X(\muone(x)-\muzero(x))}{h(x)}=\frac{\sigmaone(x)^2}{e^{\widehat{\ALG}}(x) g^{(1)}(x)}=-\frac{\sigmazero(x)^2}{\left(1-e^{\widehat{\ALG}}(x)\right)g^{(0)}(x)}=1\]
for all $x$.
In other words, for any $x\in X$ we define 
\begin{equation}
    \begin{aligned}
        h(x)&=\muone(x)-\muzero(x)-\E_X(\muone(x)-\muzero(x)),\\
        g^{(1)}(x)&=\frac{\sigmaone(x)^2}{e^{\widehat{\ALG}}
        (x)},\\
        g^{(0)}(x)&=-\frac{\sigmazero(x)^2}{1-e^{\widehat{\ALG}}(x)},
    \end{aligned}
\end{equation}
 and we have proved  that $V^{*}$ a lower bound.

 Furthermore,
 we can analyze the Cauchy-Schwarz inequality in (\ref{equ-variance-cauchy}) 
 to have
 \begin{equation}
 \begin{aligned}
       &\Var(T(\F_n))-\frac{\E^2(Z(\F_n)T(\F_n)))}{\E(Z^2(\F_n))}\\
       =&\E\left(T(\F_n)-\E(T(\F_n))-\frac{\E(T(F_n)Z(\F_n))}{\E(Z^2(\F_n))}Z(\F_n)\right)^2\\
       =&\E\left(T(\F_n)-\E(T(\F_n))-\frac{1}{n}Z(\F_n)\right)^2\\
       =&\E\left(T(\F_n)-\E(T(\F_n))-\frac{1}{n}\sum_{t=1}^n \left(\tau(X_t)-\tau+ 
\frac{W_t(Y_t-\muone(X_t)}{e^{\ALG}(X_t)}-\frac{(1-W_t)(Y_t-\muzero(X_t)}{1-e^{\ALG}(X_t)}\right)\right)^2, 
 \end{aligned}
 \end{equation}
 where the first equality is simply given by the residual term of Cauchy-Schwarz inequality, the second inequality is shown by equality (\ref{equ-def-gh}), and the third one is by plugging in our specification of $g^{(1)}(x)$, $g^{(0)}(x)$ and $h(x)$. Specifically, we show that under our specification, the score function $Z(\F_n)$ is actually the AIPW estimator under allocation mechanism $\widehat{\ALG}$, where we denote $\tau(X)=\muone(X)-\muzero(X)$ as the conditional ATE and $\tau=\E(\tau(X))$ as the ATE. Combining everything together, we show that for all unbiased estimator $T$ and possibly adaptive allocation mechanism $\ALG$, 
 $V^{*}$ is a valid variance lower bound, and it's achieved only when the allocation algorithm is the generalized Neyman allocation $e^{*}(X)=\sigma^{(1)}(X)/(\sigma^{(1)}(X)+\sigma^{(0)}(X))$ and the estimator is the idealized AIPW estimator 
 \[
   \widehat{\tau}^{*X}=
        \sum_{i=1}^n \frac{1}{n} \left(\mu^{(1)}(X_i)-\mu^{(0)}(X_i)+ \frac{Y_i-\mu^{(1)}(X_i)}{e^*(X_i)}W_i -\frac{Y_i-\mu^{(0)}(X_i)}{1-e^*(X_i)}(1-W_i) \right).
 \]
 \subsection{Proof of  Corollary \ref{cor-lowerbound-Neyman}}
 The proof of \cref{cor-lowerbound-Neyman} follows exactly the same as theorem \ref{thm-lowerbound-GNeyman} by ignoring the existence of covariates $X$ and distribution $P_X$.

 \subsection{Proof of Theorem \ref{thm-CLT}}
The sufficient conditions in Theorem 1 in (\cite{cook2024semiparametric})
is already given by our offline regression oracle \ref{asm:oracle with convariate} and propensity score convergence guarantee in theorem \ref{thm-upperbound-nofeature} and lemma \ref{lem:propensity optimization regret-feature}.
\end{document}